%% file: main.tex
\documentclass[journal]{IEEEtran}
\newtheorem{thm}{Theorem}[section]

\newtheorem{prop}[thm]{Proposition}

\newtheorem{defn}[thm]{Definition}
\newtheorem{rem}[thm]{Remark}
\usepackage{authblk}
\usepackage{enumerate}
\usepackage{amsmath}
\usepackage{amssymb}
\usepackage{graphicx}
\usepackage[subrefformat=parens,labelformat=parens]{subfig}\usepackage{authblk}
\usepackage{enumerate}
\usepackage{amsmath}
\usepackage{amssymb}
\usepackage{textcomp}
\usepackage{graphicx}
\usepackage[subrefformat=parens,labelformat=parens]{subfig}
\usepackage{multirow}
\usepackage{graphicx} % for pdf, bitmapped graphics files
\usepackage{epsfig} % for postscript graphics files
\usepackage{epstopdf}
\usepackage{epsfig} % for postscript graphics files
\usepackage{cite}
\usepackage{algorithm}
\usepackage{algorithmic}
\usepackage{multirow}
\usepackage{rotating}
\usepackage{color}
\usepackage{url}
\graphicspath{{figures/}}

\begin{document}

\title{Control of Magnetic Microrobot Teams for Temporal Micromanipulation Tasks}% to perform efficient microassembly operations}

\author{Yiannis Kantaros,~\IEEEmembership{Student Member,~IEEE,} Benjamin V. Johnson,~\IEEEmembership{Student Member,~IEEE,}, Sagar Chowdhury,~\IEEEmembership{Member,~IEEE,} David J. Cappelleri,~\IEEEmembership{Member,~IEEE,} and Michael M. Zavlanos,~\IEEEmembership{Member,~IEEE}
        % <-this % stops a space
\thanks{Y, Kantaros and M. Zavlanos are with the Department
of Mechanical Engineering \& Material Science, Duke University, Durham,
NC, 27708 USA, e-mail: \{yiannis.kantaros,michael.zavlanos\}@duke.edu.  B.V. Johnson, S. Chowdhury, and D. Cappelleri are with the School of Mechanical Engineering, Purdue University, West Lafayette, IN, 47907 USA,  e-mail: \{john1360@purdue.edu,sagar353@gmail.com,dcappell@purdue.edu\}.  This work was supported by NSF grants IIS-1358446 and IIS-1302283. Opinions expressed are those of the authors and do not necessarily reflect opinions of the sponsors.  
}% <-this % stops a space
% \thanks{B. Johnson, S. Chowdhury, and D. Cappelleri are with the School of Mechanical Engineering, Purdue University, West Lafayette, IN, 47907 USA,  e-mail: \{john1360,sagar353,dcappell\}@purdue.edu}% <-this % stops a space
%\thanks{Manuscript received June X, 2017; revised Month XX, 2017.}
}

\maketitle

\begin{abstract}
\input{abstract}

\end{abstract}

\IEEEpeerreviewmaketitle

%%%%%%%%%%%%%%%%%%%%%%%%%%%%%%%%%
\section{Introduction}\label{sec:Intro}
% \input{Introduction}
% %%%%%%%%%%%%%%%%%%%%%%%%%%%%%%%%%

% %%%%%%%%%%%%%%%%%%%%%%%%%%%%%%%%%
% \section{Related Work}\label{sec:related}
\input{RelatedWork}
\section{Local Magnetic Field Actuation}\label{sec:LocalMagneticFieldActuation}
\input{LocalMagneticFieldActuation}

%%%%%%%%%%%%%%%%%%%%%%%%%%%%%%%%%

% %%%%%%%%%%%%%%%%%%%%%%%%%%%%%%%%%
% \section{LTL Motion Planning}\label{sec:LTLMotionPlanning}
% \input{LTLPlanning}
% %%%%%%%%%%%%%%%%%%%%%%%%%%%%%%%%%

%%%%%%%%%%%%%%%%%%%%%%%%%%%%%%%%%
\section{Problem Definition}\label{sec:ProblemDefinition}
\input{ProblemDefinition}
%%%%%%%%%%%%%%%%%%%%%%%%%%%%%%%%%

%%%%%%%%%%%%%%%%%%%%%%%%%%%%%%%%%
\section{Temporal Logic Planning for Microrobot Teams}\label{sec:TemporalPlanning}
\input{PlanningApproach}
%%%%%%%%%%%%%%%%%%%%%%%%%%%%%%%%%

%%%%%%%%%%%%%%%%%%%%%%%%%%%%%%%%%
\section{Simulation Results}\label{sec:Simulations}
\input{Simulation}

%%%%%%%%%%%%%%%%%%%%%%%%%%%%%%%%%

%\FloatBarrier
%%%%%%%%%%%%%%%%%%%%%%%%%%%%%%%%%%
\section{Experimental Results}\label{sec:ExperimentalResults}

\subsection{Mobile Microrobot System}\label{sec:microsystem}
\input{MobileMicrorobotSystem}

\subsection{Validation Experiments}
\input{Experiments}

%%%%%%%%%%%%%%%%%%%%%%%%%%%%%%%%%%

%%%%%%%%%%%%%%%%%%%%%%%%%%%%%%%%%%%
\section{Conclusions}\label{sec:Conclusions}
\input{Conclusion}
%%%%%%%%%%%%%%%%%%%%%%%%%%%%%%%%%%%

\bibliographystyle{IEEEtran}   
\bibliography{YK_bib,manipulation,Reference_sim_micro_multi_robot}

\end{document}

%% file: abstract.tex
In this paper, we present a control framework that allows magnetic microrobot teams to accomplish complex micromanipulation tasks captured by global Linear Temporal Logic (LTL) formulas. 
%Constraints specified by the physics of the problem need to be satisfied for all time as the microrobots navigate the workspace to accomplish the assigned tasks. These constraints are expressed by a global LTL specification. 
To address this problem, we propose an optimal control synthesis method that constructs discrete plans for the robots that satisfy both the assigned tasks as well as proximity constraints between the robots due to the physics of the problem. Our proposed algorithm relies on an existing optimal control synthesis approach combined with a novel sampling-based technique to reduce the state-space of the product automaton that is associated with the LTL specifications. The synthesized discrete plans are executed by the microrobots independently using local magnetic fields. Simulation studies show that the proposed algorithm can address large-scale planning problems that cannot be solved using existing optimal control synthesis approaches. %Finally, experimental results of the proposed microrobot control framework are presented using local magnetic fields.
Moreover, we present experimental results that also illustrate the potential of our method in practice. To the best of our knowledge, this is the first control framework that allows independent control of teams of magnetic microrobots for temporal micromanipulation tasks.

\textit{Keywords}:  Path Planning for Multiple Mobile Robot Systems, Micro/Nano Robots, Temporal Logic Planning, Optimal Control Synthesis

%% file: RelatedWork.tex
%Manipulation part: \textbf{Purdue}\textcolor{red}{SagarSagar}
%While manipulation with multiple robots have been widespread in macroscale in various industrial and academic settings, it faces numerous challenges in microscale not only because of the nature of physics at microscale but also the limited capabilities of the robots due to the size constraints  associated to the

%Manipulation in microscale requires us to build robots that are small in size that do not cover the whole workspace and capable of manipulating microscale objects. 

\IEEEPARstart{M}{anipulation} of microscale objects can be characterized by a number of unique features: (1) The manipulator has to be small relative to the workspace such that it does not cover the entire workspace, (2) The mass of the microscale object is very small which makes the surface related forces (e.g. surface tension, drag, viscous forces etc.) dominate the inertial forces (e.g. weight of the object), (3) Highly parallelized operation requires the robots/manipulators to be capable of coordinating with each other in manipulating a single object or multiple objects in a sequential manner. The ability to manipulate such objects autonomously in a sequential manner can potentially revolutionize the micro-assembly operations where a microscale component will be moved with a micro-robot to the workspace to be assembled with its counterpart carried by another robot to create a heterogenous product. These unique features give rise to a new set of constraints on the design of the robots as well as provide unique opportunities for the development of novel actuation and planning techniques.

In this paper, we propose a novel control framework for teams of magnetic microrobots responsible for accomplishing temporal micromanipulation tasks. We have utilized an 8 $\times$ 8 planar magnetic coil array from our previous work~\cite{Capp14,Chow15,Chow15a} and  
developed a new 11 $\times$ 11 planar magnetic coil array for use as an actuation system for independently controlling multiple microrobots. Each coil can be switched on or off independently creating a local magnetic field to only actuate the microrobot in its vicinity.  Therefore, the coil array is suitable for our control framework to achieve temporal actuation of microrobots. This magnetic substrate partitions the workspace into square cells giving rise to a transition system that models the robot motion. The micromanipulation tasks that the robots need to accomplish are captured using global Linear Temporal Logic (LTL) formulas that also incorporate proximity constraints that the robots need to satisfy during their motion, that are due to the physics of the problem. To generate high-level plans for the robots, we propose a novel optimal control synthesis algorithm that combines an existing temporal logic planning method with a new sampling-based technique that reduces the state-space of the product automaton associated with the LTL statement. Given the transition systems that abstract robot mobility, the key idea is to create smaller transition systems
that may not be as expressive as the original ones but they can still generate motion plans that satisfy the assigned LTL tasks. In this way, our method can solve large planning problems arising from large magnetic coil arrays, large numbers of robots, and complex temporal tasks, which is not possible using existing optimal control synthesis methods. Finally, given discrete high-level plans for the robots, we control the magnetic fields within each cell in the workspace so that the robots can navigate across cells as specified by the discrete high-level plans. We present local magnetic field actuation experiments that illustrate our approach. To the best of our knowledge, this is the first control framework for magnetic microrobots accomplishing temporal micromanipulation tasks that has also been successfully demonstrated in practice. A preliminary description of the proposed planning method is summarized in the extended abstract \cite{kantaros15asilomar}. Compared to \cite{kantaros15asilomar}, we analyze the complexity of the proposed algorithm while extensive simulation and experimental studies are provided.

% The unique features mentioned above impose some unique constraints on the design of the robots as well as open us some unique opportunities to develop actuation techniques for the robots. 

% These unique features give rise to a new set of constraints on the design of the robots as well as provide unique opportunities for the development of novel actuation and planning techniques.
%
\subsection{Related Work}
Contact based manipulation where two to four micromanipulators with point contacts are coordinated to grasp an object and transport it to a goal location have been demonstrated in~\cite{david1, seven, WasonTRO12, cappelleri2012cooperative, cappelleri2011two, cappelleri2011caging}. Rather than using multiple manipulators, a flexible microgripper attached to a multi DOF manipulator can be used to realize a pick-and-place operations as demonstrated in~\cite{YangNelsonGaines2005,Dechev04}. Manipulator and micro-gripper based manipulation for micro-assembly operations based on micro snap fasteners has also been demonstrated~\cite{Popa07, Ren07,Rabenorosoa09,Tamadazte09a}. In many of these applications, several parts need to be soldered for making stable electrical connections. Venkatesan and Cappelleri~\cite{Venkatesan17} have replaced one of the micromanipulators in their assembly cell with a soldering iron along with an automated solder feed mechanism to realize a flexible soldering station for micro-assembly operations.  These pick-and-place approaches are suitable for fast serialized operations where the gripper or manipulators are designed to handle only one type of object.

Another powerful approach for micromanipulation tasks is to replace the fixed manipulator with mobile microrobots. Each microrobot can be equipped with a customized gripper suitable for handling a particular object. Due to their small size, global actuation fields (e.g. magnetic field, optical field, physiological energy) are typically utilized as actuation mechanisms for microscale robots \cite{Chow15b}. However, actuating multiple robots with a global field is challenging as it is difficult to apply individual and/or independent control on the robots. 
%Researchers have proposed several approaches to address the individual robot with a global field to facilitate individual control. 
To address this challenge, optical manipulation methods have been recently developed where a single laser beam is split into multiple paths with the help of a spatial light modulator or high speed mirror technology~\cite{Bane12, Bane14, Chea15, Chea15a, Chow13b, Chow13a, Haoyao12, Hu11b, Hu12, Rahm17,Thak14}. However, optical manipulation is only suitable for objects that are less than 10 $\mu m$ in size due to the very small actuation force (on the order of pico-Newtons) that can be generated by the laser beam. 

Magnetic fields can generate forces as high as on the order of micro-Newtons and can also be modulated into various forms with minimum effort. Hence, magnetic fields have become a very useful tool to address a variety of applications from biology~\cite{Mair17,Mair17a,Jing14,ufsmm_tase,Jing14b,bi2018design} to micro assembly~\cite{Saka14}. To obtain independent control, there have been various approaches that introduce heterogeneity into the robots so that they can be affected by the same field in a significantly different fashion \cite{Dill12,Frut10,Chea14}. In this way, controllers can be developed that can control multiple robots somewhat independently of each other with the same global input \cite{Devo09}. Another approach is to utilize the influence zone of a particular magnetic coil and partition the workspace accordingly so that identical robots can be controlled by  dedicated coils without affecting the others \cite{Wong15}. Hybrid approaches, where a magnetic field actuates a robot while an electrostatic surface is selectively activated to make the other robot static have also been proposed for independent control with magnetic global fields~\cite{Pawa09a}. While the approaches presented above allow for different behaviors of the microrobots in the workspace, these motions are still coupled to some degree. Therefore, they are not useful when the robots need to truly independently operate in the workspace. Moreover, these approaches are not scalable since the complexity of coupled dynamics increases as the number of robots increases in the workspace. In addition, the controller needs to account for the motion constraints of the robots that might make some manipulation tasks in the workspace unattainable by the robots. In a truly independent system, the robots will not have any motion constraints due to the nature of actuation and hence will operate under fewer constraints that will help in achieving complex manipulation tasks.

To overcome the challenges in controlling microrobots independently using global magnetic fields, Pelrine {\em et al.}~\cite{Pelr12} developed a four layer printed circuit board that generates local magnetic fields for independent control of multiple homogeneous mm-scale robots.  Similarly, arrays of multiple mm-scale planar coils have been proposed in~\cite{Capp14} that generate local magnetic fields large enough to actuate multiple small scale robots independently.  This design was successfully experimentally demonstrated for autonomous navigation in~\cite{Chow15,Chow15a}. Chowdhury {\em et al.}~\cite{Chow16, Chow17} proposed a similar small scale array of current carrying wires arranged on a planar substrate.  The array can be provided with current individually to actuate micron sized robots independent of each other. 

The planning algorithm proposed in~\cite{Chow15a, Chow17} computes the waypoints for each robot separately, without considering the other robots, and does not consider proximity constraints between the robots due to the physics of the problem, e.g., the magnetic forces applied between robots.
Hence, the waypoints computed in~\cite{Chow15a, Chow17} need to be recomputed frequently to account for such constraints. 
Moreover, the planning problems addressed in~\cite{Chow15a, Chow17} are restricted to point-to-point navigation and can not incorporate high-level tasks, such as sequential or temporal navigation and operation, that are typical in manipulation.
Control synthesis for mobile robots under complex tasks, such as those of interest here, can be captured by Linear Temporal Logic (LTL) formulas.  These build upon either bottom-up approaches when independent LTL expressions are assigned to robots \cite{kress2009temporal,kress2007s,bhatia2010sampling} or top-down approaches when a global LTL formula describing a collaborative task is assigned to a team of robots \cite{chen2011synthesis,chen2012formal}. Common in the above works is that they rely on model checking theory \cite{baier2008principles,clarke1999model} to find paths that satisfy LTL-specified tasks, without optimizing task performance. Optimal control synthesis under local and global LTL specifications has been addressed in \cite{smith2010optimal,smith2011optimal,guo2013revising, guo2013reconfiguration,guo2015multi} and \cite{kloetzer2010automatic,ulusoy2013optimality,ulusoy2014optimal}, respectively. In top-down approaches \cite{kloetzer2010automatic,ulusoy2013optimality,ulusoy2014optimal}, optimal discrete plans are derived for every robot using the individual transition systems that capture robot mobility and a Non-deterministic B$\ddot{\text{u}}$chi Automaton (NBA) that represents the global LTL specification. Specifically, by taking the synchronous product among the transition systems and the NBA, a synchronous product automaton can be constructed. Then, representing the latter automaton as a graph and using graph-search techniques, optimal motion plans can be derived that satisfy the global LTL specification and optimize a cost function. As the number of robots or the size of the NBA increases, the state-space of the product automaton grows exponentially and, as a result, graph-search techniques become intractable. Consequently, these motion planning algorithms scale poorly with the number of robots and the complexity of the assigned task. To the contrary, the method we propose here scales to much larger planning problems that are typical for micromanipulation tasks. Sampling-based approaches for optimal temporal logic control synthesis have been recently proposed by the authors in \cite{kantaros2017Csampling}. Although \cite{kantaros2017Csampling} completely avoids the construction of the product automaton and, therefore, it is more resource efficient, it is probabilistically complete which means that it theoretically requires an infinite number of iterations to find a feasible motion plan. To the contrary, here we show that the proposed algorithm can find a feasible solution, if it exists, in a finite number of iterations. 

\subsection{Contributions}
The contributions of this work are the following. First, to the best of our knowledge, we present the first planning method for the independent control of teams of magnetic microrobots for temporal micromanipulation tasks, that has also been demonstrated successfully in practice. Second, our proposed planning method is able to solve much larger planning problems, that are typical for micromanipulation tasks, compared to existing optimal control synthesis techniques. Third, with respect to local magnetic field actuation of mobile microrobots, we have identified new diagonal static equilibrium points in the workspace that allow for additional waypoints for path planning and diagonal robot trajectories. We have also systematically characterized and utilized the surrounding coils of the robot to obtain predictable motions between workspace waypoints.  Finally, the experimental results of the proposed algorithm exhibit temporal synchronization of multiple magnetic microrobots being independently controlled in the workspace for the first time. Experiments with two, three, and four microrobots are presented.  Additionally, a two robot microassembly experiment has been demonstrated to show the potential for using the system for micromanipulation and assembly tasks.

% Finally,... [is there a new contribution from the perspective of microrobot design? what was challenging with the experiments that has not been addressed before, but were able to address?]

The rest of this paper is organized as follows. In Sec.~\ref{sec:LocalMagneticFieldActuation}, we present the theory behind using local magnetic field actuation for magnetic microrobot control.  The problem we are solving is defined in Sec.~\ref{sec:ProblemDefinition}.  Next, the proposed method for temporal planning for microrobot teams is described in Sec.~\ref{sec:TemporalPlanning} followed by simulation results using the method in Sec.~\ref{sec:Simulations}.  Our microrobot experimental testbed is described in Sec.~\ref{sec:ExperimentalResults} along with validation and multi-robot LTL experiments.  Finally, Sec.~\ref{sec:Conclusions} concludes the paper.

%% file: LocalMagneticFieldActuation.tex
\begin{figure}
\centering
\includegraphics[width=2.25in]{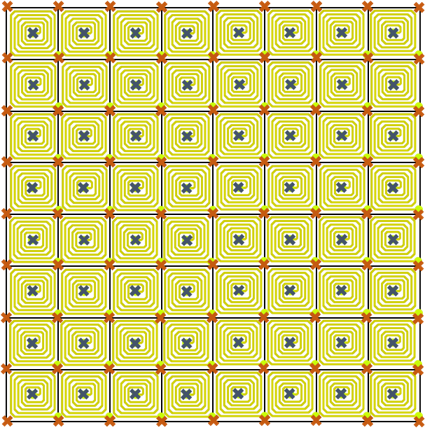}
\caption{Schematic of 8 $\times$ 8 planar magnetic coil array. Waypoints for the developed path planning algorithm consist of each coils center and corner (workspace diagonal) positions, all marked with an ``x".}
\label{fig:CoilArray}
\end{figure}

Consider the 8 $\times$ 8 planar array of magnetic coils developed in~\cite{Capp14, Chow15, Chow15a} and shown schematically in Fig.\ref{fig:CoilArray}, and permanent magnetic disc-shaped microrobots with diameters smaller than the footprint of a single coil.  Each magnetic coil has a radius of influence, $R_{I_{coil}}$, that corresponds to a distance from the coil center.  If a robot is within a distance of $R_{I_{coil}}$ from the center of the coil, it will be influenced when the coil is activated. If the magnetic robot is greater than $R_{I_{coil}}$ from the coil center, the activated coil will not affect the robot.

The magnetic field generated by the a single planar coil can be evaluated by considering it as a series of finite wire segments in a spiral fashion. 
The force $\vec{F}$ on a body with magnetization $\vec{m}$ due to an external magnetic flux density $\vec{B}$ generated by single coil is $\vec{F} = (\vec{m}\cdot\nabla)\vec{B}$. 
 %
% \begin{equation}
% \vec{F} = (\vec{m}\cdot\nabla)\vec{B}
% \end{equation}
%
A disc shaped robot has the peak magnetization in the Z-direction ($\vec{a_z}$), and hence the value $\vec{m}$ can be approximated as $\vec{m} = m_z\vec{a_z}$ and the force components in the X, Y, and Z-directions computed accordingly~\cite{Chow15a}.
% \begin{equation}
% \vec{m} = m_z\vec{a_z}
% \end{equation}

% The force equation can be expanded as
% \begin{equation}
% \vec{F} = m_z \frac{{\partial}}{{\partial}z}\begin{bmatrix}
% B_x \\ B_y\\B_z
% \end{bmatrix}
% \end{equation}
% \begin{equation}
% F_x = 0
% \end{equation}
% \begin{equation}
% F_y = m_z \frac{{\partial}B_y}{{\partial}z}
% \end{equation}
% \begin{equation}
% F_z = m_z \frac{{\partial}B_z}{{\partial}z}
% \end{equation}

\begin{figure}
\centering
\includegraphics[width=1\columnwidth]{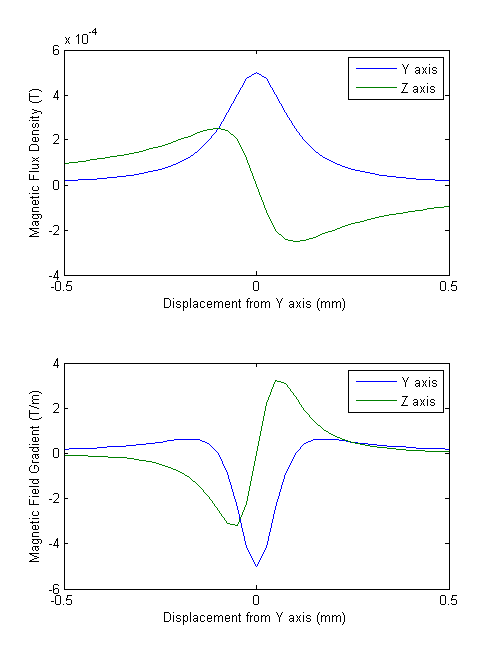}
\vspace{-0.20in}
\caption{Magnetic field density and magnetic flux gradient on a magnetic microrobot in the Y and Z-directions as a function of distance along the Y-axis from an infinitely long wire. Current is flowing in the positive X-direction.}% for constant Z height on the YZ plane.}
\label{fig:force_plots}
\vspace{-0.10in}
\end{figure}

\begin{figure}
\centering
\includegraphics[width=1\columnwidth]{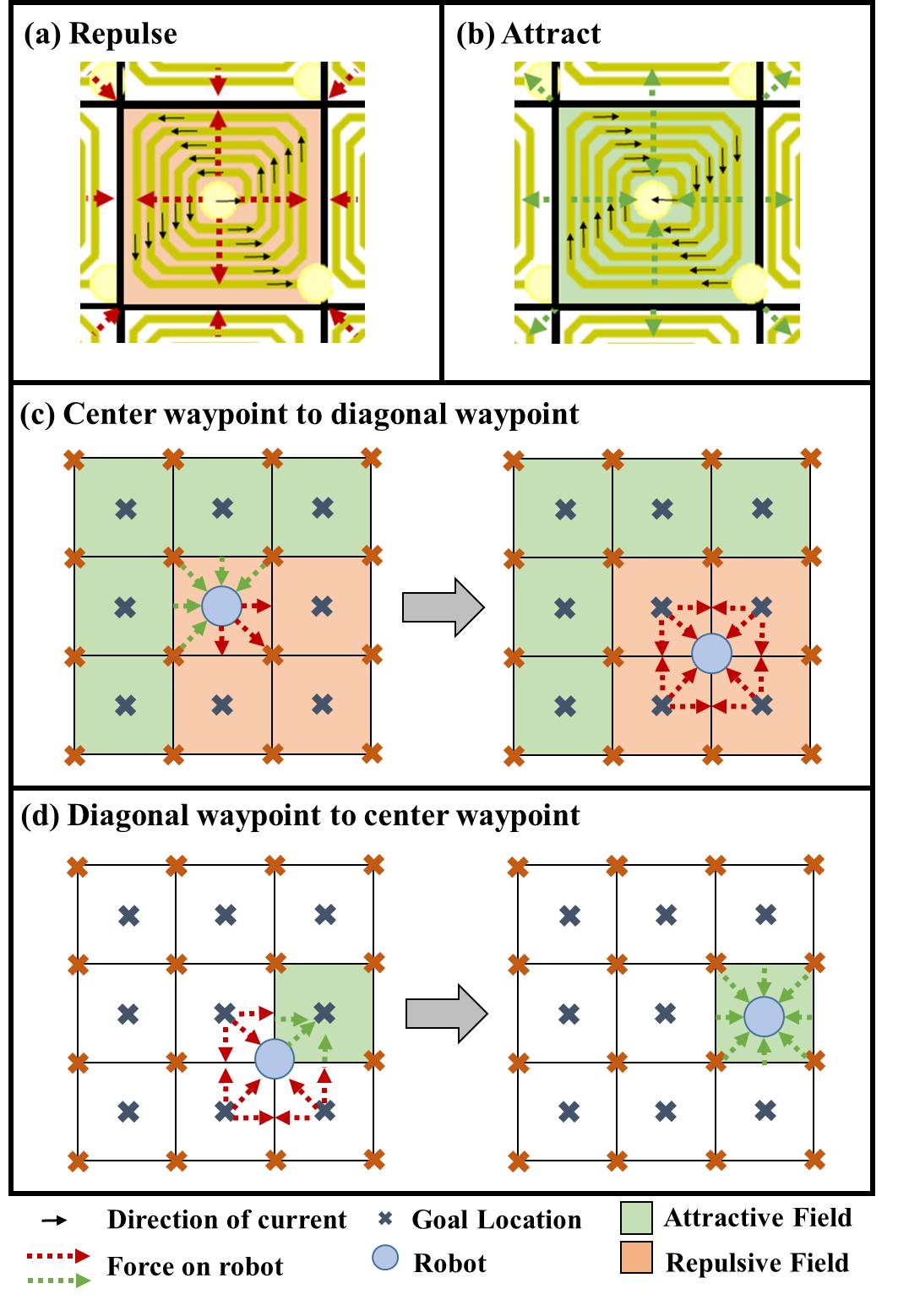}
\caption{Local magnetic field actuation. (a) Repulsive force due to clockwise wire current. (b) Attractive force due to anti-clockwise current in the planar coil. (c) Coil actuation states needed to actuate a microrobot to a diagonal waypoint from a center waypoint; (d) States needed to actuate a microrobot to a center waypoint from a diagonal waypoint.
}
\label{fig:LocalFields}
\vspace{-0.20in}
\end{figure}

For a constant height at which the field gradient is measured, the magnetic flux densities and gradients
%\textcolor{red}{(plots show flux density values, not forces.)} 
experienced by the magnetic robot along the Y-axis in the YZ plane are shown in Fig.~\ref{fig:force_plots}. Therefore, for a current flowing in the positive X-axis direction, the magnetic robot experiences a force in the negative Y-axis due to negative magnetic gradients. Thus, for a series of wires laid parallel to the X-axis, with currents all flowing in the same direction, the magnetic robot will translate in the Y-direction.
The currents in a spiral coil then will either repulse, force the robot outward, or attract, pull the robot towards its center, depending on the direction of current flowing through the coil. The net forces on a robot for a fixed direction of current are shown in Fig.~\ref{fig:LocalFields}(a) and (b). Each coil acts like an electromagnet. When the robot has attractive potential, it is like the coil having a magnetic moment in the same direction as that of the robot.  This will attract the robot if it's inside the bounds of the coil. Meanwhile, if the robot is out of the bounds of the coil, the coil and robot will repel each other like two magnets aligned parallel in the workspace. %This effect is only found when the robots are close to the coil, but not overlapping. The regions of the robot just outside its body exhibit magnetic moment field lines in the opposite direction. Therefore, when the body of the robot is just outside the border of the coil, the magnetic moment generated by the coil would repel the robot.
Therefore, a repulsive potential generated by the coil will push the magnet to the outer border of the coil. %For example, an attractive coil current attracts the robot when it's body is inside the coil borders and repulses the robot when its body is outside the borders of the coil.

To define waypoints (goal locations) for the robot, it is important that they are at equilibrium points, i.e points at lowest potential. In our previous work, it was believed that these equilibrium points only existed at the center of each coil, limiting the waypoints for path planning and trajectories to orthogonal moves in the workspace.  Here we have identified an additional equilibrium point at the corner positions of each coil (diagonal positions in the workspace).
%This occurs only at the center of a coil and at it's corner positions (diagonal positions in the workspace). 
The robot remains in the center of the coil if the coil ``attracts" the robot, and it remains in the diagonal (corner) position if all the surrounding coils ``repel" the robot.  However, to reliably move from a center waypoint to a diagonal waypoint, all nine neighboring coils of the robot are needed, as shown in Fig.\ref{fig:LocalFields}(c).  The  
%For diagonal goals, the robot has to be at the center location of the nearest coil. 
repulsing coils push the robot to the diagonal goal while the outer surrounding attracting coils help direct the robot towards the diagonal. Since the robot is outside the borders of these attracting coils, they actually repulse the robot.
%The surrounding attract coil here repulses the robot because the robot is outside its border. 
This makes sure that the robot moves in a predictable manner to the diagonal waypoint.  In order to move from a diagonal waypoint to a center waypoint, the robot simply has to reside in the influence radius, $R_{I_{coil}}$, of the corresponding coil.  It will then get pulled to the the static equilibrium at the center of the attracting coil, as shown in ~(Fig.~\ref{fig:LocalFields}(d)).  All of these possible center and diagonal equilibrium points in the workspace yield the potential waypoints for path planning. 
% \begin{figure*}
% \centering
% \includegraphics[width=1\textwidth]{ben_figures/coil_map_all.png}
% \caption{(a) Goal locations in the coil. (b) Robot moving to the diagonal position. (c) Robot moving to the center position}
% \label{fig:goal_map}
% \end{figure*}

%% file: ProblemDefinition.tex
Consider a team of $N$ microrobots residing in a magnetic workspace with $R$ waypoints/locations described in Sec.~\ref{sec:LocalMagneticFieldActuation}. The positions of the microrobots and the waypoints are denoted by ${\bf{x}}_i(t)$ and ${\bf{c}}_j$, respectively, where $i\in\{1,2,\dots,N\}$ and $j\in\{1,2,\dots,R\}$. 

\subsection{Discrete Abstraction of Robot Mobility}
\begin{figure}
\centering
\includegraphics[width=2.25in]{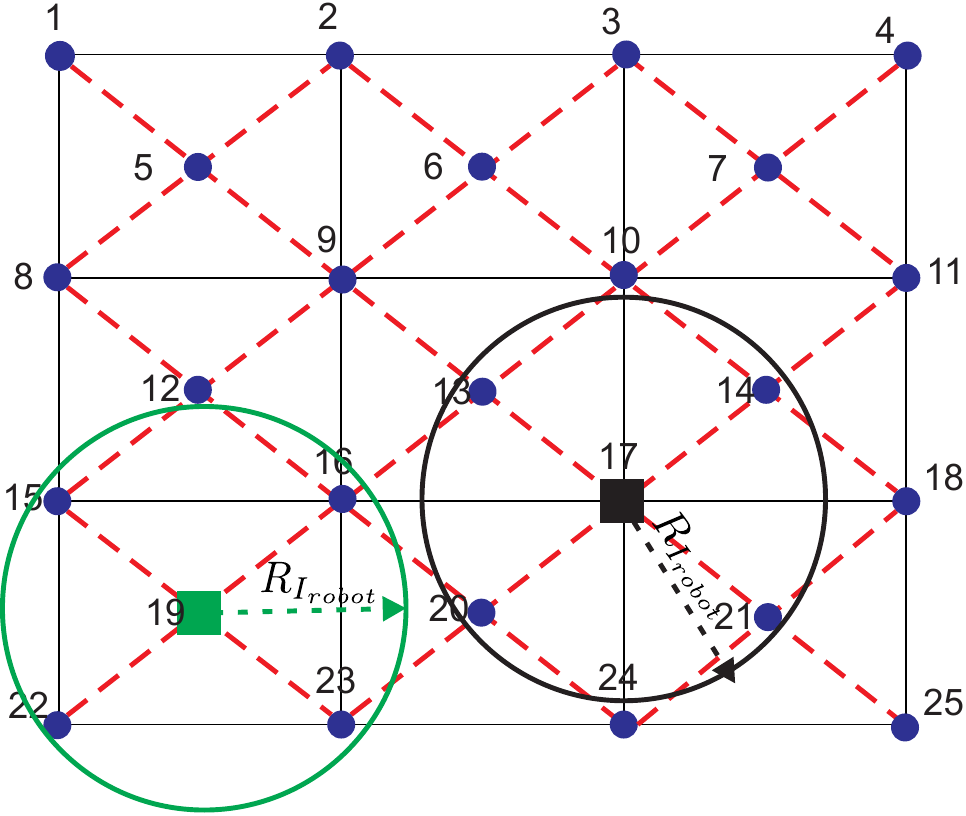}
\caption{Graphical depiction of a wTS as per Definition \ref{def:wts}. Blue dots denote the states of the wTS and red dashed lines depict the available transitions between the states. The red and black square denote two robots that have to accomplish a temporal logic task while maintaining a distance between them that is always greater than $R_{I_{robot}}$.}
\label{fig:concept}
\end{figure}

Robot mobility in the workspace can be represented by a weighted Transition System (wTS) defined as follows \cite{baier2008principles}; see also Figure \ref{fig:concept}.

\begin{defn}[weighted Transition System]\label{def:wts}
A \textit{weighted Transition System} $\text{wTS}_i$ associated with robot $i$ is a tuple $\left(\mathcal{Q}_i, q_i^0,\mathcal{A}_i,\rightarrow_i, w_i,\mathcal{AP}_i,L_{i}\right)$ where: 
\begin{itemize}
\item $\mathcal{Q}_{i}=\{q_i^{{\bf{c}}_j}\}_{j=1}^{R}$ is the set of states. A state $q_i^{{\bf{c}}_j}\in\mathcal{Q}_i$ is associated with the presence of robot $i$ in location ${\bf{c}}_j$.
\item $q_{i}^0\in\mathcal{Q}_{i}$ is the initial state of robot $i$; 
\item $\mathcal{A}_{i}$ is a set of actions; The available actions at state $q_i^{{\bf{c}}_j}\in\mathcal{Q}_i$ are move diagonally: up and left, up and right, down and left, down and right.
\item $\rightarrow_{i}\subseteq\mathcal{Q}_{i}\times\mathcal{A}_{i}\times\mathcal{Q}_{i}$ is the transition relation. Transition from state $q_i^{{\bf{c}}_j}$ to $q_i^{{\bf{c}}_k}$ exists if there is an action $a_i\in\mathcal{A}_i$ at state $q_i^{{\bf{c}}_j}$ that can drive the robot $i$ to state $q_i^{{\bf{c}}_k}$. 
\item $w_{i}:\mathcal{Q}_{i}\times\mathcal{Q}_{i}\rightarrow \mathbb{R}_+$ is a cost function that assigns weights/cost to each possible transition in the wTS. For example, these costs can be associated with the distance between two states $q_i^{{\bf{c}}_j}$ and $q_i^{{\bf{c}}_k}$; 
\item $\mathcal{AP}_i=\big\{\pi_i^{{\bf{c}}_j}\big\}_{j=1}^R$ is the set of atomic propositions, 
where the atomic proposition $\pi_i^{{\bf{c}}_j}$ is true if $\lVert {\bf{x}}_i - {\bf{c}}_j \rVert]\leq \epsilon$, for a sufficiently small $\epsilon>0$ and false, otherwise,
%where the atomic proposition $\pi_i^{{\bf{c}}_j}$ is true if robot $i$ is in location ${\bf{c}}_j$ and false otherwise,
and 
\item $L_{i}:\mathcal{Q}_{i}\rightarrow \mathcal{AP}_i$ is an observation/output function defined as $L_i(q_i^{{\bf{c}}_j})=\pi_i^{{\bf{c}}_j}$, for all $j\in\{1,\dots,W\}$.
\end{itemize}
\end{defn} 

Next, given $\text{wTS}_i$ for all robots, the \textit{Product Transition System} (PTS) is constructed, which captures all the possible combinations of robots' states in their respective $\text{wTS}_{i}$ and is defined as follows \cite{baier2008principles}.

\begin{defn}[Product Transition System]
Given $N$ transition systems $\text{wTS}_{i}=\big(\mathcal{Q}_{i}, q_{i}^0,\mathcal{A}_{i},$ $\rightarrow_{i},w_i,\mathcal{AP}_i,L_{i}\big)$, the \textit{weighted Product Transition System} $\text{PTS}=\text{wTS}_{1}\otimes\text{wTS}_{2}\otimes\dots\otimes\text{wTS}_{N}$ is a tuple $\left(\mathcal{Q}_{\text{PTS}},q_{\text{PTS}}^0,\mathcal{A}_{\text{PTS}},\longrightarrow_{\text{PTS}},\mathcal{AP},L_{\text{PTS}}\right)$ where: 
\begin{itemize}
\item $\mathcal{Q}_{\text{PTS}}=\mathcal{Q}_{1}\times\mathcal{Q}_{2}\times\dots\times\mathcal{Q}_{N}$ is the set of states, 
\item $q_{\text{PTS}}^0=(q_{1}^0,q_{2}^0,\dots,q_{N}^0)\in\mathcal{Q}_{\text{PTS}}$ is the initial state,
\item $\mathcal{A}_{\text{PTS}}=\mathcal{A}_{1}\times\mathcal{A}_{2}\times\dots\times\mathcal{A}_{N}$ is a set of actions,
\item
$\longrightarrow_{\text{PTS}}\subseteq\mathcal{Q}_{\text{PTS}}\times\mathcal{A}_{\text{PTS}}\times\mathcal{Q}_{\text{PTS}}$ is the transition relation defined by the rule\footnote{The notation here is adopted from \cite{baier2008principles}. It means that if the proposition above the solid line is true, then so is the proposition below the solid line. The state $q_{\text{PTS}}$ stands for $\left(q_{1},\dots,q_{N}\right)\in\mathcal{Q}_{\text{PTS}}$, where with slight abuse of notation $q_i\in\mathcal{Q}_i$. Also, $q_{\text{PTS}}'$ is defined accordingly.} $\frac{\bigwedge _{\forall i}\big(q_{i}\xrightarrow{a_{i}}_{i}q_{i}'\big)}{q_{\text{PTS}}\xrightarrow{a_{\text{PTS}}=(a_{1},\dots,a_{ij_{N}})}_{\text{PTS}}q_{\text{PTS}}'}$,
\item $w_{\text{PTS}}(q_{\text{PTS}},q_{\text{PTS}}')=w(q_{\text{PTS}},q_{\text{PTS}}')=\sum_{i=1}^{N} w_i(q_i,q_i')$, 
%see also \eqref{eq:w};  
\item $\mathcal{AP}=\bigcup_{i=1}^{N}\mathcal{AP}_i$ is the set of atomic propositions; and 
\item $L_{\text{PTS}}=\bigcup_{i}L_{i}$ is an observation/output relation giving the set of atomic propositions that are satisfied at a state. 
\end{itemize}
\label{def:pts}
\end{defn} 

In what follows, we give definitions related to the $\text{PTS}$, that we will use throughout the rest of the paper. An \textit{infinite path} $\tau$ of a $\text{PTS}$ is an infinite sequence of states, $\tau=\tau(1)\tau(2)\tau(3)\dots$ such that $\tau(1)=q_{\text{PTS}}^0$, $\tau(k)\in\mathcal{Q}_{\text{PTS}}$, and $(\tau(k),\tau_{i}(k+1))\in\rightarrow_{\text{PTS}}$, $\forall k\in\mathbb{N}_+$, where $k$ is an index that points to the $k$-th entry of $\tau$ denoted by $\tau(k)$. A \textit{finite path} of a $\text{PTS}$ can be defined accordingly. The only difference with the infinite path is that a finite path is defined as a finite sequence of states of a $\text{PTS}$. Given the definition of the weights $w_{\text{PTS}}$ in Definition \ref{def:pts}, the \textit{cost} of a finite path $\tau$, denoted by $J_f(\tau)$, can be defined as
\begin{equation}\label{eq:cost}
J_f(\tau)=\sum_{k=1}^{|\tau|-1}w_{\text{PTS}}(\tau(k),\tau(k+1)).
\end{equation} 
In \eqref{eq:cost}, $|\tau|$ stands for the number of states in $\tau$. In words, the cost \eqref{eq:cost} captures the total cost incurred by all robots during the execution of the finite path $\tau$. The \textit{trace} of an infinite path $\tau=\tau(1)\tau(2)\tau(3)\dots$ of a PTS, denoted by $\texttt{trace}(\tau)\in\left(2^{\mathcal{AP}}\right)^{\omega}$, where $\omega$ denotes infinite repetition, is an infinite word that is determined by the sequence of atomic propositions that are true in the states along $\tau$, i.e., $\texttt{trace}(\tau)=L(\tau(1))L(\tau(2))\dots$. 
The trace of a transition system $\text{PTS}$ is defined as $\texttt{trace}(\text{PTS})=\bigcup_{\tau\in \mathcal{P}}\texttt{trace}(\tau_{i})$, where $\mathcal{P}$ is the set of all infinite paths $\tau$ of $\text{PTS}$. 

\subsection{Task Specification}
In what follows, we assume that the microrobot team has to accomplish a complex collaborative task encapsulated by a global LTL statement $\phi$ defined over the set of atomic propositions $\mathcal{AP}=\left\{\{\pi_{i}^{{\bf{c}}_j}\}_{j=1}^{R}\right\}_{i=1}^N$. 
The basic ingredients of an LTL formula are a set of atomic propositions $\mathcal{AP}$, the boolean operators, i.e., conjunction $\wedge$, and negation $\neg$, and two temporal operators, next $\bigcirc$ and until $\mathcal{U}$. For the sake of brevity we abstain from presenting the derivations of other Boolean and temporal operators, e.g., \textit{always} $\square$, \textit{eventually} $\lozenge$, \textit{implication} $\Rightarrow$, which can be found in  \cite{baier2008principles}. For instance, consider the following temporal task:

\begin{align}\label{eq:example1}
\phi_{\text{task}}=&(\Diamond \pi_1^{{\bf{c}}_{30}})\wedge(\pi_1^{{\bf{c}}_{30}} \rightarrow\Diamond\pi_1^{{\bf{c}}_{54}}) \wedge (\pi_1^{{\bf{c}}_{54}} \rightarrow\Diamond\pi_2^{{\bf{c}}_{30}})\nonumber\\& \wedge( \pi_2^{{\bf{c}}_{30}} \rightarrow\Diamond(\pi_2^{{\bf{c}}_{28}}\wedge(\Diamond\pi_3^{{\bf{c}}_{28}})))
\end{align}
where $\pi_i^{{\bf{c}}_j}$ is an atomic proposition that is true only if robot $i$ is located at  ${\bf{c}}_j$ of the magnetic workspace. In words, the task described in \eqref{eq:example1} requires robot 1 to move part 1 of an object to location ${\bf{c}}_{30}$. Once this happens, robot 1 moves back to location ${\bf{c}}_{54}$, Then robot 2, moves to location ${\bf{c}}_{30}$ to take over the task. Then, robot 2 moves part 1 of this object to location ${\bf{c}}_{54}$ and then eventually robot 3 that carries part 2 of the considered object moves to location ${\bf{c}}_{28}$ to finalize the assembly task. %The robot trajectories for this assembly scenario are illustrated in Figure \ref{fig:coil}.

Also, we require that as the microrobots move to accomplish the assigned temporal task, the distance between them is always greater than $R_{I_{robot}} > 0$, where $R_{I{robot}}$ is the radius of influence of a robot with respect to another in the same workspace; see also Figure \ref{fig:concept}. Note that $R_{I{robot}} > R_{I{coil}}$. Robots closer together than $R_{I{robot}}$ will attract or repel each other. This way, we ensure that the magnetic fields used for mobility of robot $i$ do not induce movement of any robot $j\neq i$. This requirement can encapsulated by the following LTL statement.

\begin{equation}\label{eq:collision}
\phi_c=\square\neg_{\forall i,j}( \left\|{\bf{x}}_i(t)-{\bf{x}}_j(t)\right\|\leq R_{I{robot}}) 
\end{equation}

Consequently, the robots need to navigate the magnetic workspace so that the global LTL statement 
\begin{equation}\label{eq:phi}
\phi=\phi_{\text{task}}\wedge\phi_c
\end{equation}
is satisfied.

\subsection{Cost of Infinite Paths $\tau\models\phi$}
Given an LTL formula $\phi$, we define the \textit{language} $\texttt{Words}(\phi)=\left\{\sigma\in (2^{\mathcal{AP}})^{\omega}|\sigma\models\phi\right\}$, where $\models\subseteq (2^{\mathcal{AP}})^{\omega}\times\phi$ is the satisfaction relation, as the set of infinite words $\sigma\in (2^{\mathcal{AP}})^{\omega}$ that satisfy the LTL formula $\phi$. Given such a global LTL formula $\phi$ and an infinite path $\tau$ of a $\text{PTS}$ satisfies $\phi$ if and only if $\texttt{trace}(\tau)\in\texttt{Words}(\phi)$, which is equivalently denoted by $\tau\models\phi$. 

In what follows, we describe how an infinite path $\tau$ that satisfies an LTL formula $\phi$ can be written in a finite representation and then, we define its cost.
First, we translate the LTL	formula $\phi$ defined over a set of atomic propositions $\mathcal{AP}$ into a Nondeterministic B$\ddot{\text{u}}$chi Automaton (NBA) that is defined as follows \cite{vardi1986automata,baier2008principles}.
\begin{defn}
A \textit{Nondeterministic B$\ddot{\text{u}}$chi Automaton} (NBA) $B$ over $2^{\mathcal{AP}}$ is defined as a tuple $B=\left(\mathcal{Q}_{B}, \mathcal{Q}_{B}^0,\Sigma,\rightarrow_B,\mathcal{Q}_B^F\right)$, where 
\begin{itemize}
\item $\mathcal{Q}_{B}$ is the set of states, \item $\mathcal{Q}_{B}^0\subseteq\mathcal{Q}_{B}$ is a set of initial states, 
\item $\Sigma=2^{\mathcal{AP}}$ is an alphabet, \item $\rightarrow_{B}\subseteq\mathcal{Q}_{B}\times \Sigma\times\mathcal{Q}_{B}$ is the transition relation, and 
\item $\mathcal{Q}_B^F\subseteq\mathcal{Q}_{B}$ is a set of accepting/final states. 
\end{itemize}
\end{defn}

Given the $\text{PTS}$ and the NBA $B$ that corresponds to the LTL $\phi$, we can now define the \textit{Product B$\ddot{\text{u}}$chi Automaton} (PBA) $P=\text{PTS}\otimes B$, as follows \cite{baier2008principles}.

\begin{defn}[Product B$\ddot{\text{u}}$chi Automaton]
Given the product transition system $\text{PTS}=\big(\mathcal{Q}_{\text{PTS}},q_{\text{PTS}}^0,\mathcal{A}_{\text{PTS}},\longrightarrow_{\text{PTS}},$ $\mathcal{AP},L_{\text{PTS}}\big)$ and the NBA $B=\left(\mathcal{Q}_{B}, \mathcal{Q}_{B}^0,2^{\mathcal{AP}},\rightarrow_{B},\mathcal{Q}_{B}^F\right)$, the \textit{Product B$\ddot{\text{u}}$chi Automaton} $P=\text{PTS}\otimes B$ is a tuple $\left(\mathcal{Q}_{P}, \mathcal{Q}_{P}^0,\longrightarrow_{P},w_P,\mathcal{Q}_{P}^F\right)$ where: 
\begin{itemize}
\item $\mathcal{Q}_{P}=\mathcal{Q}_{\text{PTS}}\times\mathcal{Q}_{B}$ is the set of states,
\item $\mathcal{Q}_{P}^0=q_{\text{PTS}}^0\times\mathcal{Q}_{B}^0$ is a set of initial states,
\item $\longrightarrow_{P}\subseteq\mathcal{Q}_{P}\times\mathcal{A}_{\text{PTS}}\times 2^{\mathcal{AP}}\times\mathcal{Q}_{P}$ is the transition relation defined by the rule $\frac{\left(q_{\text{PTS}}\xrightarrow{a_{\text{PTS}}}q_{\text{PTS}}'\right)\wedge\left( q_{B}\xrightarrow{L_{\text{PTS}}\left(q_{\text{PTS}}\right)}q_{B}'\right)}{q_{P}=\left(q_{\text{PTS}},q_{B}\right)\xrightarrow{a_{\text{PTS}}}_{P}q_{P}'=\left(q_{\text{PTS}}',q_{B}'\right)}$, 
\item $w_P(q_{\text{P}},q_{\text{P}}')=w_{\text{PTS}}(q_{\text{PTS}},q_{\text{PTS}}^{'})$,  where $q_{\text{P}}=(q_{\text{PTS}},q_B)$ and $q_{\text{P}}'=(q_{\text{PTS}}',q_B^{'})$, and 
\item $\mathcal{Q}_{P}^F=\mathcal{Q}_{\text{PTS}}\times\mathcal{Q}_{B}^F$ is a set of accepting/final states. 
\end{itemize}
\label{def:pba}
\end{defn}

Given the PBA, we can represent an infinite path $\tau\models\phi$ in a finite form. Specifically, any infinite path $\tau$ that satisfies an LTL formula $\phi$ can be written in a finite representation, called prefix-suffix structure, i.e., $\tau=\tau^{\text{pre}}[\tau^{\text{suf}}]^{\omega}$, where the prefix part $\tau^{\text{pre}}$ is executed only once followed by the indefinite execution of the suffix part $\tau^{\text{suf}}$ \cite{guo2013revising,guo2013reconfiguration,guo2015multi}. The prefix part $\tau^{\text{pre}}$ is the projection of a finite path of the PBA, i.e., a finite sequence of states of the PBA, denoted by $p^{\text{pre}}$, onto $\mathcal{Q}_{\text{PTS}}$, which has the following structure
\begin{align}
p^{\text{pre}}=(q_{\text{PTS}}^0,q_B^0)(q_{\text{PTS}}^1,q_B^1)\dots (q_{\text{PTS}}^K,q_B^K),
\end{align}
with $(q_{\text{PTS}}^K,q_B^K)\in\mathcal{Q}_B^F$. The suffix part $\tau^{\text{suf}}$ is the projection of a finite path of the PBA, denoted by $p^{\text{suf}}$, onto $\mathcal{Q}_{\text{PTS}}$, which has the following structure 
\begin{align}
p^{\text{suf}}=&(q_{\text{PTS}}^{K},q_B^K)(q_{\text{PTS}}^{K+1},q_B^{K+1})\dots (q_{\text{PTS}}^{K+S},q_B^{K+S})\nonumber\\&(q_{\text{PTS}}^{K+S+1},q_B^{K+S+1}),
\end{align}
where $(q_{\text{PTS}}^{K+S+1},q_B^{K+S+1})=(q_{\text{PTS}}^{K},q_B^{K})$. Then our goal is to compute a plan
\begin{equation}
\tau=\tau^{\text{pre}}[\tau^{\text{suf}}]^{\omega}=\Pi|_{\text{PTS}}p^{\text{pre}}[\Pi|_{\text{PTS}}p^{\text{pre}}]^{\omega}, 
\end{equation}
where $\Pi|_{\text{PTS}}$ stands for the projection on the state-space $\mathcal{Q}_{\text{PTS}}$, so that the LTL formula $\phi$ in \eqref{eq:phi} is satisfied and the following objective function is minimized
\begin{align}\label{eq:cost2}
J(\tau)=J_f(\tau^{\text{pre}})+J_f(\tau^{\text{suf}}),
\end{align}
which captures the total cost incurred by all robots during the execution of the prefix and a single execution of the suffix part. In \eqref{eq:cost2}, $J_f(\tau^{\text{pre}})$ and $J_f(\tau^{\text{suf}})$ stands for the cost of the prefix and suffix part, where the cost function $J_f(\cdot)$ is defined in \eqref{eq:cost}, i.e., $J_f(\tau^{\text{pre}})=\sum_{k=1}^{K-1}w_{\text{PTS}}(\Pi|_{\text{PTS}}p^{\text{pre}}(k),\Pi|_{\text{PTS}}p^{\text{pre}}(k+1)),~
J_f(\tau^{\text{suf}})=\sum_{k=K}^{K+S}w_{\text{PTS}}(\Pi|_{\text{PTS}}p^{\text{suf}}(k)\Pi|_{\text{PTS}}p^{\text{suf}}(k+1))$.

The problem that we address in this paper is summarized as follows:

\textit{Problem:}
Determine a motion plan $\tau$ for the team of microrobots such that $\phi$ in \eqref{eq:phi} is accomplished, i.e., the assigned temporal task captured by $\phi_{\text{task}}$ is accomplished and independent mobility of all robots is ensured as per $\phi_c$, while minimizing the total cost \eqref{eq:cost2}.

% \begin{problem}\label{pr:pr1}
% Determine a motion plan $\tau$ for the team of microrobots such that $\phi$ in \eqref{eq:phi} is accomplished, i.e., the assigned temporal task captured by $\phi_{\text{task}}$ is accomplished and independent mobility of all robots is ensured as per $\phi_c$, while minimizing the total cost \eqref{eq:cost2}.
% \end{problem}

%% file: PlanningApproach.tex
To solve the problem as defined, %\ref{pr:pr1}
known optimal control synthesis techniques can be employed that rely on graph search techniques applied to the product automaton defined in Definition~\ref{def:pba}. In \ref{sec:model_checking}, we briefly summarize such methods. Then, in Section \ref{sec_proposed_model_checking}, we develop a modification to this algorithm that allows us to construct weighted Transitions Systems (wTS) with smaller state-spaces that generate words that satisfy the LTL formula $\phi=\phi_{\text{task}}\wedge\phi_c$.

\subsection{Existing Optimal Control Synthesis Methods}\label{sec:model_checking}

The problem at hand is typically solved by applying graph-search methods to the PBA, see e.g., \cite{guo2013revising,guo2013reconfiguration,guo2015multi}. Specifically, to generate a motion plan $\tau$ that satisfies $\phi$ and minimizes the cost function \eqref{eq:cost2}, the PBA is viewed as a weighted directed graph $\mathcal{G}_P=\{\mathcal{V}_P, \mathcal{E}_P, w_P\}$, where the set of nodes $\mathcal{V}_P$ is indexed by the set of states $\mathcal{Q}_P$, the set of edges $\mathcal{E}_P$ is determined by the transition relation $\longrightarrow_P$, and the weights assigned to each edge are determined by the function $w_P$. Then, we find the shortest paths from the initial states to all reachable final states $q_P\in\mathcal{Q}_P^F$ and projecting these paths onto the $\text{PTS}$ results in the prefix parts $\tau^{\text{pre},f}$, where $f=\{1,\dots,|\mathcal{Q}_P^F|\}$. The respective suffix parts $\tau^{\text{suf},f}$ are constructed similarly by computing the shortest cycle around the $f$-th final state. All the resulting motion plans $\tau^f=\tau^{\text{pre},f}[\tau^{\text{suf},f}]^{\omega}$ satisfy the LTL specification $\phi$. Among all these plans, we can easily compute the optimal plan that minimizes the cost function defined in \eqref{eq:cost2} by computing the cost $J(\tau^f)$ for all plans and selecting the one with the smallest cost; see e.g. \cite{guo2013reconfiguration,guo2015multi,ulusoy2013optimality,ulusoy2014optimal}. 

\begin{figure}%[h] 
\centering
\subfloat[]{\label{fig:example_shortest_paths}\includegraphics[width=0.30\textwidth]{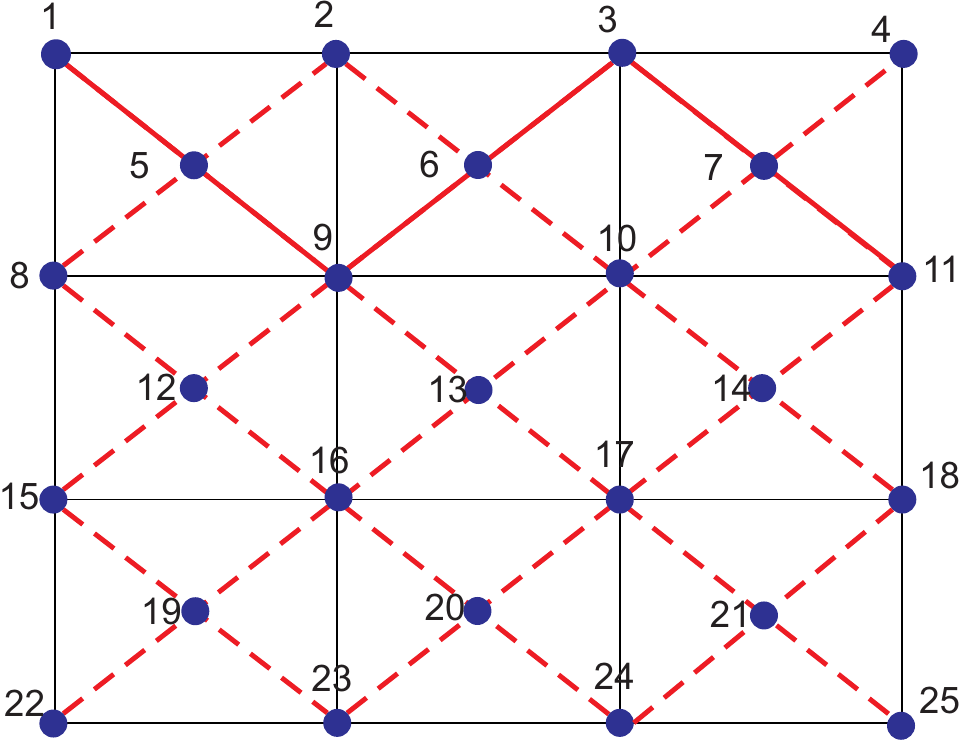}}\\
%\hspace*{0.2cm}
\subfloat[]{\label{fig:add_states}\includegraphics[width=0.30\textwidth]{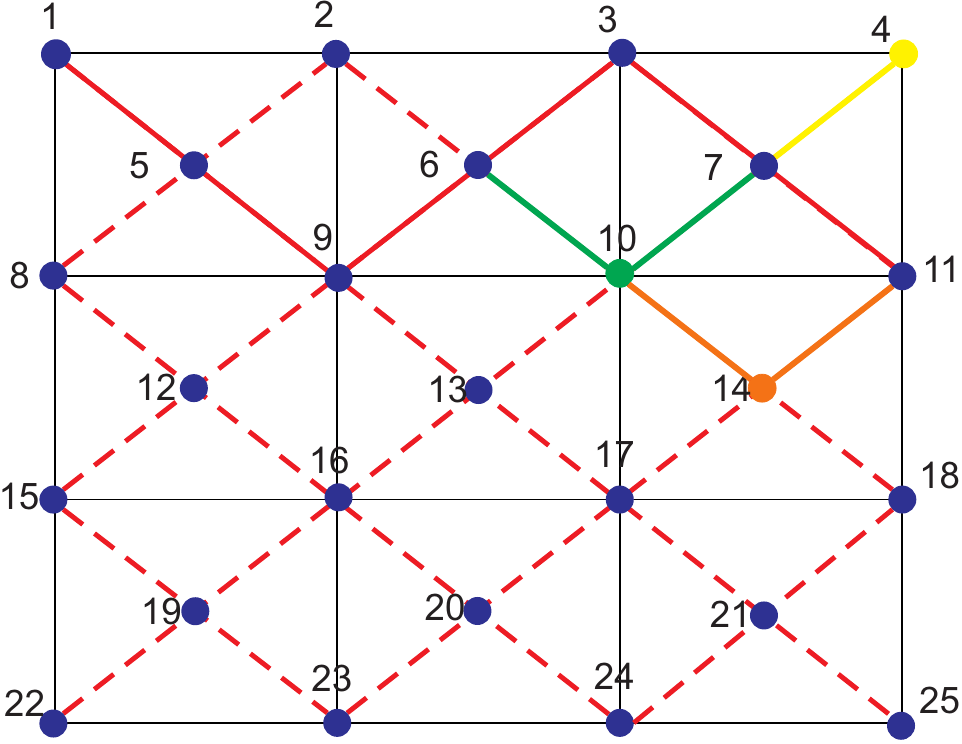}}
\caption[]{(a): Blue dots and red (solid and dashed) edges represent the states and all possible transitions among these states in $\text{wTS}_1$, respectively. Red solid edges represent the transitions among the states of $\text{wTS}_1^0$ when the global LTL formula is described by \eqref{eq:example}. (b): Yellow (green) [orange] dots and edges represent the states and  transitions that are included in $\text{wTS}_1^0$ at the first (second) [third] iteration of the while-loop of Algorithm \ref{alg:update}, respectively. Given the LTL expression \eqref{eq:example}, $\overline{\Pi}_1\cap L_1(\mathcal{Q}_1^0)=\{\pi_1^{{\bf{c}}_7}\}$ for robot 1 and, therefore, $\mathcal{S}_1=\{{\bf{c}}_7\}$.}
\label{fig:coil}
\end{figure}

\subsection{Proposed Optimal Control Synthesis  Approach}\label{sec_proposed_model_checking}
The optimal control synthesis algorithm discussed in Section \ref{sec:model_checking} relies on the construction of a synchronous product transition system among all robots in the network. As a result, it suffers from the state explosion problem and, therefore, it is resource demanding and scales poorly with the number of robots. Specifically, the worst case complexity of that algorithm is $\mathcal{O}((|\mathcal{E}_P|+|\mathcal{V}_P|\log(|\mathcal{V}_P|))(|\mathcal{Q}_{P}^0|+|\mathcal{Q}_{P}^F|))$, since it is based on executing the Dijkstra algorithm $|\mathcal{Q}_{P}^0|+|\mathcal{Q}_{P}^F|$ times. 

To decrease the complexity of this model checking algorithm we develop a novel method that aims to reduce the state-space of the product transition system and, consequently, the cardinality of the sets $\mathcal{E}_P$ and $\mathcal{V}_P$ by taking into account the atomic propositions that appear in the LTL expression $\phi$. Specifically, the algorithm checks at which regions these atomic propositions are satisfied and then  constructs paths towards those regions. This way, different LTL expressions will result in different wTS's. To achieve this, we construct wTS's, denoted by $\text{wTS}_i'$ so that $\text{wTS}_i=\left(\mathcal{Q}_i, q^0_i,\mathcal{A}_i,\rightarrow_i,w_i,\mathcal{AP},L_i\right)$, defined in Definition \ref{def:wts} and $\text{wTS}_i'=\left(\mathcal{Q}_i', q^0_i,\mathcal{A}_i,\rightarrow_i,w_i,\mathcal{AP},L_i\right)$ satisfy the following trace-inclusion property (see Chapter 3.2.4 in \cite{baier2008principles})
\begin{equation}\label{eq:traceIncl}
\texttt{trace}(\text{wTS}_{i}')\subseteq\texttt{trace}(\text{wTS}_{i}).
\end{equation}
%\begin{defn}[Trace-included wTS]\label{def:traceinc}  
%Consider weighted Transitions Systems $\text{wTS}_i=\left(\mathcal{Q}_i, q^0_i,\mathcal{A}_i,\rightarrow_i,w_i,\mathcal{AP},L_i\right)$ and $\widetilde{\text{wTS}}_i=\big(\tilde{\mathcal{Q}}_i, \tilde{q}^0_i,\tilde{\mathcal{A}}_i,\rightarrow_{\tilde{i}},\tilde{w}_i,\mathcal{AP},L_i\big)$ both defined over a set of atomic propositions $\mathcal{AP}$; $\widetilde{\text{wTS}}_i$ is trace-included by ${\text{wTS}}_i$ if $\texttt{trace}(\widetilde{\text{wTS}}_{i})\subseteq\texttt{trace}(\text{wTS}_{i})$.
%\end{defn}
%
In other words, our goal is to construct $\text{wTS}_i'$ that may not be as ``expressive'' as $\text{wTS}_i$, since $\texttt{trace}(\text{wTS}_{i}')\subseteq\texttt{trace}(\text{wTS}_{i})$, but  have  smaller state-spaces and are able to generate motion plans that satisfy the LTL formula $\phi$. Note that for $\text{wTS}_{i}'$ and $\text{wTS}_i$ it holds that $\mathcal{Q}_{i}'\subseteq\mathcal{Q}_{i}$ while both wTSs are defined over the same set of actions $\mathcal{A}_i$, the same transition rule $\rightarrow_i$, and the same function $w_i$. The wTSs $\text{wTS}_i'$ are constructed iteratively as per Algorithm \ref{alg:update} until they become expressive enough to satisfy the LTL formula $\phi$. Specifically, the wTS constructed at iteration $m$ of Algorithm \ref{alg:update} for robot $i$ is denoted by $\text{wTS}_{i,m}'=\left(\mathcal{Q}_{i,m}', q^0_i,\mathcal{A}_i,\rightarrow_i,w_i,\mathcal{AP},L_i\right)$.%The PBA based on the wTS constructed by Algorithm \ref{alg:update} at iteration $k\in\{0,1,\dots\}$ is denoted by $\text{wTS}_{i,k}'=\left(\mathcal{Q}_{i,k}', q^0_i,\mathcal{A}_i,\rightarrow_i,w_i,\mathcal{AP},L_i\right)$ the wTS constructed for robot $i$ at iteration $k$ of Algorithm \ref{alg:update}.

\begin{algorithm}[t]
\caption{Update $\text{wTS}_i$ }
\label{alg:update}
\begin{algorithmic}[1]
%\STATE \textcolor{blue}{$k=0$};
%\STATE \textcolor{blue}{Construct $\text{wTS}_{i}'$ and $\text{P}_0'=(\otimes_{\forall i}\text{wTS}_{i}')\otimes B$ };
\IF {$\overline{\Pi}_i\cap L_i(\mathcal{Q}_{i,0}')\neq\varnothing$}\label{upd:line1}
\STATE $\mathcal{S}_i=\{q_i\in\mathcal{Q}_{i,0}' | L_i(q_i)\in\overline{\Pi}_i\}$;\label{upd:line2}
\ELSE
\STATE $\mathcal{S}_i=\mathcal{Q}_{i,0}'$;\label{upd:line4}
\ENDIF
\STATE $m=0$, $n_i=1$, $\kappa_i=1$, $\forall i\in\{1,\dots,N\}$;\label{upd:line6}
%\STATE \textcolor{blue}{$\text{PTS}_k'=\otimes_{\forall i}\text{wTS}_{i,k}'$,$\text{P}_k'=\text{PTS}_k'\otimes B$}; \label{upd:line7}
\WHILE {$\texttt{trace}(\text{PTS}_m)\cap\mathcal{L}_B=\varnothing$} \label{upd:line8}
\STATE Pick a state from $f_i:\mathcal{N}_{\mathcal{S}_i(\kappa_i)}^{n_i-\text{hops}}\rightarrow (0,1]$;\label{upd:line9}
\STATE Construct the PTS state $q_{\text{PTS}}^{\text{new}}=(q_1^{\text{new}},\dots,q_N^{\text{new}})$; \label{upd:line10}
\STATE $P_{m+1}=\texttt{UpdatePBA}(P_{m},q_{\text{PTS}}^{\text{new}},\mathcal{Q}_B)$ (Alg. \ref{alg:updatePBA});\label{upd:line11}
\STATE $\kappa_i=\kappa_i+1$;\label{upd:line12}
\IF {$\kappa_i>|\mathcal{S}_i|$}\label{upd:line13}
\STATE $\kappa_i=1$, $n_i=n_i+1$;\label{upd:line14}
\ENDIF	
\STATE $m=m+1$;
\ENDWHILE
\end{algorithmic}
\end{algorithm}

\begin{algorithm}[t]
\caption{$P_{m+1}=\texttt{UpdatePBA}(P_m,q_{\text{PTS}}^{\text{new}},\mathcal{Q}_B)$}
\label{alg:updatePBA}
\begin{algorithmic}[1]
\STATE $P_{m+1}=P_{m}$;
\FOR{$q_P=\{q_\text{PTS},q_B\}\in\mathcal{Q}_{P,m}$}
\IF {$q_\text{PTS}\xrightarrow{a_{\text{PTS}}}_{\text{PTS}}q_{\text{PTS}}^{\text{new}}$}
\FOR{$q_B^{\text{new}}\in\mathcal{Q}_B$}
\IF{$q_B\xrightarrow{L_{\text{PTS}}(q_{\text{PTS}})}_{B}q_B^{\text{new}}$}
\STATE Add state $q_P^{\text{new}}=(q_{\text{PTS}}^{\text{new}},q_B)$ to $\mathcal{Q}_{P,m+1}'$;
\STATE Add transition $(q_P,q_P^{\text{new}})$ with cost $w_{\text{PTS}}(q_\text{PTS},q_\text{PTS}^{\text{new}})$;
\ENDIF
\STATE
\ENDFOR
\ENDIF
\IF {$q_\text{PTS}^{\text{new}}\xrightarrow{a_{\text{PTS}}}_{\text{PTS}}q_{\text{PTS}}$}
\FOR{$q_B^{\text{new}}\in\mathcal{Q}_B$}
\IF{$q_B^{\text{new}}\xrightarrow{L_{\text{PTS}}(q_{\text{PTS}}^{\text{new}})}_{B}q_B$}
\STATE $q_P^{\text{new}}=(q_{\text{PTS}}^{\text{new}},q_B)$ to $\mathcal{Q}_{P,k+1}'$, if it does not already exist;
\STATE Add transition $(q_P^{\text{new}},q_P)$ with cost $w_{\text{PTS}}(q_\text{PTS}^{\text{new}},q_\text{PTS})$;
\ENDIF
\ENDFOR
\ENDIF
\ENDFOR
\end{algorithmic}
\end{algorithm}

\paragraph{Initialization} First, we present the construction of initial wTS denoted by $\text{wTS}_{i,0}'=\left(\mathcal{Q}_{i,0}', q^0_i,\mathcal{A}_i,\rightarrow_i,w_i,\mathcal{AP},L_i\right)$ given a global LTL expression $\phi$ and original wTS $\text{wTS}_i$. Given an LTL formula $\phi$ we define the following sets of atomic propositions. Let $\Pi_i$ be an ordered set that collects all atomic propositions $\pi_i^{{\bf{c}}_e}$ associated with robot $i$ that appear in $\phi$ without the negation operator $\neg$ in front of them, including the atomic proposition that is true at $q_i^0$. Also, let $\overline{\Pi}_i$ be a set that collects all atomic propositions $\pi_i^{{\bf{c}}_e}$ associated with robot $i$ that appear in $\phi$ with the negation operator $\neg$ in front of them. If an atomic proposition appears in $\phi$ more than once, both with and without the negation operator, then it is included in both sets. For example, consider the following  $\phi$:
\vspace{-2mm}
\begin{align}\label{eq:example} 
\phi=&(\square\Diamond \pi_1^{{\bf{c}}_{11}})\wedge (\square\Diamond \pi_1^{{\bf{c}}_3}) \wedge (\square\Diamond \pi_1^{{\bf{c}}_7}) \wedge (\square \neg \pi_1^{{\bf{c}}_{2}})
\nonumber\\&\wedge ( \neg\pi_1^{{\bf{c}}_7} \mathcal{U} \pi_1^{{\bf{c}}_{11}})\wedge(\square\Diamond \pi_2^{{\bf{c}}_9}).
\end{align}
Then, $\Pi_1$ and $\overline{\Pi}_1$ become $\Pi_1=\{\pi_1^{{\bf{c}}_1},\pi_1^{{\bf{c}}_3},\pi_1^{{\bf{c}}_{7}}, \pi_1^{{\bf{c}}_{11}}\}$ and $\overline{\Pi}_1=\{\pi_1^{{\bf{c}}_{2}},\pi_1^{{\bf{c}}_{7}}\}$ where $\pi_i^{{\bf{c}}_1}$ denotes the atomic proposition that is satisfied at the initial state $q_i^0$ of robot $i$.

To construct the state-space $\mathcal{Q}_{i,0}'$ of $\text{wTS}_{i,0}'$, we view the corresponding original transition system $\text{wTS}_i$ as a weighted directed graph $G_i=\{\mathcal{V}_i,\mathcal{E}_i\}$, where the set of vertices $\mathcal{V}_i$ and the set of indices $\mathcal{E}_i$ are determined by $\mathcal{Q}_i$ and $\rightarrow_i$, respectively. Weights on edges are assigned by the function $w_i$. Then, we compute the shortest paths from the location where $\Pi_i(n)$ is true to the location where $\Pi_i(n+1)$ is true avoiding all locations associated with atomic propositions in $\bar{\Pi}_i\setminus\Pi_i$, for all $n\in\{1\dots,|\Pi_i|\}$; see Figure \ref{fig:example_shortest_paths}. All states of $\text{wTS}_i$ that appear in these shortest paths comprise the set $\mathcal{Q}_{i,0}'$.\footnote{The order in the sets $\Pi_i$ can be selected either randomly or by checking which order results in the smaller state-space $\mathcal{Q}_{i,0}$.} Given $\text{wTS}_{i,0}'$, we construct the corresponding PBA defined as $\text{P}_0=(\otimes_{\forall i}\text{wTS}_{i,0}')\otimes B$. 
%Note that $\text{wTS}_{i,0}'$ may not be expressive enough to satisfy $\phi$, i.e., a prefix-suffix plan that satisfies $\phi$ may not exist based on the corresponding PBA $\text{P}_0'$. In this case, 
Given the sets of states $\mathcal{Q}_{i,0}'$, we define the sets $\mathcal{S}_i\subseteq\mathcal{Q}_{i,0}'$ that collect states in $q_i\in\mathcal{Q}_{i,0}'$ that if visited by robot $i$, then the LTL formula might be violated. These sets will be used for the re-construction of the PBA at iteration $k\in\{1,\dots\}$ of Algorithm \ref{alg:update}. The sets $\mathcal{S}_i$ are defined as $\mathcal{S}_i=\{q_i\in\mathcal{Q}_{i,0}' | L_i(q_i)\in\overline{\Pi}_i\}$,  $\overline{\Pi}_i\cap L_i(\mathcal{Q}_{i,0}')\neq\varnothing$. Otherwise, $\mathcal{S}_i=\mathcal{Q}_{i,0}'$ [line 4].

\paragraph{Iterative Update} %\textcolor{blue}{
%If $\text{wTS}_{i,0}'$ are not expressive enough, i.e., a prefix-suffix plan that satisfies $\phi$ cannot be found based on the corresponding PBA $\text{P}_0'$, then we enrich the wTSs, by adding new states, and then we construct the respective PBA. This iterative process terminates when the resulting PBA can generate a prefix-suffix motion plan that satisfies $\phi$; see Algorithm \ref{alg:update}.}
%
%
The PBA $\text{P}_m=(\otimes_{\forall i}\text{wTS}_{i,m}')\otimes B$ constructed at iteration $m\in\{0,1,\dots\}$ of Algorithm \ref{alg:update} may or may not satisfy the LTL specification $\phi$. In the latter case all robots design new wTSs, denoted by $\text{wTS}_{i,m+1}$, with states-spaces $\mathcal{Q}_{i,m+1}'$ that are constructed by adding a new state to their respective $\mathcal{Q}_{i,m}'$, as per Algorithm \ref{alg:update}.
The candidate state to be added to $\mathcal{Q}_{i,m}'$ is selected from the neighborhood of the states that belong to an ordered set $\mathcal{S}_i$ since, intuitively, visitation of these states may lead to violation $\phi$ [line 2]. Specifically, to construct $\mathcal{Q}_{i,m+1}'$ a new state $q_i^{\text{new}}$ is added to $\mathcal{Q}_{i,m}'$ that is generated by a discrete probability density function $f_i:\mathcal{N}_{\mathcal{S}_i(\kappa_i)}^{n_i-\text{hops}}\rightarrow (0,1]$ that is bounded away from zero on the $\mathcal{N}_{\mathcal{S}_i(\kappa_i)}^{n_i-\text{hops}}$. The set $\mathcal{N}_{\mathcal{S}_i(\kappa_i)}^{n_i-\text{hops}}$ that contains the $n_i$-hops connected neighbors of the $\kappa_i$-th state in $\mathcal{S}_i$, denoted by $\mathcal{S}_i(\kappa_i)$, in the graph $G_i=\{\mathcal{V}_i,\mathcal{E}_i\}$ excluding the states that already belong to the state space of $\mathcal{Q}_{i,m}'$; see Figure \ref{fig:add_states}. 
%The state that will be included in $\tilde{\mathcal{Q}}_i$ can be selected from the set $\mathcal{N}_{\mathcal{S}_i(\kappa)}^{n-\text{hops}}$ based on various criteria. For example it can be selected randomly, or based on the degree of that state in the graph $G_i$, or based on the sum of weights of the transitions that will be included in $\rightarrow_{\tilde{i}}$. 

Once the states $q_i^{\text{new}}$ are generated and the state-spaces $\mathcal{Q}_{i,m+1}'$ are constructed, the corresponding PBA $\text{P}_{m+1}=(\otimes_{\forall i}\text{wTS}_{i,m+1}')\otimes B$ is constructed as well. Instead of reconstructing the PBA $\text{P}_{m+1}$ from scratch, which is a computationally expensive step, we construct it by updating $\text{P}_{m}$ as per Algorithm \ref{alg:updatePBA}. Specifically, $\text{P}_{m+1}$ is initialized as $\text{P}_{m+1}=\text{P}_{m}$ [line 1, Alg. \ref{alg:updatePBA}]. Then, we check  if there exists feasible transitions 
from states $q_P=(q_{\text{PTS}},q_B)\in\mathcal{Q}_{P,m}$ to $q_P^{\text{new}}=(q_{\text{PTS}}^{\text{new}},q_B^{\text{new}})$ that satisfy $q_P\rightarrow_P q_P^{\text{new}}$, where $q_{\text{PTS}}^{\text{new}}=(q_1^{\text{new}},\dots,q_N^{\text{new}})$, for all $q_B^{\text{new}}\in\mathcal{Q}_B$ [lines 2-11, Alg. \ref{alg:updatePBA}]. Similarly, next, we identify all possible transitions from $q_P^{\text{new}}=(q_{\text{PTS}}^{\text{new}},q_B^{\text{new}})$ to $q_P=(q_{\text{PTS}},q_B)\in\mathcal{Q}_{P,k}$, for all $q_B^{\text{new}}\in\mathcal{Q}_B$ [lines 12-20, Alg. \ref{alg:updatePBA}].
If for some $n_i$, all the states from $\mathcal{N}_{\mathcal{S}_i(\kappa_i)}^{n_i-\text{hops}}$ have been included then the index $n_i$ is increased by one and $\kappa_i$ is reset to one [line 14]. We then have the following result:
\begin{prop}[Completeness]\label{prop:correctness}
Assume that $\texttt{trace}(\text{PTS})\cap\texttt{Words}(\phi)\neq\varnothing$, i.e., that the initial transition systems $\text{wTS}_i$ can generate motions plans that satisfy $\phi$. Then, Algorithm \ref{alg:update} can construct wTS's that can generate feasible motion plans $\tau$ that satisfy $\phi$ after finite number of iterations $m$.
\end{prop}
%\begin{proof}
{\textit{Proof:}} The proof is based on the fact that at the worst case scenario, Algorithm \ref{alg:update} will incorporate all states from $\mathcal{Q}_i$ into $\mathcal{Q}_{i,m}'$, i.e., $\mathcal{Q}_i=\mathcal{Q}_{i,m}'$, for all robots $i$. Note that this will happen after a finite number of iterations of Algorithm \ref{alg:update}, since $\mathcal{Q}_i$ is a finite set of states, by assumption, and $\mathcal{Q}_{i,m+1}'\subseteq \mathcal{Q}_i$ contains exactly one state that does not belong to $\mathcal{Q}_{i,m}'\subseteq\mathcal{Q}_i$, for all robots $i$ and iterations $m$, by construction of the sets $\mathcal{N}_{\mathcal{S}_i}^{n_i-\text{hops}}$. Therefore, there exists a finite iteration $m$ of Algorithm \ref{alg:update}, when it holds that $\texttt{trace}(\text{PTS}_{m})=\texttt{trace}(\text{PTS})$, where $\text{PTS}_{m}=\otimes_{i}\text{wTS}_{i,m}'$. Consequently, we have that $\texttt{trace}(\text{PTS}_{m})\cap\texttt{Words}(\phi)\neq\varnothing$ completing the proof.
%\end{proof}
%, since by assumption we have $\texttt{trace}(\text{PTS})\cap\texttt{Words}(\phi)\neq\varnothing$

\begin{rem}[Optimality]
Let $J_{\text{PTS}}^*$ denote the optimal cost when the motion plan $\tau$ is computed over $\text{PTS}=\otimes_{\forall i}\text{wTS}_i$. Given that $\mathcal{Q}_{i,m}'\subseteq\mathcal{Q}_i$, for all robots $i$ and for all $m\in\{0,1\dots\}$, it holds that $\mathcal{Q}_{\text{PTS},m}\subseteq\mathcal{Q}_{\text{PTS}}$, $\mathcal{Q}_{\text{PTS},m}$ is the state-space of $\text{PTS}_{m}=\otimes_{i}\text{wTS}_{i,m}'$. Therefore, in general, for the optimal cost computed over $\text{PTS}_{m}$ we have that $J_{\text{PTS}_{k}}^*\geq J_{\text{PTS}}^*$. 
\end{rem}

\subsection{Complexity Analysis}
In this section, we discuss the computational complexity of Algorithm \ref{alg:update}. At every iteration $m$ of Algorithm \ref{alg:update}, we check if prefix-suffix plans that satisfy the assigned LTL specification can be generated based on the corresponding PBA $P_{m}$. To check that, we compute the shortest paths from the initial states to the final states of the PBA (prefix parts) and the shortest cycles around these final states (suffix parts). Viewing the constructed PBA $P_{m}$ as a graph with set of nodes and edges denoted by $\mathcal{V}_{P,m}$ and $\mathcal{E}_{P,m}$, respectively, we get that the computational complexity of this process is $O((|\mathcal{E}_{P,m}|+|\mathcal{V}_{P,m}|\log(|\mathcal{V}_{P,m}|))(|\mathcal{Q}_{P,m}^0|+|\mathcal{Q}_{P,m}^F|))$, since it is based on executing the Dijkstra algorithm $|\mathcal{Q}_{P,m}^0|+|\mathcal{Q}_{P,m}^F|$ times. 
If a feasible plan cannot be generated by $P_{m}$, then a new state $q_{\text{PTS}}^{\text{new}}$ is sampled and added to the wTSs and the corresponding PBA is constructed by Algorithm \ref{alg:updatePBA}. The complexity of Algorithm \ref{alg:updatePBA} is $O(|\mathcal{V}_{P,m}N\mathcal{Q}_B|)$. Therefore, the computational complexity per iteration $m$ of Algorithm \ref{alg:updatePBA} is $O([|\mathcal{V}_{P,m}\mathcal{Q}_B|+(|\mathcal{E}_{P,m}|+|\mathcal{V}_{P,m}|\log(|\mathcal{V}_{P,m}|))(|\mathcal{Q}_{P,m}^0|+|\mathcal{Q}_{P,k}^F|)])$. Observe that the computational complexity of the Algorithm 1 increases with respect to iterations $m$ since the size of $\mathcal{V}_{P,m}$ and $\mathcal{E}_{P,m}$ increases.

Finally, viewing the PBA as a graph the memory required to store it using its adjacency list is $O(|\mathcal{V}_{P,m}|+|\mathcal{E}_{P,m}|)$. Note that this is much smaller than the memory $O(|\mathcal{V}_{P,m}|+|\mathcal{E}_{P,m}|)$ required to store the original PBA, since $\mathcal{V}_{P,m}\subseteq\mathcal{V}_P$ and $\mathcal{E}_{P,m}\subseteq\mathcal{E}_P$, for all $m\in\{0,1,\dots\}$ by construction. This allows us to solve larger planning problems that the existing approaches cannot manipulate due to memory requirements. 

\begin{rem}[Construction of wTS]
Note that there may exist cases where the resulting transition systems, constructed by Algorithm 1, are the same as the original ones. For example, this can happen if $\Pi_i=L_i(\mathcal{Q}_i)$, for all robots $i$, i.e., if all states in the original transition systems are associated with atomic propositions that appear in the assigned task $\phi$, since then $\mathcal{Q}_{i,0}'=\mathcal{Q}_i$. Also, if an infeasible LTL task is assigned to the robots then eventually our algorithm will eventually construct wTSs with $\mathcal{Q}_{i,m}'=\mathcal{Q}_i$, for all robots $i$. In these cases, our algorithm cannot reduce the computational cost of synthesizing motion plans.
\end{rem}

% \begin{figure*}[h]
% \centering
% \subfloat[]{\includegraphics[width=0.30\linewidth]{prefix_e8.eps}\label{p8}}
% \subfloat[]{\includegraphics[width=0.30\linewidth]{prefix_e10.eps}\label{p10}}
% \subfloat[]{\includegraphics[width=0.30\linewidth]{prefix_e20.eps}\label{p20}}
% \caption{Case Study I: Robot trajectories due to the execution of the prefix part $\tau^{\text{pre}}$. (a), (b), and (c) illustrate the states $\tau^{\text{pre}}(k)$ of the prefix part, for $k\in\{1,\dots,8\}$, $k\in\{8,9,10\}$, and $k\in\{10,\dots,20\}$. The projection of the states $\tau^{\text{pre}}(8)$, $\tau^{\text{pre}}(10)$, and $\tau^{\text{pre}}(20)$ onto the trace-included wTS of the red and blue robot are depicted by red and blue disks, respectively. The red and blue circles illustrate the part of the workspace that falls within the influence radius of the red and blue robot in these states.} %when their states are determined by $\tau^{\text{pre}}(8)$, $\tau^{\text{pre}}(10)$, and $\tau^{\text{pre}}(20)$.}
% \label{sim1pre}
% \end{figure*}

\begin{figure}%[h]
\centering
\subfloat[]{\includegraphics[width=0.62\linewidth]{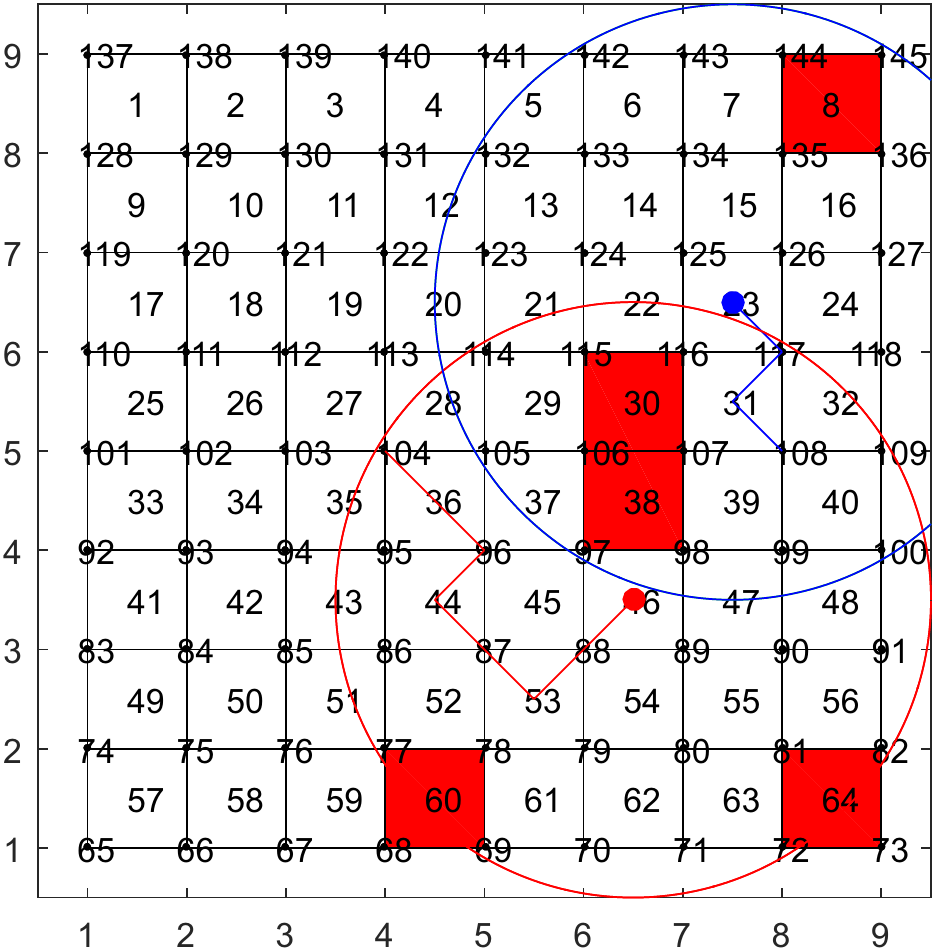}\label{p8}}\\
\subfloat[]{\includegraphics[width=0.62\linewidth]{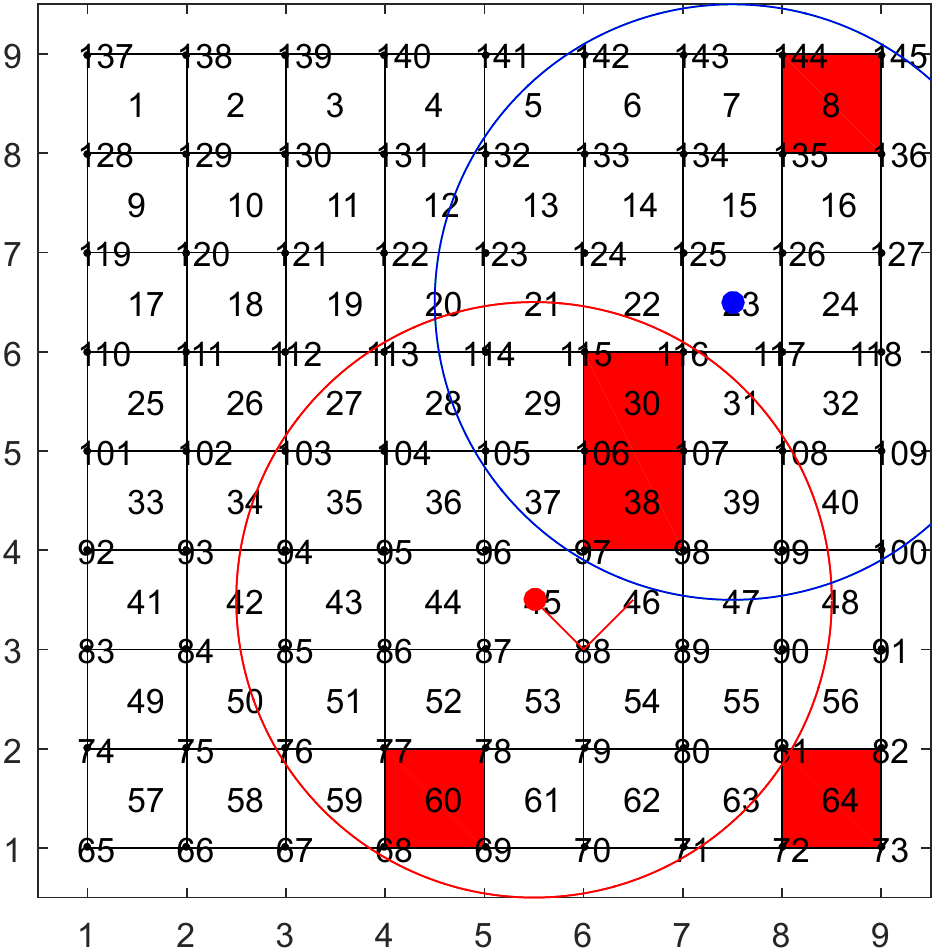}\label{p10}}\\
\subfloat[]{\includegraphics[width=0.62\linewidth]{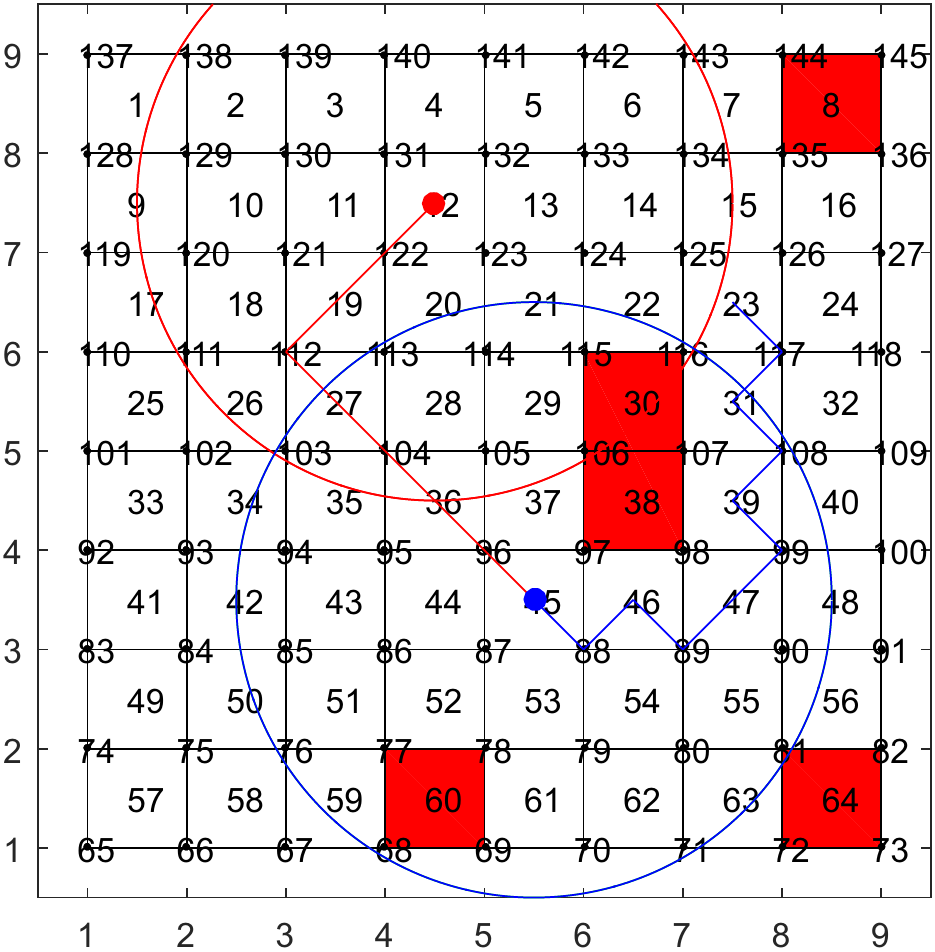}\label{p20}}
\caption{Case Study I: Robot trajectories due to the execution of the prefix part $\tau^{\text{pre}}$. (a), (b), and (c) illustrate the states $\tau^{\text{pre}}(k)$ of the prefix part, for $k\in\{1,\dots,8\}$, $k\in\{8,9,10\}$, and $k\in\{10,\dots,20\}$. The projection of the states $\tau^{\text{pre}}(8)$, $\tau^{\text{pre}}(10)$, and $\tau^{\text{pre}}(20)$ onto the trace-included wTS of the red and blue robot are depicted by red and blue disks, respectively. The red and blue circles illustrate the part of the workspace that falls within the influence radius of the red and blue robot in these states.  Filled in red regions are obstacles in the workspace.} %when their states are determined by $\tau^{\text{pre}}(8)$, $\tau^{\text{pre}}(10)$, and $\tau^{\text{pre}}(20)$.}
\label{sim1pre}
\end{figure}

%% file: Simulation.tex
% \begin{figure*}[h]
% \centering
% \subfloat[]{\includegraphics[width=0.30\linewidth]{prefix_e8.eps}\label{p8}}
% \subfloat[]{\includegraphics[width=0.30\linewidth]{prefix_e10.eps}\label{p10}}
% \subfloat[]{\includegraphics[width=0.30\linewidth]{prefix_e20.eps}\label{p20}}
% \caption{Case Study I: Robot trajectories due to the execution of the prefix part $\tau^{\text{pre}}$. (a), (b), and (c) illustrate the states $\tau^{\text{pre}}(k)$ of the prefix part, for $k\in\{1,\dots,8\}$, $k\in\{8,9,10\}$, and $k\in\{10,\dots,20\}$. The projection of the states $\tau^{\text{pre}}(8)$, $\tau^{\text{pre}}(10)$, and $\tau^{\text{pre}}(20)$ onto the trace-included wTS of the red and blue robot are depicted by red and blue disks, respectively. The red and blue circles illustrate the part of the workspace that falls within the influence radius of the red and blue robot in these states.} %when their states are determined by $\tau^{\text{pre}}(8)$, $\tau^{\text{pre}}(10)$, and $\tau^{\text{pre}}(20)$.}
% \label{sim1pre}
% \end{figure*}

In this section we present three case studies, implemented using MATLAB R2015b on a computer with Intel Core i7-2670QM at 2.2 GHz and 4 GB RAM, that illustrate our proposed algorithm. The considered case studies pertain to motion planning problems with PBA that have $273,325$, $21,340,375$, and $512, 778, 725, 000\approx 5\cdot 10^{11}$ states, respectively. Recall that the state-space of the PBA defined in Definition \ref{def:pba} has $\Pi_{i=1}^N|\mathcal{Q}_i||\mathcal{Q}_B|$ states. Note that the last two case studies pertain to planning problems that cannot be solved by the standard optimal control synthesis algorithms \cite{kloetzer2010automatic,ulusoy2013optimality,ulusoy2014optimal}, discussed in Section ~\ref{sec:Intro}, that rely on the explicit construction of the PBA defined in Section \ref{sec:model_checking} due to memory requirements. In fact, our Matlab implementation of the algorithm  described in Section \ref{sec:model_checking} cannot synthesize plans for PBA with more than few millions of states and transitions. 
Also, the sampling-based optimal control synthesis algorithm in \cite{kantaros2017Csampling} using uniform density functions failed to find a feasible solution within $55$ hours for the third case study although, in general, it can solve problems with order $10^{10}$ states. The algorithm proposed here can solve larger planning problems than \cite{kantaros2017Csampling}, if the size of the sets $|\Pi_i|$, i.e., the number of atomic propositions that appear in the tasks, are small and/or if the length (number of states) of the shortest paths computed to construct the initial wTS's $\text{wTS}_{0,i}’$ are small.
In all case studies, the costs $w_i$, defined in Definition \ref{def:wts} are associated with the distance between states. Also, we select $R_{I_{robot}}=3$, $R_{I_{robot}}=3.7$, and $R_{I_{robot}}=0.9$ distance units for the first, second and third case study. Finally, notice that the off-the-shelf model checkers SPIN \cite{holzmann2004spin} and NuSMV \cite{cimatti2002nusmv} cannot be used to synthesize plans that satisfy the LTL formula in \eqref{eq:phi}. The reason is that weights associated with distance need to be assigned to transitions of the wTS's, which is not possible in either SPIN or NuSMV.  Videos of the simulations for all case studies can be viewed in Supplemental Video 1. %\textcolor{cyan}{Do we have updated video showing the new case studies in DropBox folder?}

\subsection{Case Study I}

\begin{figure}%[h]
\centering
\subfloat[Case Study I]{\includegraphics[width=0.28\textwidth]{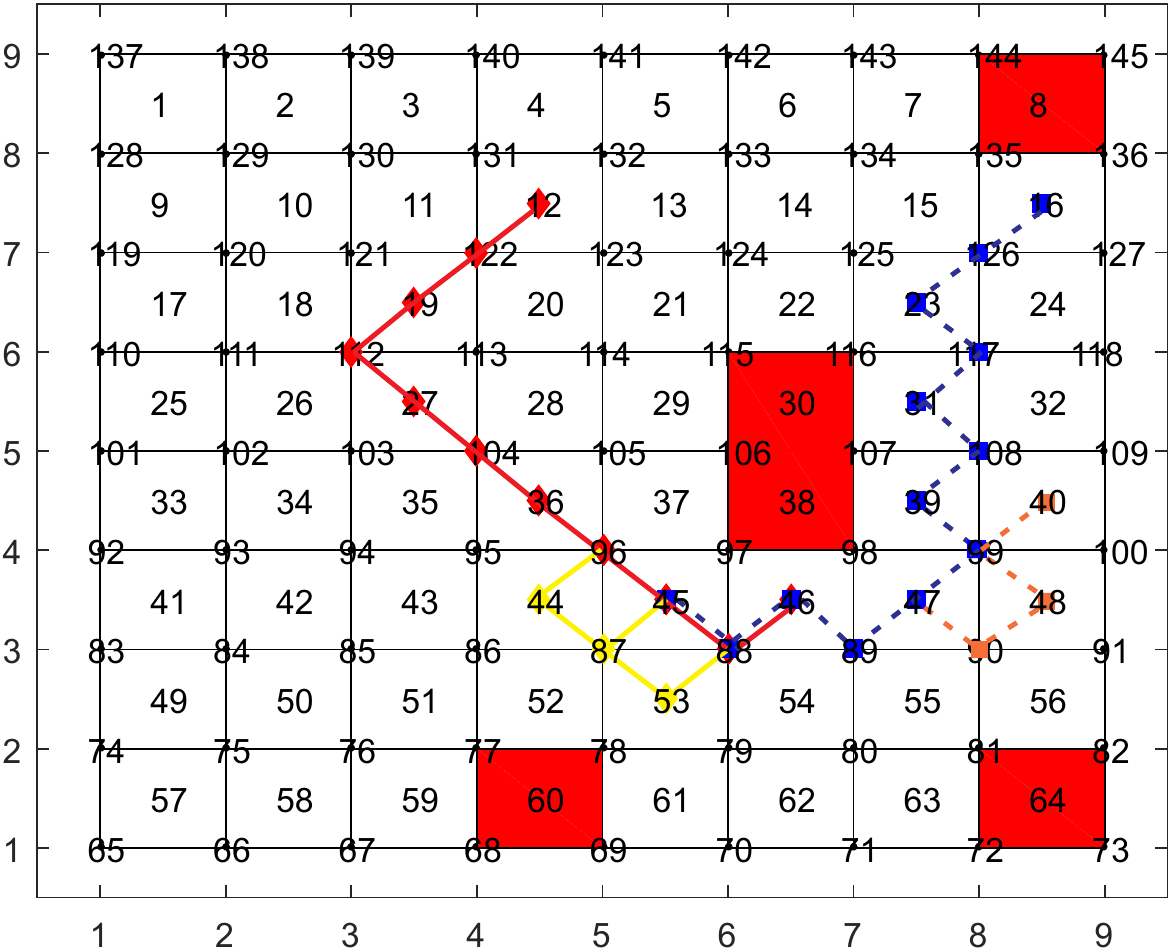}\label{trTS1}}\\
\subfloat[Case Study II]{\includegraphics[width=0.28\textwidth]{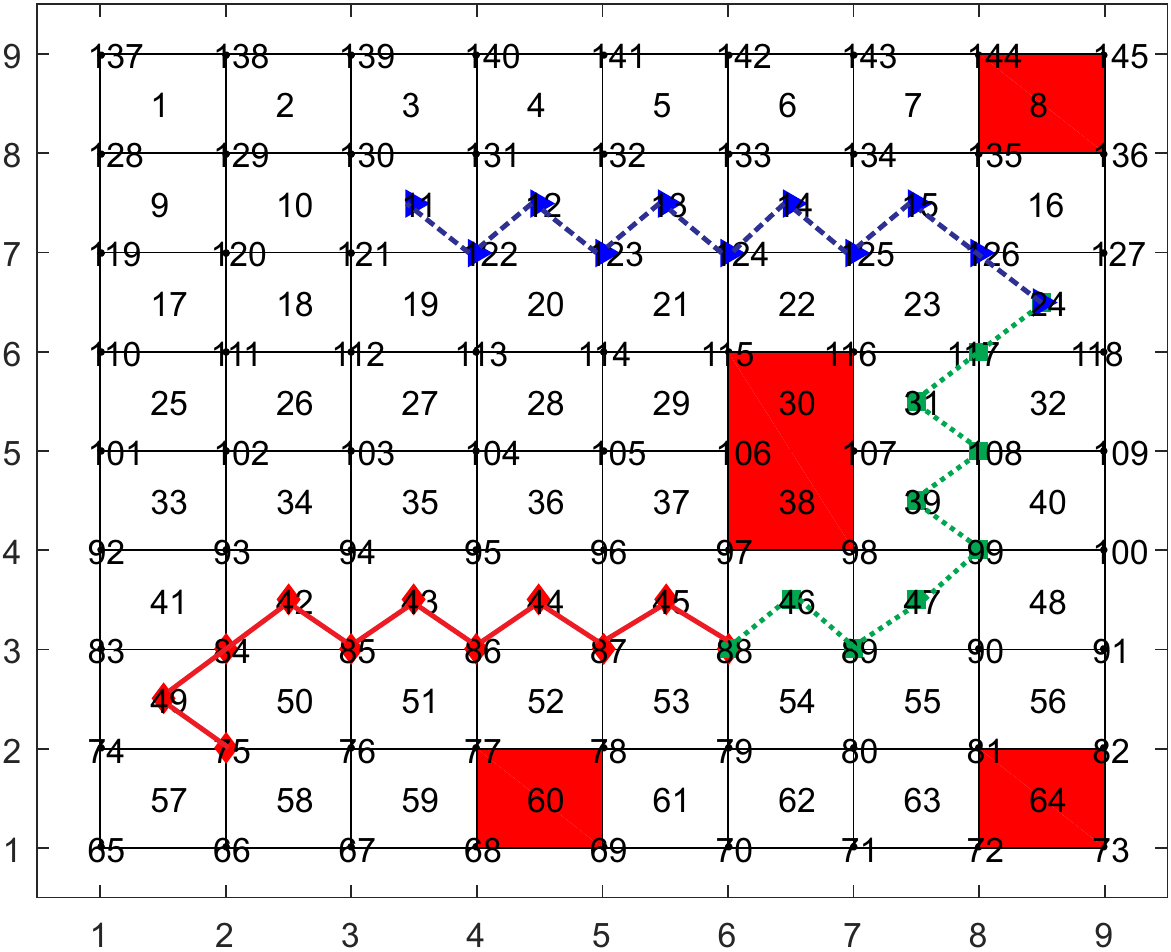}\label{trTS2}}\\
\subfloat[Case Study III]{\includegraphics[width=0.28\textwidth]{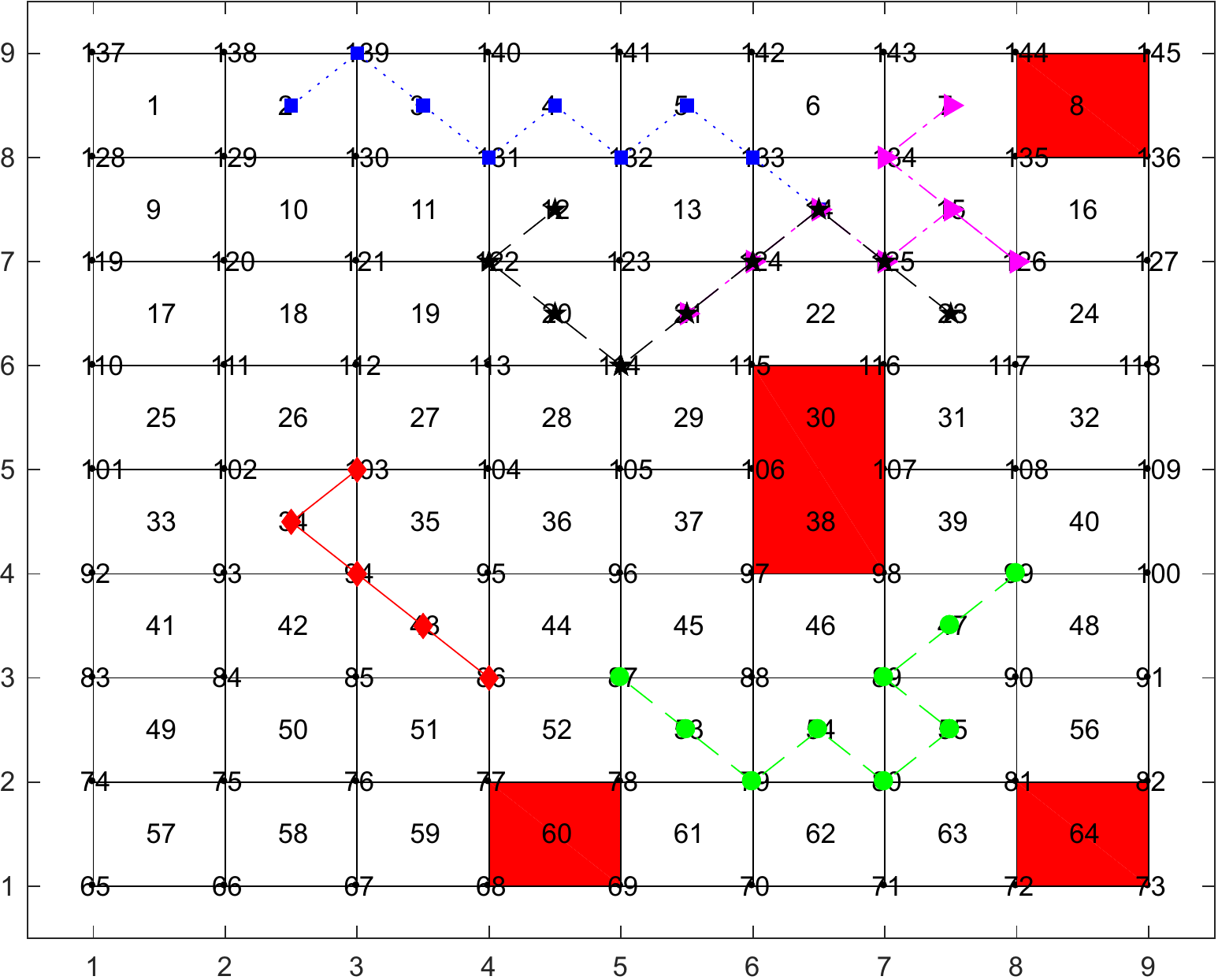}\label{trTS3}}
\caption{Graphical depiction of the wTS constructed for each robot for the first (a), the second (b), and the third (c) case study. The transition system for the red, blue, green, magenta, and black robot are depicted with corresponding color. In (a) the yellow diamonds and brown squares correspond to states that were added to the wTS at the first and the second iteration of Algorithm \ref{alg:update}. Red filled regions stand for obstacles in the workspace.}
\label{ts}
\end{figure}

% \begin{figure*}[h]
% \centering
% \subfloat[]{\includegraphics[width=0.30\linewidth]{suf_e3.eps}\label{s3}}
% \subfloat[]{\includegraphics[width=0.30\linewidth]{suf_e7.eps}\label{s7}}
% \subfloat[]{\includegraphics[width=0.30\linewidth]{suf_e13.eps}\label{s13}}\\
% \subfloat[]{\includegraphics[width=0.30\linewidth]{suf_e15.eps}\label{s15}}
% \subfloat[]{\includegraphics[width=0.30\linewidth]{suf_e19.eps}\label{s19}}
% \subfloat[]{\includegraphics[width=0.30\linewidth]{suf_e25.eps}\label{s25}}
% \caption{Case Study I: Robot trajectories due to a single execution of the suffix part $\tau^{\text{suf}}$. (a)-(f) illustrate the states $\tau^{\text{suf}}(k)$ of the suffix part, for $k\in\{1,2,3\}$, $k\in\{3,\dots,7\}$, $k\in\{7,\dots,13\}$, $k\in\{13,\dots,15\}$, $k\in\{15,\dots,19\}$, and $k\in\{19,\dots,25\}$. The projection of the states $\tau^{\text{suf}}(3)$, $\tau^{\text{suf}}(7)$, $\tau^{\text{suf}}(13)$, $\tau^{\text{suf}}(15)$, $\tau^{\text{suf}}(19)$, and $\tau^{\text{suf}}(25)$ onto the trace-included wTS of the red and blue robot are depicted by red and blue disks, respectively, in each figure. The red and blue circles illustrate the part of the workspace that falls within the influence radius of the red and blue robot are in these states.}% are determined by $\tau^{\text{suf}}(3)$, $\tau^{\text{suf}}(7)$, $\tau^{\text{suf}}(13)$, $\tau^{\text{suf}}(15)$, $\tau^{\text{suf}}(19)$, and $\tau^{\text{suf}}(25)$.}
% \label{sim1suf}
% \end{figure*}

\begin{figure}[h]
\centering
\subfloat[]{\includegraphics[width=0.42\linewidth]{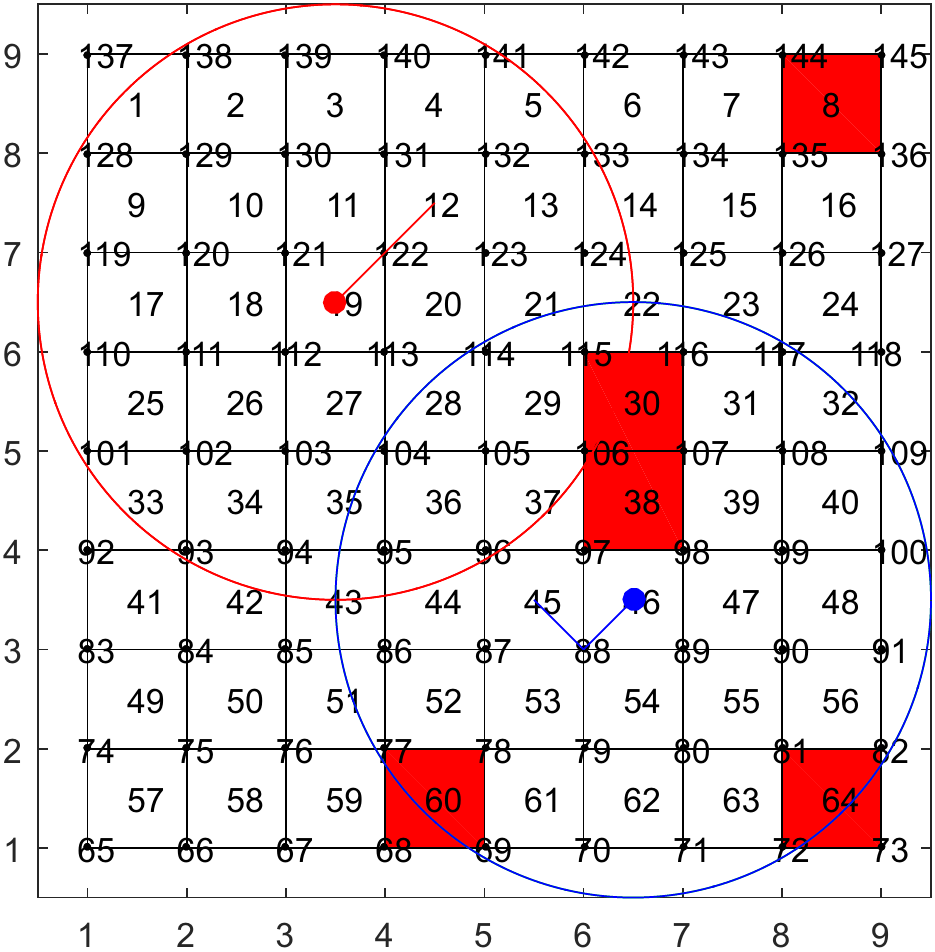}\label{s3}}
\subfloat[]{\includegraphics[width=0.42\linewidth]{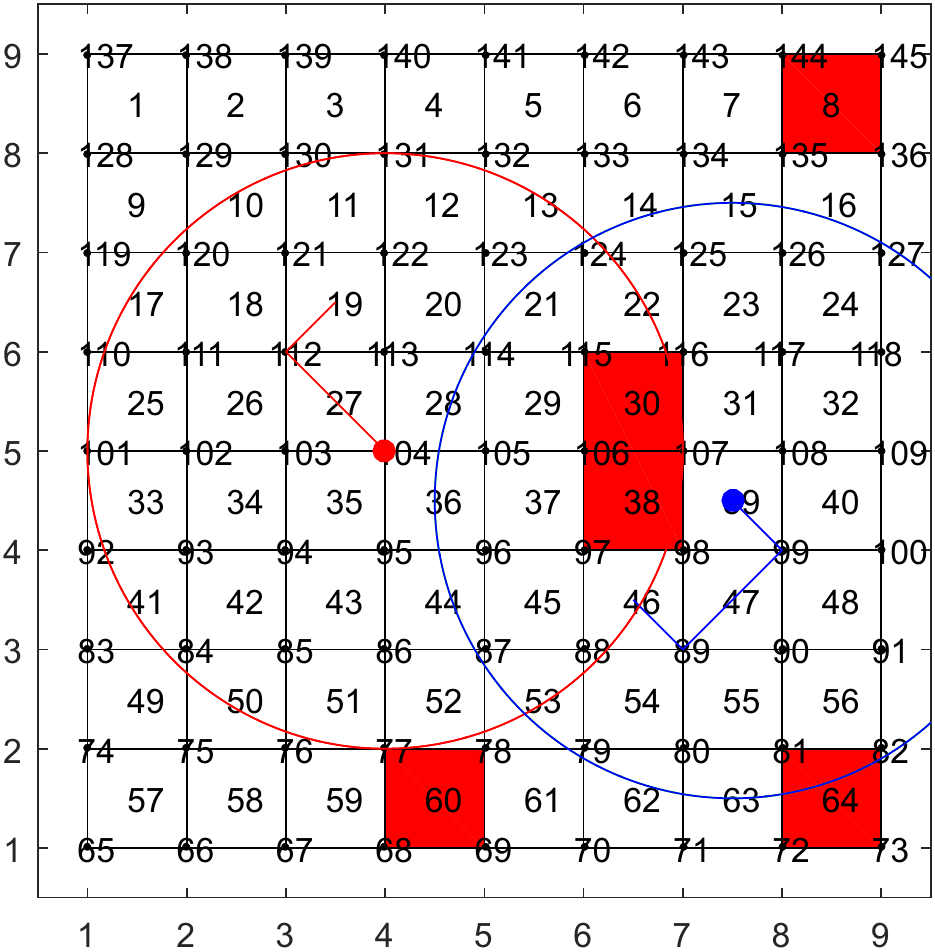}\label{s7}}\\
\subfloat[]{\includegraphics[width=0.42\linewidth]{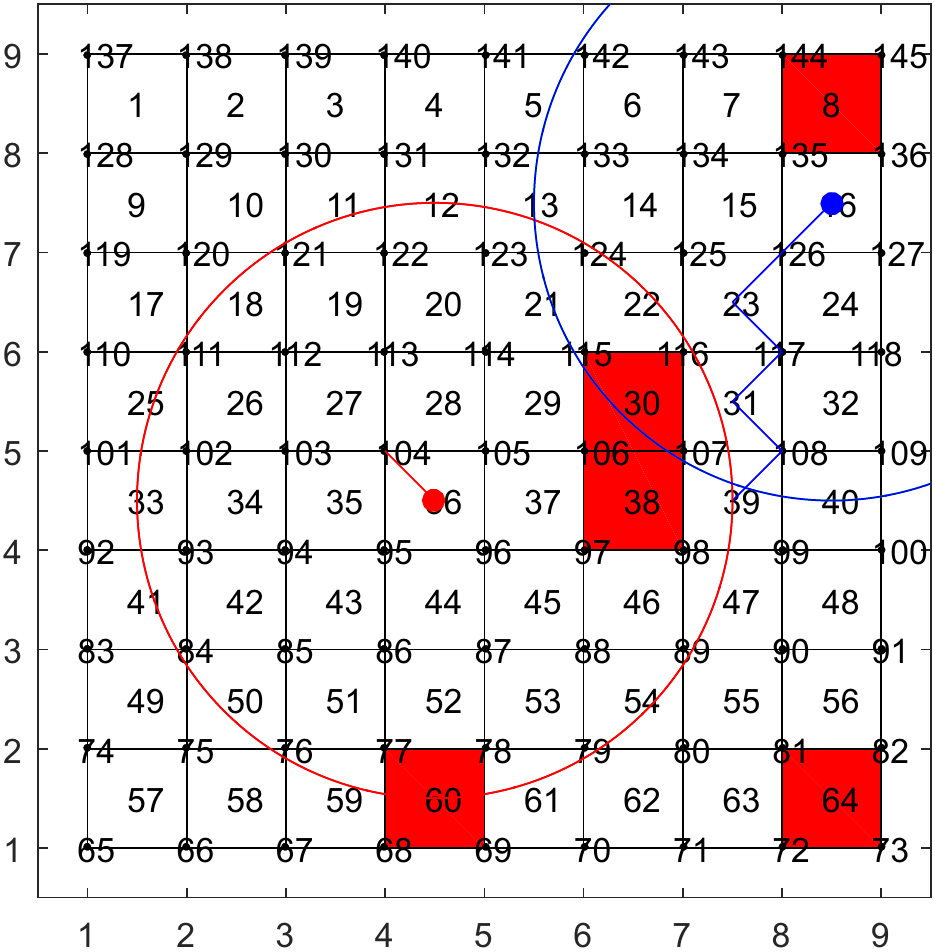}\label{s13}}
\subfloat[]{\includegraphics[width=0.42\linewidth]{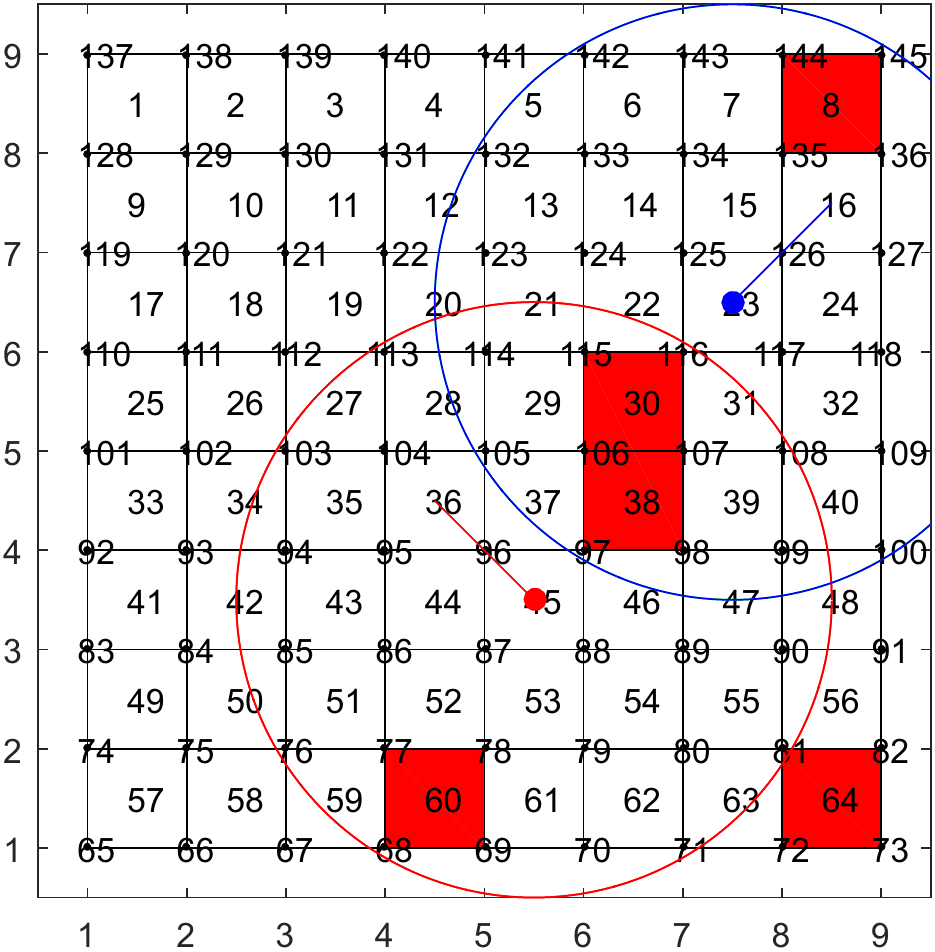}\label{s15}}\\
\subfloat[]{\includegraphics[width=0.42\linewidth]{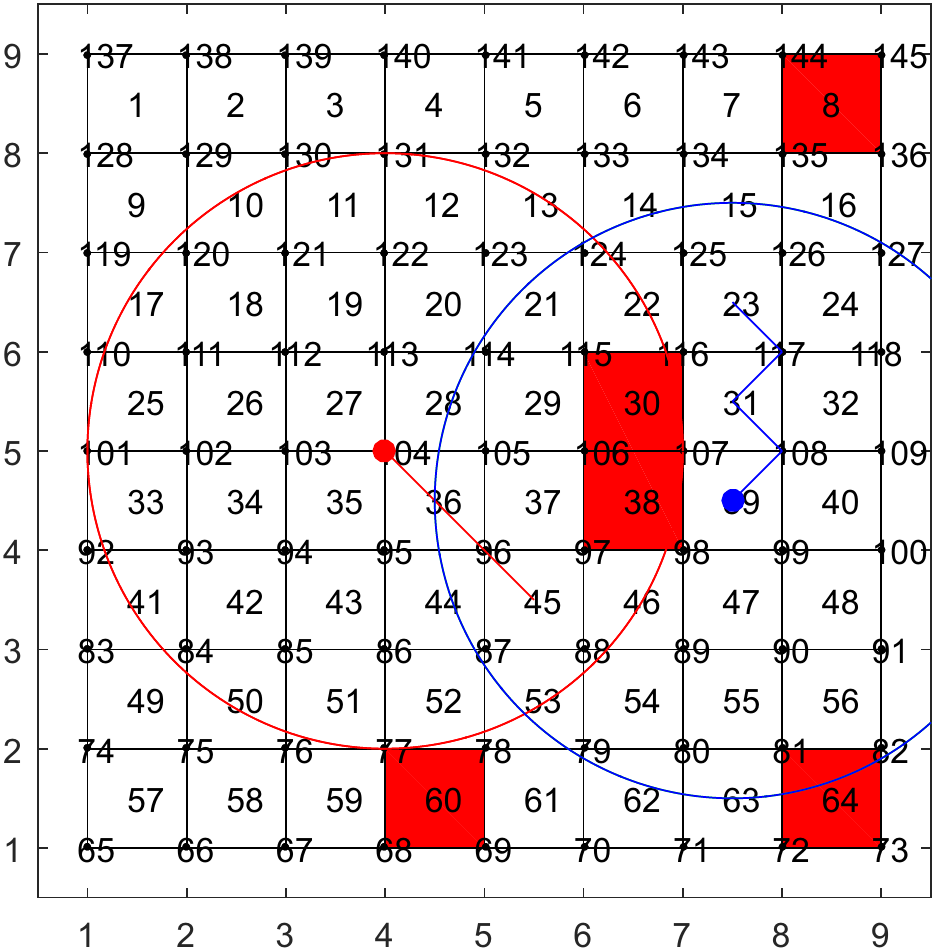}\label{s19}}
\subfloat[]{\includegraphics[width=0.42\linewidth]{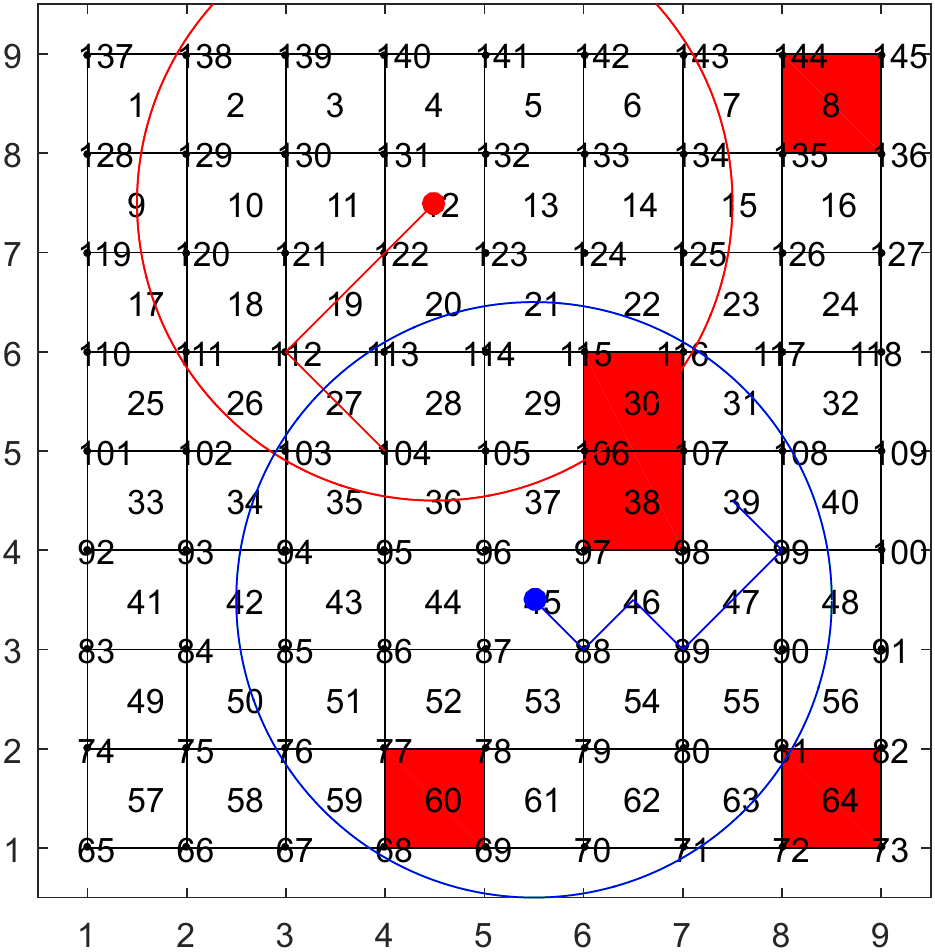}\label{s25}}
\caption{Case Study I: Robot trajectories due to a single execution of the suffix part $\tau^{\text{suf}}$. (a)-(f) illustrate the states $\tau^{\text{suf}}(k)$ of the suffix part, for $k\in\{1,2,3\}$, $k\in\{3,\dots,7\}$, $k\in\{7,\dots,13\}$, $k\in\{13,\dots,15\}$, $k\in\{15,\dots,19\}$, and $k\in\{19,\dots,25\}$. The projection of the states $\tau^{\text{suf}}(3)$, $\tau^{\text{suf}}(7)$, $\tau^{\text{suf}}(13)$, $\tau^{\text{suf}}(15)$, $\tau^{\text{suf}}(19)$, and $\tau^{\text{suf}}(25)$ onto the wTS of the red and blue robot are depicted by red and blue disks, respectively, in each figure. The red and blue circles illustrate the part of the workspace that falls within the influence radius of the red and blue robot, respectively.}% are in these states.}% are determined by $\tau^{\text{suf}}(3)$, $\tau^{\text{suf}}(7)$, $\tau^{\text{suf}}(13)$, $\tau^{\text{suf}}(15)$, $\tau^{\text{suf}}(19)$, and $\tau^{\text{suf}}(25)$.}
\label{sim1suf}
\end{figure}

\begin{figure}%[h]
\centering
\subfloat[Prefix Part.]{\includegraphics[width=0.30\textwidth]{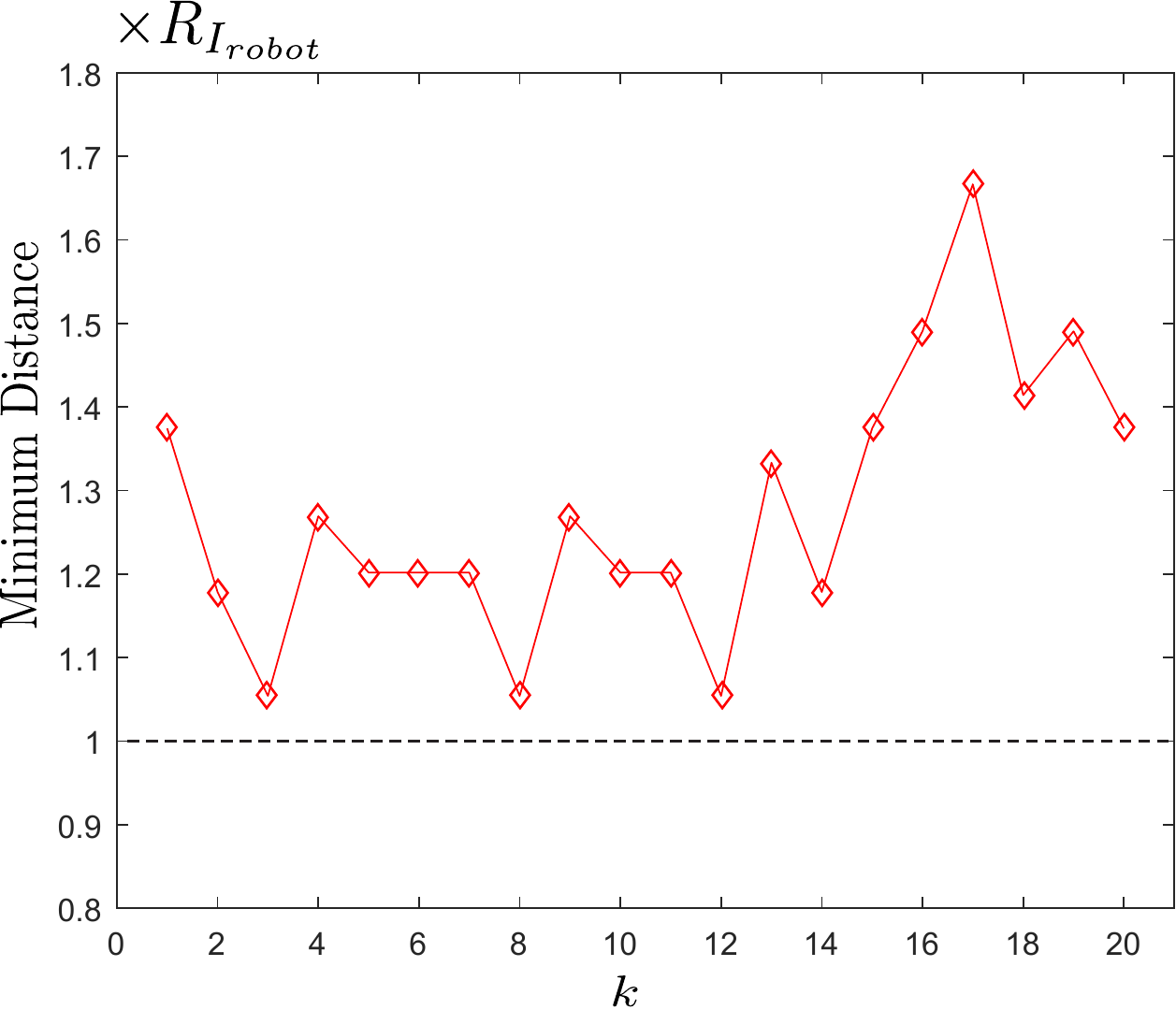}\label{dpre1}}\\
\subfloat[Suffix Part.]{\includegraphics[width=0.30\textwidth]{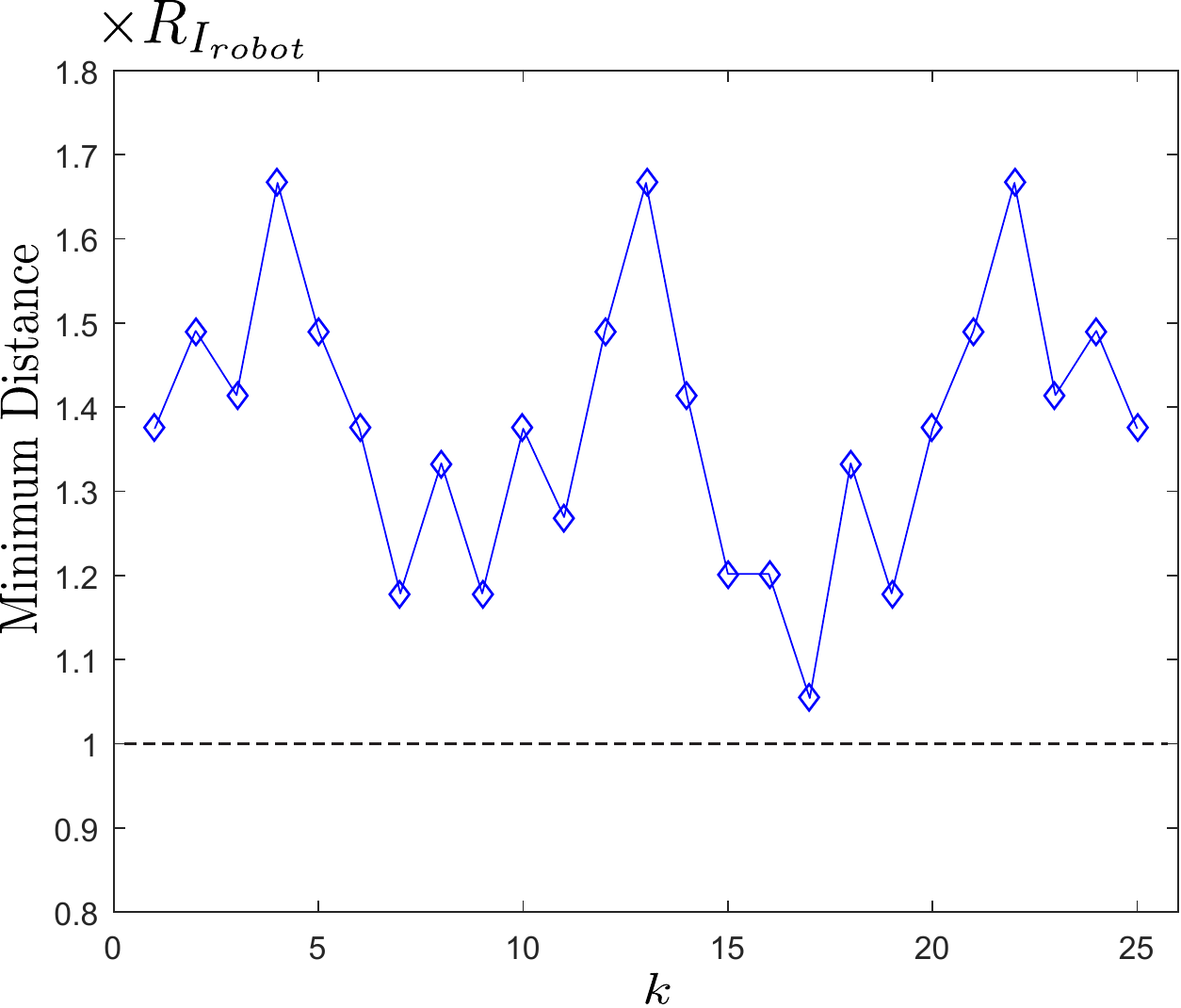}\label{dsuf1}}
\caption{Case Study I: Graphical depiction of the minimum distance among robots, when the robots are in states $\tau^{\text{pre}}(k)$, for all $k\in\{1,\dots,K\}$ (a) and $\tau^{\text{suf}}(k)$, for all $k\in\{1,\dots,S\}$ (b), where $K$ and $S$ stand for the number of states in the prefix $\tau^{\text{pre}}$ and suffix part $\tau^{\text{suf}}$.}
\label{distance1}
\end{figure}

In the first simulation study we consider a network of $N=2$ microrobots residing in a workspace, as shown in Figure~\ref{sim1pre}, with $|\mathcal{Q}_i|=145$, for both robots. The assigned temporal task is described by the following LTL formula:

\begin{align}\label{eq:ex1} 
&\phi_{\text{task}}=(\square\Diamond \pi_{r}^{{\bf{c}}_{45}})
\wedge (\neg \pi_{r}^{{\bf{c}}_{45}} \mathcal{U} \pi_{r}^{{\bf{c}}_{46}})\nonumber\\
&\wedge \square(\pi_{r}^{{\bf{c}}_{45}}\rightarrow\bigcirc(\neg \pi_{r}^{{\bf{c}}_{45}} \mathcal{U}\pi_{b}^{{\bf{c}}_{16}}))\nonumber\\ 
&\wedge (\square\Diamond \pi_{r}^{{\bf{c}}_{12}})
\wedge (\square\Diamond \pi_{b}^{{\bf{c}}_{45}})
\wedge (\square (\neg \pi_{\text{obs}})).
\end{align}

Specifically, the task in \eqref{eq:ex1} requires: (a) the red robot to visit  location ${\bf{c}}_{45}$ infinitely often, (b) the red robot to avoid location ${\bf{c}}_{45}$ until it visits once location ${\bf{c}}_{46}$, (c) once the red robot visits ${\bf{c}}_{45}$ it should avoid this location until the blue robot visits location  ${\bf{c}}_{16}$; this should occur infinitely often, (d) the red robot to visit location ${\bf{c}}_{12}$ infinitely often, (e) the blue robot to visit location ${\bf{c}}_{45}$ infinitely often, and (g) both robots to avoid the obstacles illustrated by filled red regions in the Figure~\ref{sim1pre}. This LTL formula corresponds to a NBA with $|\mathcal{Q}_B|=13$ states, $|\mathcal{Q}_B^0|=1$, $|\mathcal{Q}_B^F|=2$, and $61$ transitions. \footnote{The translation of the LTL formula to a NBA was made by the tool developed in \cite{gastin2001fast}.} 

The while-loop in Algorithm \ref{alg:update} was executed for two iterations until a motion plan is synthesized adding three states to the wTS of the red and the blue robot, respectively that belong to the neighborhood of location ${\bf{c}}_{45}\in\mathcal{S}_r$ and ${\bf{c}}_{99}\in\mathcal{S}_b$. Note that the proposed algorithm required $5$ minutes to synthesize a motion plan that satisfies the LTL formula \eqref{eq:ex1}. On the other hand, the existing optimal control synthesis method presented in Section \ref{sec:model_checking} synthesized the optimal plan in 30 hours, as it has to construct a larger PBA. 

The state-spaces of the constructed wTS's have $|\mathcal{Q}_{r,2}'|=14$ and $|\mathcal{Q}_{b,2}|=16$ states and the resulting PBA has 2912 states among which 760 are final states. Notice that state-space of the PBA has been decreased by almost 99$\%$. The constructed wTS for both robots are depicted in Figure \ref{trTS1}. Notice that $\mathcal{S}_r=\{{\bf{c}}_{45}\}$ and $\mathcal{S}_b=\mathcal{Q}_b$. The prefix and suffix part of the plan that satisfies $\phi=\phi_{\text{task}}\wedge\phi_c$, where $\phi_{\text{task}}$ is given in \eqref{eq:ex1} are depicted in Figure \ref{sim1pre} and \ref{sim1suf}, respectively. Figure \ref{distance1} shows the minimum distance among the robots during the execution of the prefix part and a single execution of the suffix part. Observe that the minimum distance between the robots is always greater than the robot influence radius $R_{I_{robot}}$, as expected.%, due to the satisfaction of the LTL specification $\phi_c$ defined in \eqref{eq:collision}. 

\subsection{Case Study II}
\label{sec:three_robot_study}
In the second simulation study, we consider a team of $N=3$ microrobots that reside in the same workspace considered in the previous case study. The microrobot team is responsible for accomplishing the following temporal task:

\begin{align}\label{eq:ex2} 
\phi_{\text{task}}=&(\square\Diamond \pi_{r}^{{\bf{c}}_{42}})
\wedge (\square\Diamond \pi_{r}^{{\bf{c}}_{88}})
\wedge (\square\Diamond \pi_{g}^{{\bf{c}}_{88}})
\wedge (\square\Diamond \pi_{g}^{{\bf{c}}_{24}})\nonumber\\
&\wedge (\square\Diamond \pi_{b}^{{\bf{c}}_{24}})
\wedge (\square\Diamond \pi_{b}^{{\bf{c}}_{12}})
\wedge (\square (\neg \pi_{\text{obs}})).
\end{align}

In words the LTL-based task in \eqref{eq:ex2} requires: 
(a) the red robot to move back and forth between locations ${\bf{c}}_{88}$ and ${\bf{c}}_{42}$ infinitely often, (b) the green robot to  move back and forth between locations ${\bf{c}}_{88}$ and ${\bf{c}}_{24}$ infinitely often, (c) the blue robot to move back and forth between locations ${\bf{c}}_{24}$ and ${\bf{c}}_{12}$ infinitely often, and (d) all robots to avoid the obstacles in the workspace. This LTL formula corresponds to a NBA with $|\mathcal{Q}_B|=7$ states, $|\mathcal{Q}_B^0|=1$, $|\mathcal{Q}_B^F|=1$, and $34$ transitions.

\begin{figure}%[h]
\centering
\subfloat[]{\includegraphics[width=0.5\linewidth]{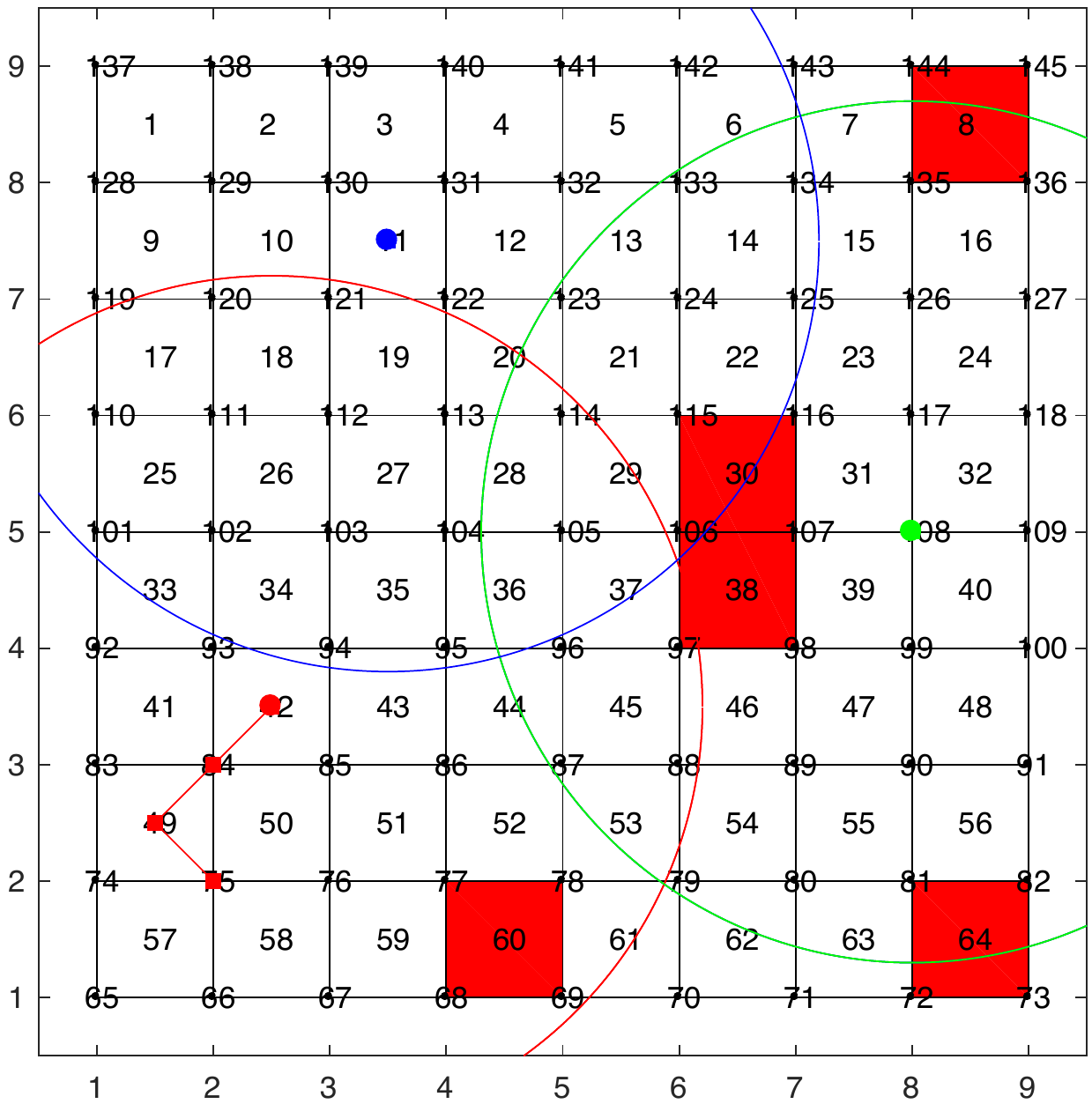}\label{p4_2}}
\subfloat[]{\includegraphics[width=0.5\linewidth]{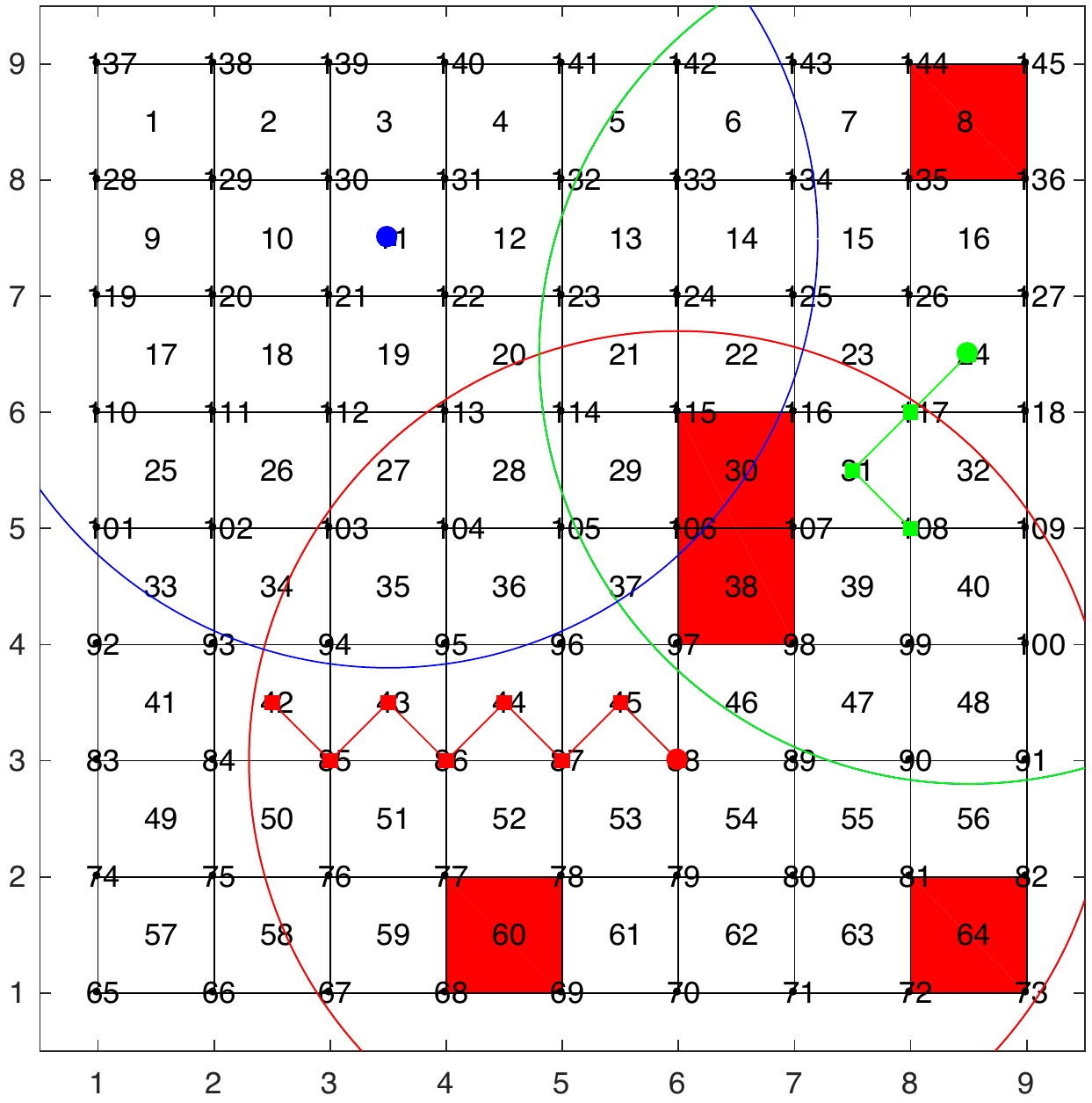}\label{p11_2}}\\
\subfloat[]{\includegraphics[width=0.5\linewidth]{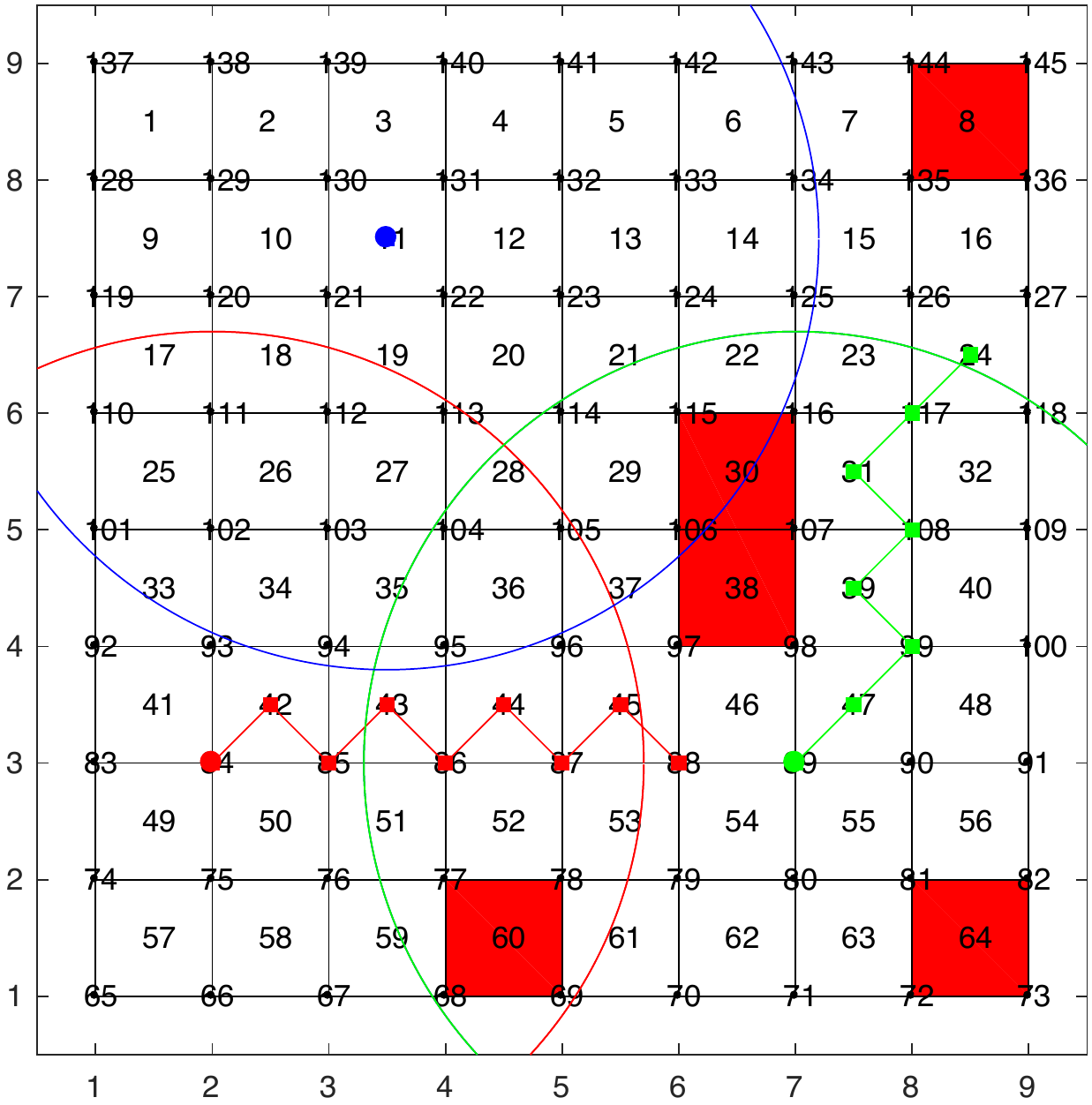}\label{p20_2}}
\subfloat[]{\includegraphics[width=0.5\linewidth]{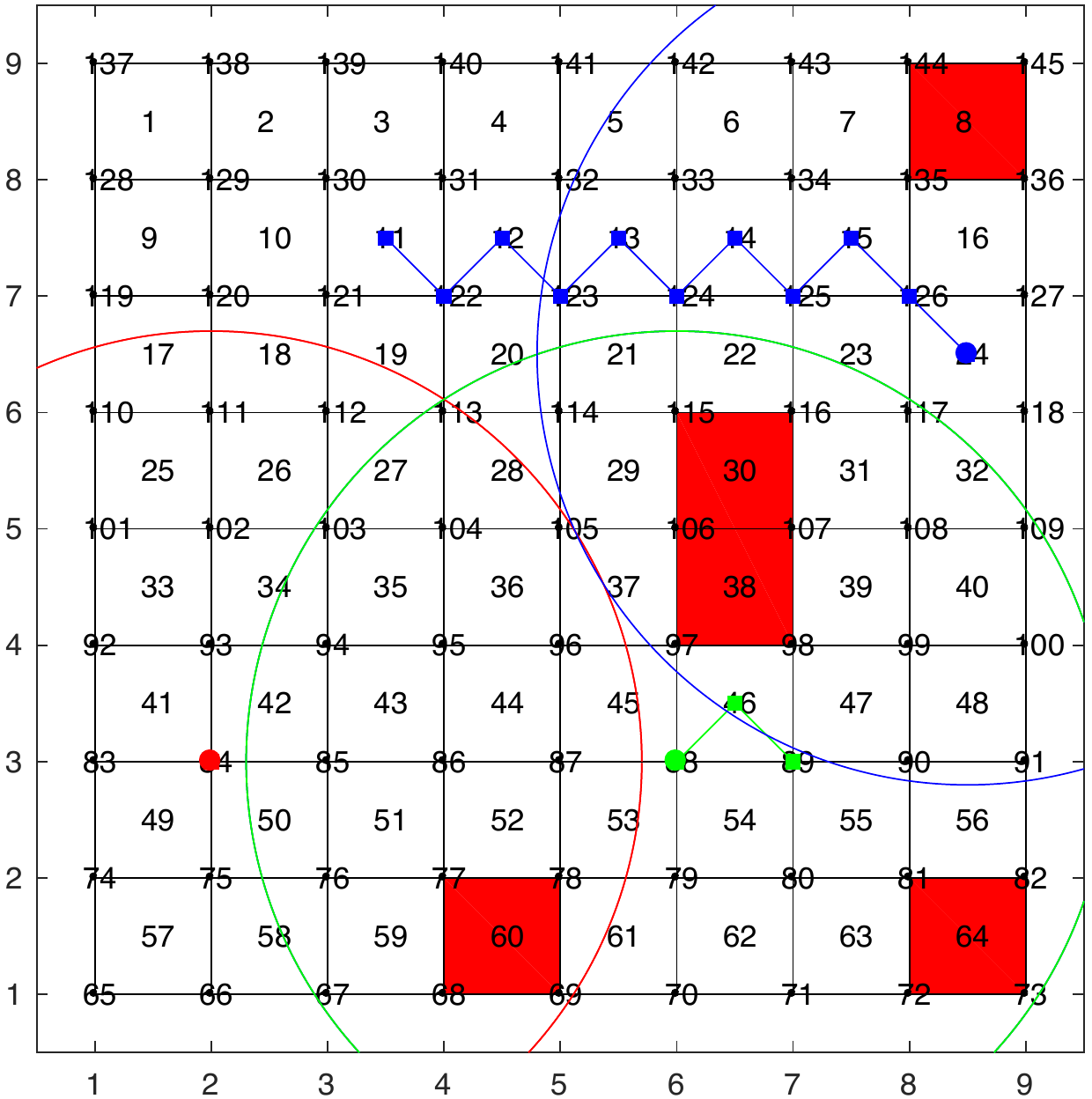}\label{p28_2}}
\caption{Case Study II: Robot trajectories due to the execution of the prefix part $\tau^{\text{pre}}$. Figures \ref{p4_2}, \ref{p11_2}, \ref{p20_2}, and \ref{p28_2} illustrate the states $\tau^{\text{pre}}(k)$ of the prefix part, for $k\in\{1,\dots,4\}$, $k\in\{4,\dots,11\}$, $k\in\{11,\dots,20\}$, and $k\in\{20,\dots,28\}$. The projection of the states $\tau^{\text{pre}}(4)$, $\tau^{\text{pre}}(11)$, $\tau^{\text{pre}}(20)$, and $\tau^{\text{pre}}(28)$  onto the wTS of the red, blue, and green robot are depicted by red, blue, and green disks, respectively. The red, blue, and green circles illustrate the part of the workspace that falls within the influence radius of the red, blue, and green robot in these states.}% are determined by $\tau^{\text{pre}}(4)$, $\tau^{\text{pre}}(11)$, $\tau^{\text{pre}}(20)$, and $\tau^{\text{pre}}(28)$.}
\label{sim2pre}
\end{figure}

\begin{figure}[h]
\centering
\subfloat[]{\includegraphics[width=0.5\linewidth]{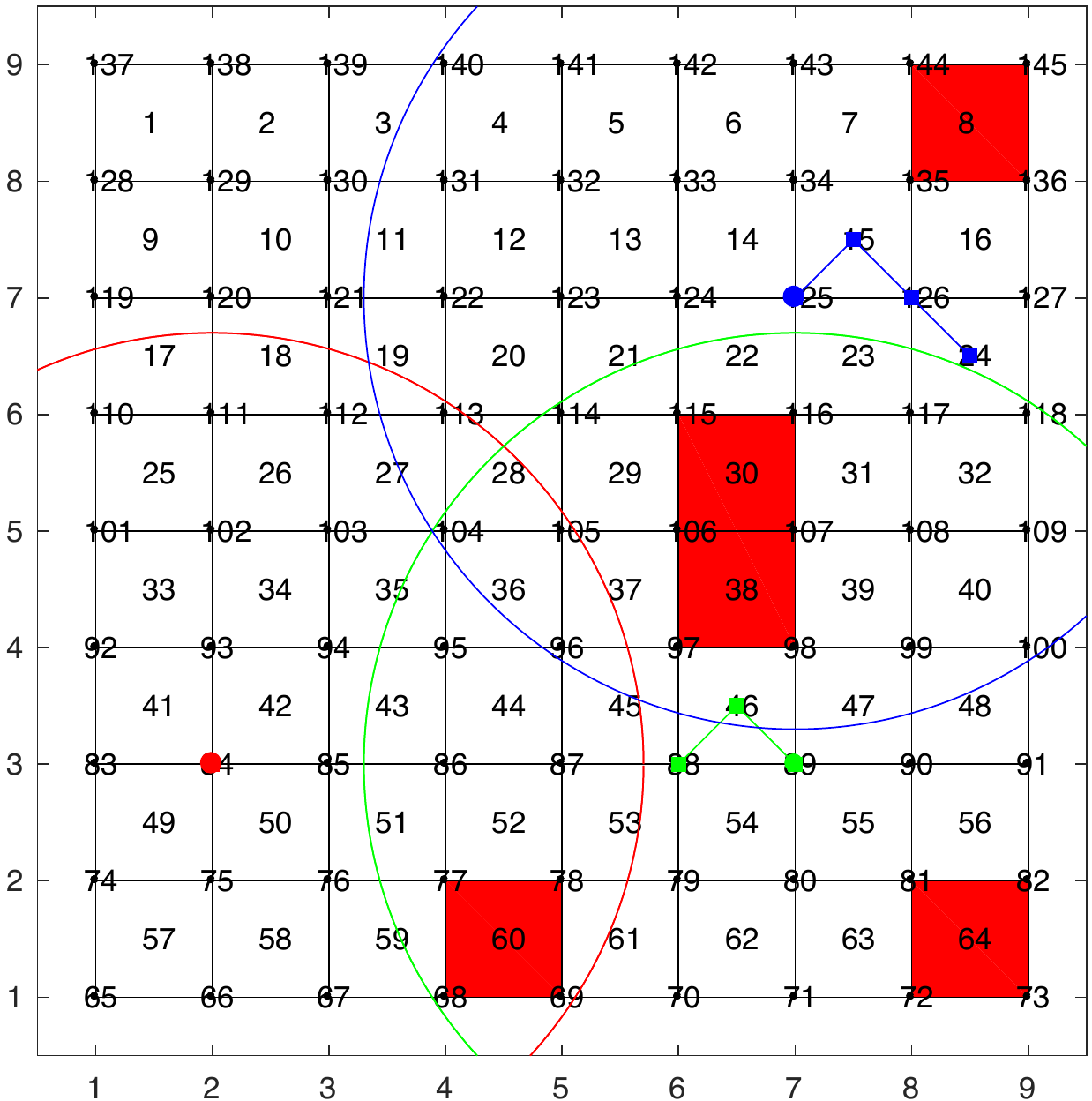}\label{s4_2}}
\subfloat[]{\includegraphics[width=0.5\linewidth]{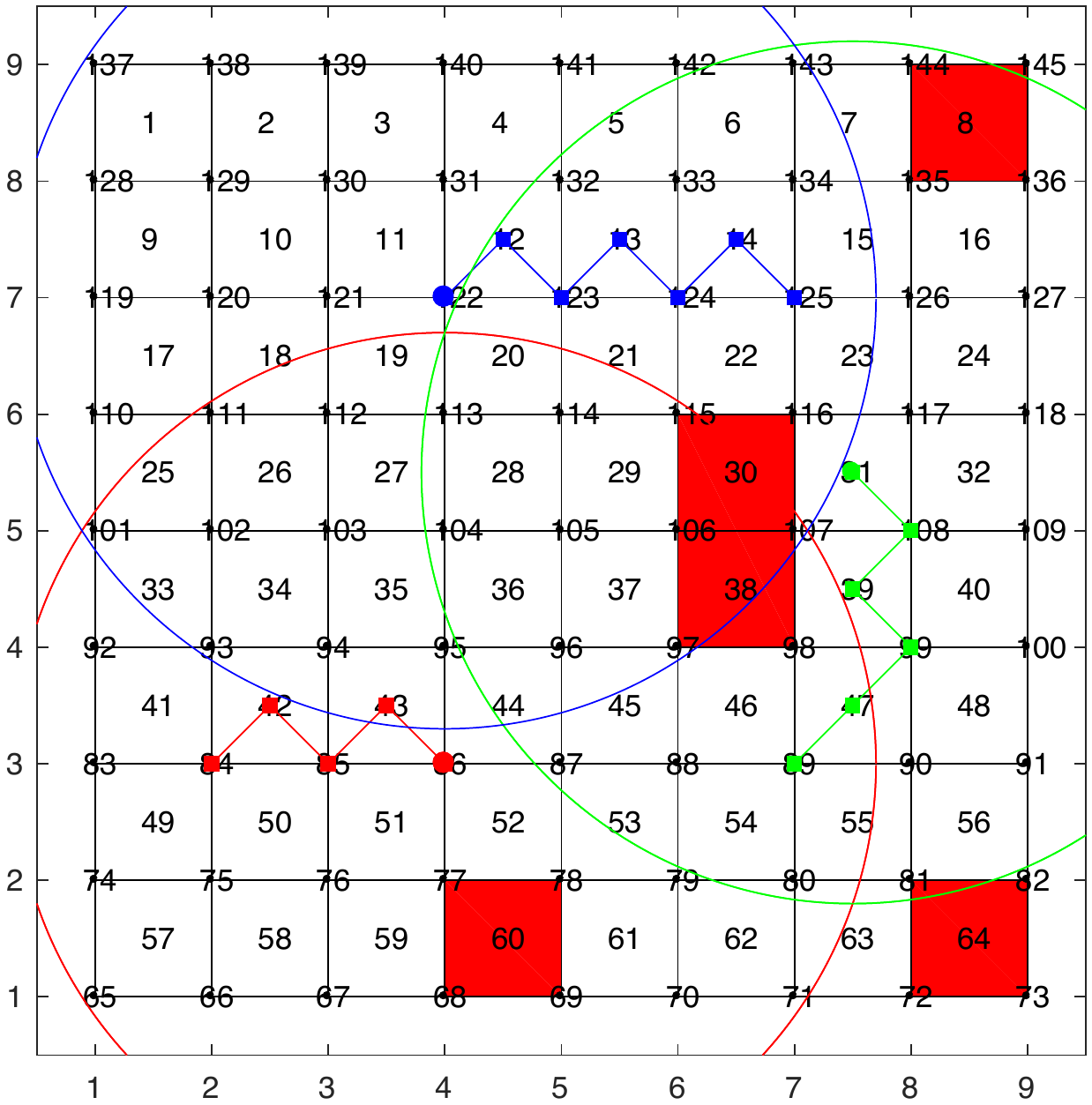}\label{s11_2}}\\
\subfloat[]{\includegraphics[width=0.5\linewidth]{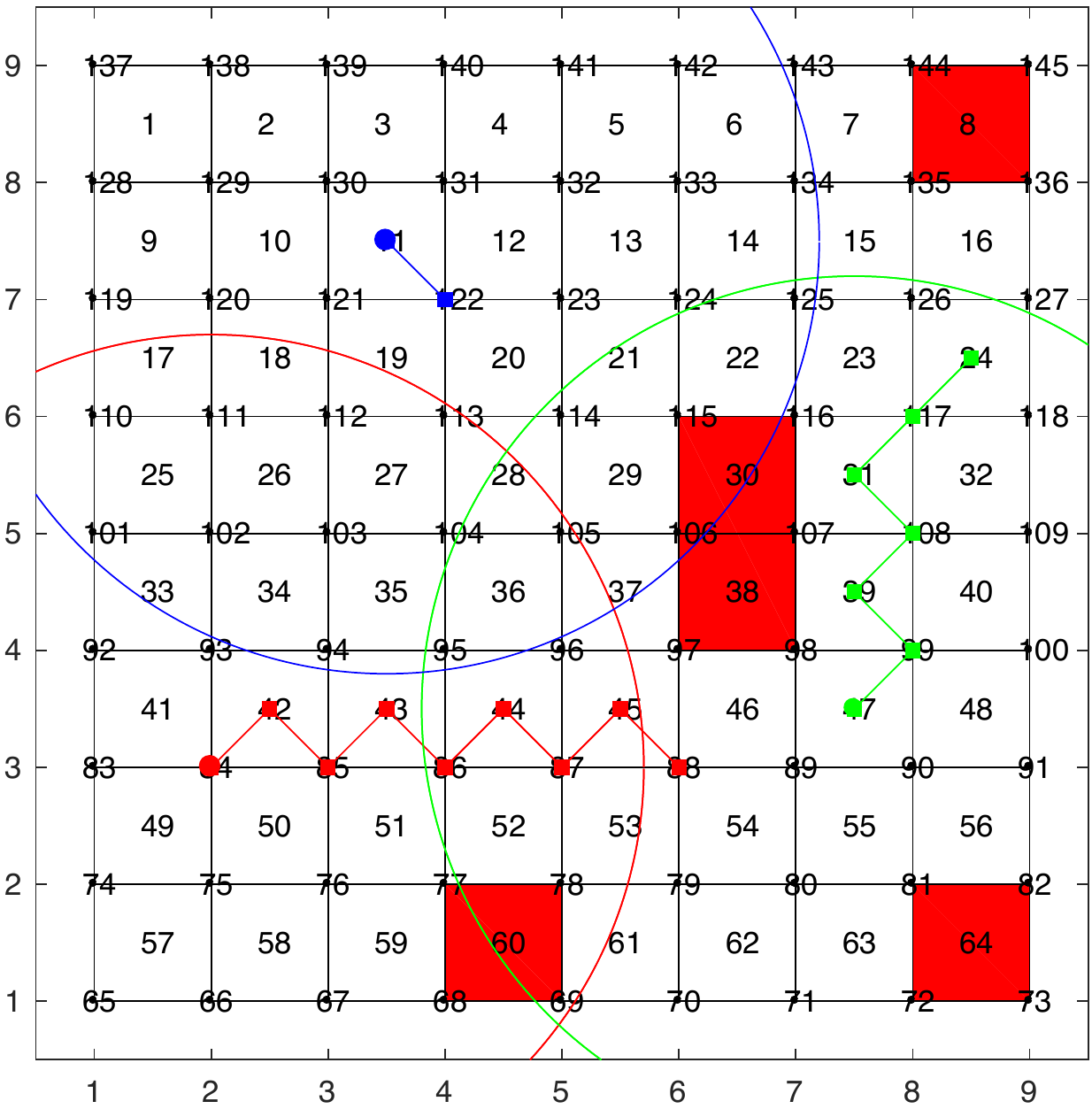}\label{s20_2}}
\subfloat[]{\includegraphics[width=0.5\linewidth]{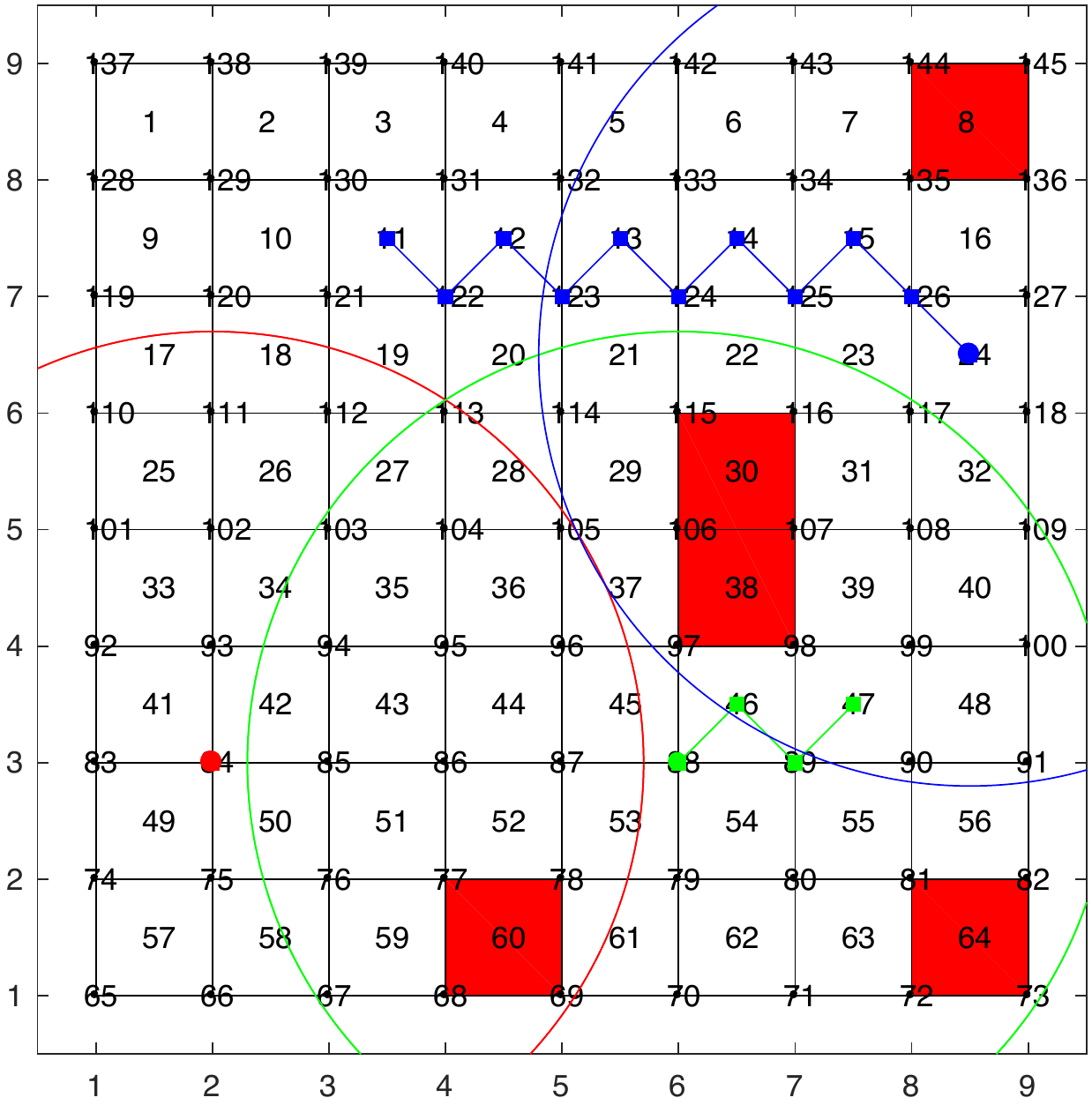}\label{s28_2}}
\caption{Case Study II: Robot trajectories due to a single execution of the suffix part $\tau^{\text{suf}}$. Figures \ref{s4_2}-\ref{s28_2} illustrate the states $\tau^{\text{suf}}(k)$ of the suffix part, for $k\in\{1,\dots,4\}$, $k\in\{4,\dots,11\}$, $k\in\{11,\dots,23\}$, and $k\in\{20,\dots,34\}$. The projection of the states $\tau^{\text{suf}}(4)$, $\tau^{\text{suf}}(11)$, $\tau^{\text{suf}}(23)$, and $\tau^{\text{suf}}(34)$ onto the wTS of the red, blue, and green robot are depicted by red, blue, and green disks, respectively, in Figures \ref{s4_2}-\ref{s28_2}. The red, blue, and green circles illustrate the part of the workspace that falls within the influence radius of the red, blue, and green robot when their states are determined by $\tau^{\text{suf}}(4)$, $\tau^{\text{suf}}(11)$, $\tau^{\text{suf}}(23)$, and $\tau^{\text{suf}}(34)$.}
\label{sim2suf}
\end{figure}

The initially constructed wTS's have state-spaces with $|\mathcal{Q}_{r,0}'|=11$, $|\mathcal{Q}_{g,0}'|=10$, and $|\mathcal{Q}_{b,0}'|=11$ states and the resulting PBA has 8470 states among which $1573$ are final states.  Notice that state-space of the PBA has been decreased by almost $99$$\%$ again. The initially constructed wTS's for all three robots are depicted in Figure \ref{trTS2}. Notice that $\mathcal{S}_r=\mathcal{Q}_r$, $\mathcal{S}_b=\mathcal{Q}_b$, and $\mathcal{S}_g=\mathcal{Q}_g$. Also, the initially constructed wTS's determined by the sets $\mathcal{S}_r$, $\mathcal{S}_g$, and $\mathcal{S}_b$ are expressive enough to generate a plan that satisfies the temporal logic task \eqref{eq:ex2}. As a result, the while-loop in Algorithm \ref{alg:update} did not introduce any states to the wTS's. Note that the proposed algorithm required $37$ minutes to synthesize a motion plan that satisfies the LTL formula \eqref{eq:ex2}. The prefix and suffix part of this plan are depicted in Figure \ref{sim2pre} and \ref{sim2suf}, respectively. The minimum distance among the robots during the execution of the algorithm always satisfy the LTL specification $\phi_c$ defined in \eqref{eq:collision}.

\subsection{Case Study III}
In the second simulation study, we consider a team of $N=5$ microrobots that reside in the same workspace considered in the previous case studies. The microrobot team is responsible for accomplishing the following temporal task:
\begin{align}\label{eq:ex3} 
\phi_{\text{task}}=&(\square\Diamond \pi_{r}^{{\bf{c}}_{89}})\wedge (\square\Diamond \pi_{b}^{{\bf{c}}_{54}})\wedge (\square\Diamond \pi_{k}^{{\bf{c}}_{114}})\nonumber\\
&\wedge (\square\Diamond (\pi_{g}^{{\bf{c}}_{87}}\wedge \pi_{r}^{{\bf{c}}_{43}}))\wedge (\square\Diamond (\pi_{g}^{{\bf{c}}_{87}}\wedge \pi_{r}^{{\bf{c}}_{43}}))\nonumber\\
&\wedge (\square\Diamond (\pi_{b}^{{\bf{c}}_{4}}\wedge \pi_{k}^{{\bf{c}}_{12}}))\wedge (\square\Diamond (\pi_{b}^{{\bf{c}}_{14}}\wedge \pi_{m}^{{\bf{c}}_{21}}))\nonumber\\
&
\wedge (\Diamond (\pi_{m}^{{\bf{c}}_{126}}\wedge \pi_{g}^{{\bf{c}}_{55}}))
\wedge (\square (\neg \pi_{\text{obs}})).
\end{align}

In words the LTL task in \eqref{eq:ex2} requires: 
(a) the red, the blue, and the black robot to visit locations ${\bf{c}}_{89}$, ${\bf{c}}_{54}$, and ${\bf{c}}_{114}$  infinitely often, (b) the green and the red robot to visit locations ${\bf{c}}_{87}$ and ${\bf{c}}_{43}$ simultaneously and infinitely often, (c) the blue and the black robot to visit locations ${\bf{c}}_{4}$ and ${\bf{c}}_{12}$ simultaneously and infinitely often, (d) the blue and the magenta robot to visit locations ${\bf{c}}_{14}$ and ${\bf{c}}_{21}$ simultaneously and infinitely often, (e) the magenta and the green robot to visit locations ${\bf{c}}_{126}$ and ${\bf{c}}_{55}$, respectively, simultaneously and infinitely often, and (d) all robots to avoid the obstacles in the workspace. This LTL formula corresponds to a NBA with $|\mathcal{Q}_B|=8$ states, $|\mathcal{Q}_B^0|=1$, $|\mathcal{Q}_B^F|=1$, and $36$ transitions.

The initially constructed wTS's $\text{wTS}_{i,0}'$ have state-spaces with $|\mathcal{Q}_{r,0}'|=5$, $|\mathcal{Q}_{g,0}'|=9$, $|\mathcal{Q}_{b,0}'|=9$, $|\mathcal{Q}_{m,0}'|=9$, and $|\mathcal{Q}_{k,0}'|=10$ states and the corresponding PBA $P_{0}$ has $291600$ states among which $19960$ are final states. These wTS's are expressive enough to generate a plan that satisfies \eqref{eq:ex3} and are depicted in Figure \ref{trTS3}.
Note that $55$ hours were required to synthesize the optimal plan.
Specifically, $18$ hours were required to construct the product automaton  $P_0$ and $37$ hours were required to construct the prefix and suffix part for each final state in $P_0$ to select the prefix-suffix plan with the minimum cost. Notice that $\mathcal{S}_i=\mathcal{Q}_{i,0}'$,  for all robots $i\in\{r,g,b,m,k\}$. The minimum distance among the robots during the execution of the algorithm always satisfy the LTL specification $\phi_c$ defined in \eqref{eq:collision}. 
%

%%%%

% \begin{figure}[t]
% \centering
% \subfloat[Prefix Part.]{\includegraphics[width=0.42\textwidth]{distance_pre2.eps}\label{dpre2}}\\
% \subfloat[Suffix Part.]{\includegraphics[width=0.42\textwidth]{distance_suf2.eps}\label{dsuf2}}
% \caption{Case Study II: Graphical depiction of the minimum distance among robots, when the robots are in states $\tau^{\text{pre}}(k)$, for all $k\in\{1,\dots,K\}$ (Figure \ref{dpre2}) and $\tau^{\text{suf}}(k)$, for all $k\in\{1,\dots,S\}$ (Figure \ref{dsuf2}), where $K$ and $S$ stand for the number of states in the prefix $\tau^{\text{pre}}$ and suffix part $\tau^{\text{suf}}$.}
% \label{distance2}
% \end{figure}

%% file: MobileMicrorobotSystem.tex
%\begin{figure*}[h] 
%\centering
%\subfloat[]{\label{fig:coil1}\includegraphics[width=0.48\textwidth]{figures/coil11.png}}
%\hspace*{0.2cm}
%\subfloat[]{\label{fig:coil2}\includegraphics[width=0.48\textwidth]{figures/coil21.png}}
%\\
%\subfloat[]{\label{fig:coil3}\includegraphics[width=0.48\textwidth]{figures/coil31.png}}
%\hspace*{0.2cm}
%\subfloat[]{\label{fig:coil4}\includegraphics[width=0.48\textwidth]{figures/coil41.png}}

%\caption[]{Sequential operations of three robots to finish an assembly task: (a) Initial position of the robots and parts; (b) Robot 1 moves to manipulate the part 1 to the desired location, (b) Robot 1 retracts after placing the part 1 to the desired location while Robot 2 to take over the task, (c) Robot 2 and Robot 3 take the parts 1  and 2 to the desired positions, (d) Finally, Robot2 and Robot 3 finish the assembly task.}
%\label{fig:coil}
%\end{figure*}
\label{sec_intro}
\begin{figure}[h]
\centering
\includegraphics[width=1\columnwidth]{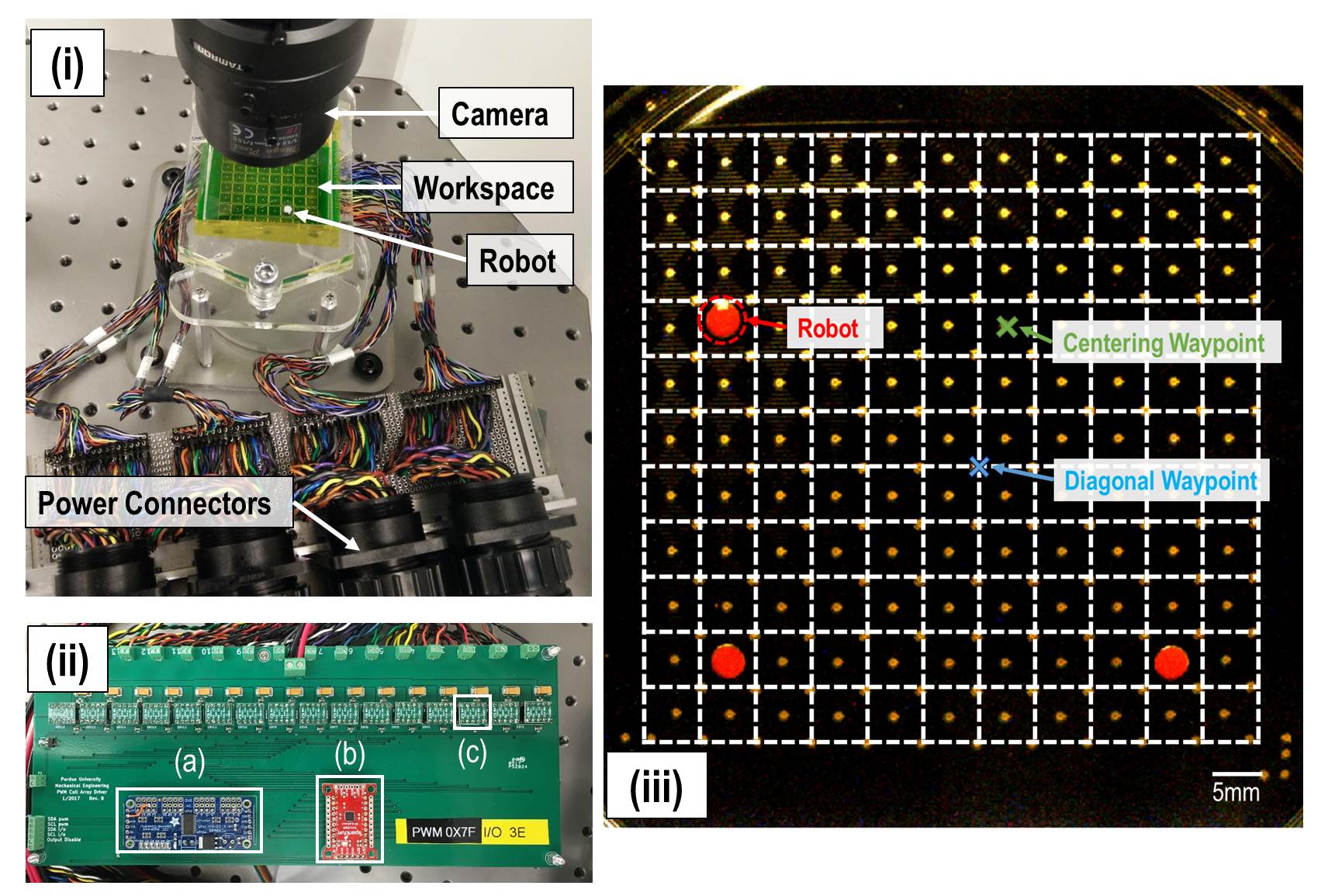}
\caption{(i) Mobile microrobot platform with the 8 $\times$ 8 planar coil array; (ii) Coil current controller unit: (a) PWM driver, (b) I/O Controller, and (c) Motor driver; (iii) Overhead view of the 11 $\times$ 11 coil array workspace and two $3.125$ mm diameter $\times$ $0.79$ mm thick magnetic robots.}
\label{fig:platform}
\end{figure}

We have developed two platforms to independently actuate multiple microrobots using local magnetic fields (Fig.~\ref{fig:platform}). The first platform is an $8$ $\times$ $8$ array of planar microcoils each with a dimension of $4.33$ mm $\times$ $4.33$ mm, shown in Fig.\ref{fig:platform}(i). Each planar microcoil has a winding width of $178$ $\mu m$, an out-of-plane winding thickness of $178$ $\mu m$, and a winding spacing of $178$ $\mu m$ with $5$ turns. The second platform has an $11$ $\times$ $11$ array of similarly sized coils is used for more advanced experiments with larger numbers of robots, shown in Fig.\ref{fig:platform}(iii). Each winding is rectangular in shape for the ease of fabrication.  Each coil is capable of generating a local magnetic field that is dominant only in its vicinity. The coils and the traces are separated by a thick insulation layer of approximately $1.5$ mm which provides sufficient isolation. A layer of Teflon or a thin glass cover slip is used as an insulation layer between the coils and the robots. The surface is also covered with a thin layer of silicone oil to reduce the friction between the robots and the workspace.

The current in each coil can be controlled through four to eight custom control units (Fig.~\ref{fig:platform}(ii)). Each control unit allows for the individual control of the magnitude and direction of current through sixteen coils. The direction of the current is switched using a voltage level shifter I/O SX1509 (Sparkfun Electronics), and the magnitude of the current is controlled by regulating the supply voltage using a PWM driver PCA9685 (Adafruit) into a motor controller DRV8838 (Pololu Corporation) which supplies the current to drive each coil. A set of $16$ coils are controlled using a single GPIO and PWM driver board, to which communications are sent over an I2C interface. The motor driver provides currents up to $1$ A to each coil, and the external current supply can supply up to $60$ A of current to the system. For a total of $64$ coils, we use $4$ controller boards and each unit is connected to the workspace using a circular power connector shown in Fig.\ref{fig:platform}(i). For the larger platform, a total of $121$ coils were controlled using $8$ controller boards. An Arduino Uno microcontroller is used to communicate with each controller board using I2C signals. The microcontroller is connected to a CPU with an Intel\textregistered Core\texttrademark i7-4771 (3.50 GHz) processor and 16 GB RAM. All computations are performed using Matlab\textregistered \hspace{0.05in} and the commands are sent to the controller boards through the Arduino microcontroller interface.

The robots used for the experiments are $3.125$ mm diameter and $0.79$ mm thick neodymium disc magnets. These magnets are of grade N52 and magnetized through their thickness. Multiple robots can move independently in this workspace provided they are approximately $15$ mm ($R_{I_{robot}}$) apart at all times. If they get closer than that, they repel/attract each other, affecting the robot's behavior if they are not in their equilibrium states. The motion of the robots are controlled by controlling the direction and magnitude of currents in the coils in the vicinity of the robot, as discussed in Section~\ref{sec:LocalMagneticFieldActuation}. 

The $45$ mm $\times$ $45$ mm view of the workspace is captured (Fig.\ref{fig:platform}(iii)) using an overhead CCD camera (FL3-U3-13E4C-C, FLIR Integrated Imaging Solutions Inc.). The possible waypoints in the workspace are plotted on the image and the robots are identified using background subtraction and Hough transform techniques. Each robot is also colored white or red to improve the image detection algorithm accuracy.  With these image processing algorithms and Matlab's Image Acquisition toolbox, the vision system is able to run at 13 frames per second.
%This camera with implemented vision algorithm is able to capture up to 13 frames per second using the Matlab Image Acquisition toolbox. 
%End effectors

%Figure~\ref{fig:coil} shows an assembly task with two parts accomplished by three robots. Every robot has an end effector that helps to push a part to a desired location (Figure~\ref{fig:coil1}). 
%The assembly task starts with the Robot 1 pushing the Part 1 to the desired location in the workspace. Robot 1 backs off after taking the part 1 to the desired location. Robot 2 then takes over the manipulation task by pushing the robot in towards the final destination where both parts will be assembled. At the same time, part 2 is pushed by the Robot 3 towards the desired location. Finally, both the parts are manipulated to finish the assembly task.

%In order to accomplish the assembly tasks  efficiently, there has to be a temporal correlation between the movements of the robots. The efficiency of the manipulation tasks can be greatly increased only with a temporal planning between the robots.  This paper will achieve the temporal planning between the robots for effective completion of assembly tasks.  

%% file: Experiments.tex
\begin{figure}
\centering
\includegraphics[width=1\columnwidth]{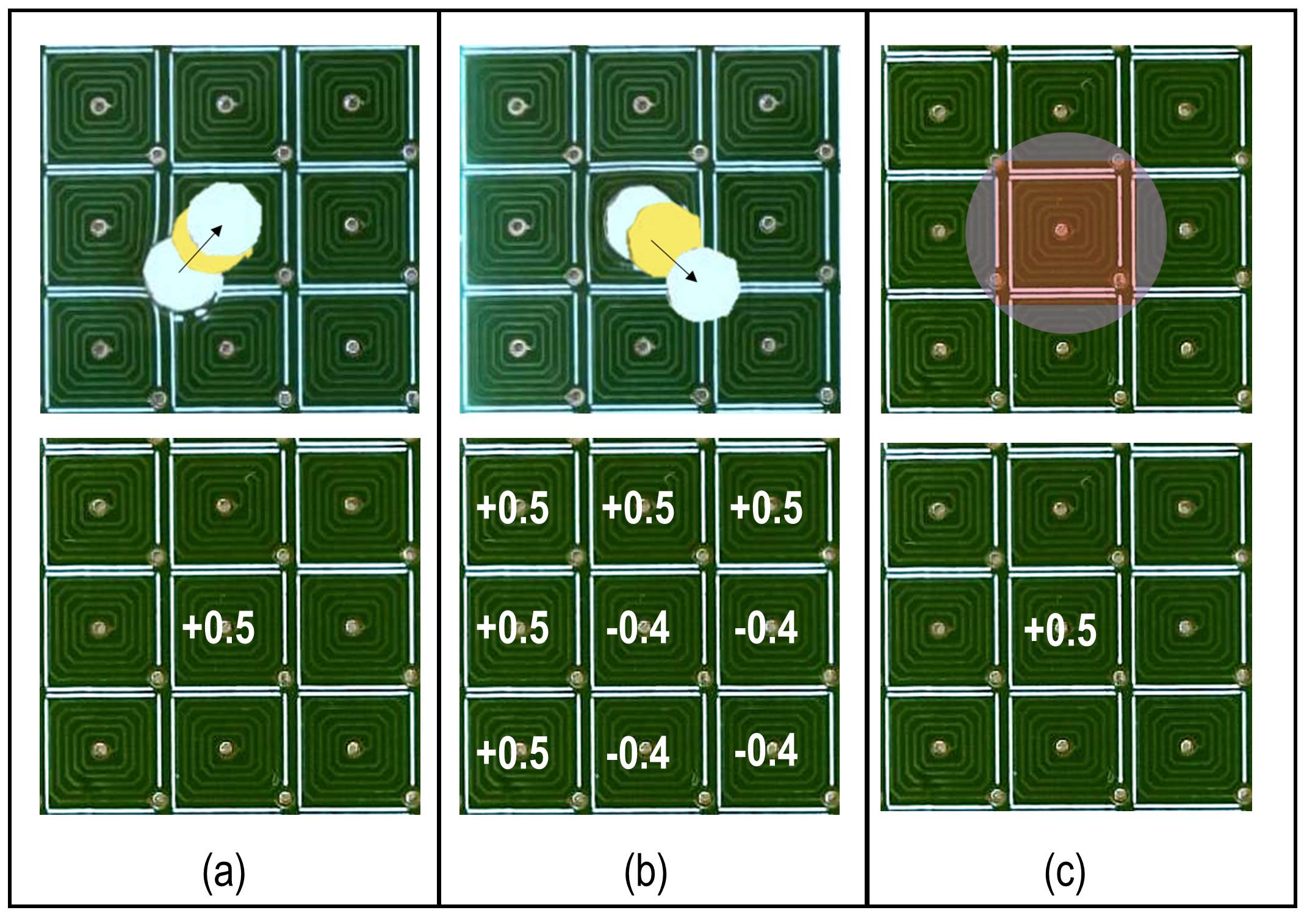}

\caption{Validation tests for (a) \textit{center} and (b) \textit{diagonal} moves. (c) $R_{I_{coil}}$ influence region for a coil. Red denotes region of attraction, and blue denotes region of repulsion for the current flowing in the clockwise direction. The direction ($+$ for attractive and $-$ for repulsive potential) and magnitude of the current in amps are shown below each moves.}
\label{fig:validation}
\end{figure}

The first set of experiments conducted were to validate the models of the system presented in Section~\ref{sec:LocalMagneticFieldActuation}.
It was confirmed that the magnetic robots have two types of static stable equilibrium points in the workspace: 1. the center of a coil when its attracting the robot, which we will refer to as the \textit{center}; and 2. the midpoint of the common diagonal of any four coils, which we refer to as the \textit{diagonal}. This was validated by manually perturbing the robot after it reaches the equilibrium point after either a center or diagonal move and checking if the robot returns to its equilibrium position (Fig.~\ref{fig:validation}(a),(b) and Supplemental Video 2). Also, the $R_{I_{coil}}$ influence region was measured for which the stable equilibrium is maintained for the task and is shown in Fig~\ref{fig:validation}(c).

The centering equilibrium point can qualitatively withstand larger perturbation forces because of the local strength of the field holding the robot in the center when compared to field applied to create a diagonal equilibrium point. The diagonal point is a result of four coils being activated which are pushing the robot away from each other. Although there are four coils, only eight traces are involved in actively pushing the robot, resulting in a weaker equilibrium point. The centering coil has sixteen traces pushing the robot to the center. This makes the diagonal equilibrium point more susceptible to errors from disturbances or uncertainties in the workspace. 

The applied magnetic force to the robots is directly proportional to the current in the coil. The current can vary from 0 to 1 A. The robot can move due to currents as low as  $2.5$ mA for a grade N52 disc magnet, but is slow and has a tendency to get stuck or deviate from its desired trajectory in the presence of external forces. At higher currents, the robots move faster to their equilibrium positions due to higher gradient fields and have better static equilibrium strength. However, the range $R_{I_{coil}}$ of the coils does not change as much with increased current due to the inverse square relationship of the field strength with the distance from the coil. 
Supplemental Video 3 shows examples of a robot moving from a diagonal waypoint to a center waypoint as well as the robot moving from a center waypoint to a diagonal waypoint.

Hence, the motion of the robot is well characterized for the wTS states used in the LTL code. The wTS states here are local equilibrium points generated from potential fields generated from the coil. The currents in the coils around the robot generates potentials which result in forces as described in Fig.\ref{fig:LocalFields}. If the robot does not reach the local equilibrium point at the point of feedback, the motion control keeps the coils switched on longer and waits for the robot to eventually reach the equilibrium state. The values of the currents used to reach the equilibrium states for the validation tests are shown at the bottom of Fig.~\ref{fig:validation}.

The robots themselves interact between each other due to their magnetization. This influence region between the robots, $R_{I_{robot}}$, can be reduced by using lower grade magnetic robots.  For example, our experiments showed that a N52 grade robot can have an $R_{I_{robot}}$ up to $20$ mm while a N42 grade robot has a $R_{I_{robot}}$ of $15$ mm.  Demagnetization of the robots can reduce their influence regions ($R_{I_{robot}}$) but it will also affect the performance of the robot since their influence from the surrounding coils, $R_{I_{coil}}$, is diminished due to their reduced magnetization.

% \begin{figure*}%[h]
% \centering
% \subfloat[LTL Experiment 1]{\includegraphics[width=0.40\textwidth]{ben_figures/2robot_march13_new.png}\label{LTLexp1}}\hspace{0.20in} 
% \subfloat[LTL Experiment 2]{\includegraphics[width=0.40\textwidth]{ben_figures/2robot_benji_new.png}\label{LTLexp2}}
% \caption{Screen shots of two multi-robot LTL experiments. In each experiment, two robots are moving in the workspace to reach the goal location marked with a $\star$, while avoiding an obstacle in workspace. For the prefix goal, Robot 1 goes towards the goal location while Robot 2 stays in place (i),(ii). Next, as suffix motion, Robot 1 returns to it's original position and Robot 2 moves towards the goal location (iii),(iv). Robot 2 then goes back to its original position and robot 1 moves to the goal location (v),(vi). For these tasks, the suffix motion then repeats infinitely. \textcolor{red}{Benji:  Keep LTL Exp. 2; add in similar figure for 3 robot experiment.}}
% \label{fig:LTLexps}
% \end{figure*}

\begin{figure}%[h]
\centering
%{\includegraphics[width=0.40\textwidth]{ben_figures/2robot_benji_new.png}\label{LTLexp2}}
{\includegraphics[width=0.40\textwidth]{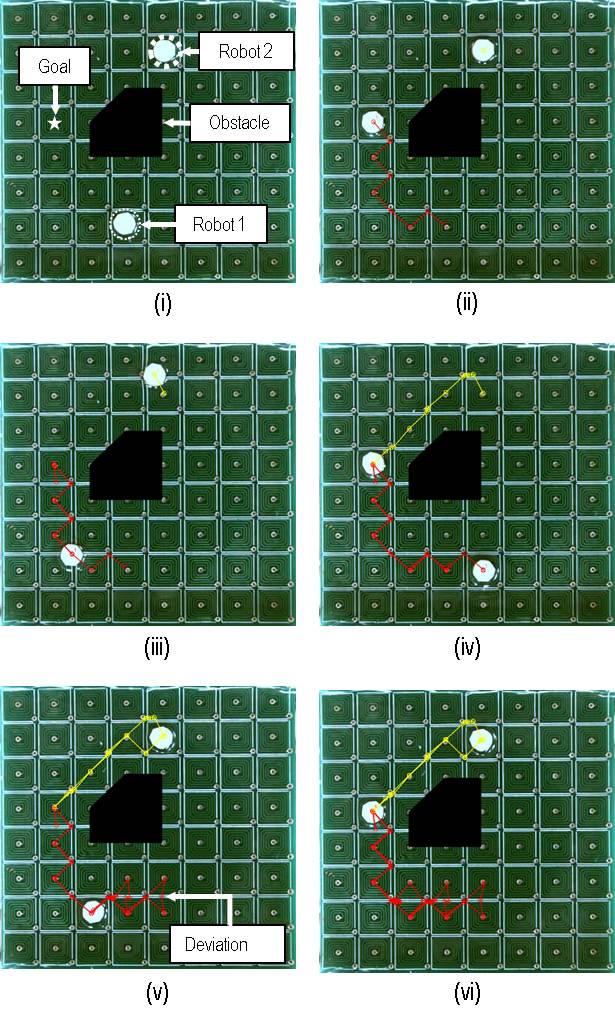}\label{LTLexp2}}
\caption{Two-robot LTL experiment: Two robots move in the workspace to reach the goal location marked with a $\star$, while avoiding an obstacle in workspace. For the prefix goal, Robot 1 goes towards the goal location while Robot 2 stays in place (i),(ii). Next, as suffix motion, Robot 1 returns to it's original position and Robot 2 moves towards the goal location (iii),(iv). Robot 2 then goes back to its original position and robot 1 moves to the goal location (v),(vi). For this task, the suffix motion then repeats infinitely. }% \textcolor{red}{Benji: move indexing labels either below figures or into top left corner of each image.}  }
\label{fig:LTLexps}
\vspace{-0.10in}
\end{figure}

% \begin{figure}[h]
% \includegraphics[width=1\columnwidth]{ben_figures/2robot_benji_new.png}
% \caption{Screen shots of two robots moving in the workspace trying to reach the goal location at $\star$, while avoiding the obstacle. For the prefix goal, robot 1 goes towards the goal location while robot 2 stays in place (a),(b). Next, as suffix motion, the robot 1 returns to it's original position and, robot 2 moves towards the goal location (c),(d). Robot 2 then goes back to it's original position and robot 1 moves to the goal location (e),(f). For the task, the suffix motion then repeats infinitely. Here the robot 2 moves diagonal during its motion as it is the shortest path.}
% \label{fig:2robot_benji}
% \end{figure}

\subsection{Multi-Robot LTL Experiments}
An experiment of two robots moving in the workspace using the LTL planning algorithm presented in Sec.~\ref{sec:TemporalPlanning} is illustrated in Fig.~\ref{fig:LTLexps} and in Supplemental Video 4. Here, we consider two robots where the assigned LTL-based task is

\begin{align}\label{ltlexp}
\phi_{\text{task}}=&(\square\Diamond\pi_1^0)\wedge (\square\Diamond\pi_1^{*})\wedge(\square\Diamond\pi_2^0)\wedge\nonumber\\& (\square\Diamond\pi_2^{*})\wedge (\square (\neg\pi_{\text{obs}})),
\end{align}
where the atomic propositions $\pi_1^{*}$ and $\pi_2^{*}$ are true if robot 1 and 2 are in the location denoted by $\star$ in Figure \ref{fig:LTLexps}, respectively. Similarly, the atomic propositions $\pi_1^0$ and $\pi_2^0$ are true if robot 1 and 2 are in the initial locations. In words the LTL formula \eqref{ltlexp} requires both robots to (a) visit the location $\star$ infinitely often, (b) visit their respective starting positions infinitely often, and (c) avoid the obstacles in the workspace. Also, the robots need to satisfy the constraints imposed by $\phi_c$ defined in \eqref{eq:collision}.  In Fig.~\ref{fig:LTLexps}, the prefix and suffix part of the synthesized plan follow: Prefix: (i) Robot 1 moves towards the $\star$ goal location while Robot 2 stays in its position. (ii) Robot 1 reaches the goal location and the prefix is completed. The suffix parts starts from (ii), (iii) Next, robot 1 returns to its starting position while Robot 2 moves towards the $\star$ position. (iv) Robot 1 reaches its starting position and Robot 2 reaches the target goal location at $\star$. (v) Robot 1 goes back to the $\star$ position while Robot 2 goes back to its original position. (vi) The robots are again in the initial suffix state shown in (ii) and the loop continues.
Note that the algorithm dictates Robot 2 to move along a diagonal trajectory in order to travel along the shortest path to the goal. By observing the experimental trajectories in the videos, it is also confirmed that the minimum robot spacing distance during each experiment is greater than the prescribed value (constraint) of $R_{I_{robot}}$.

\begin{figure}[h]
\includegraphics[width=1\columnwidth]{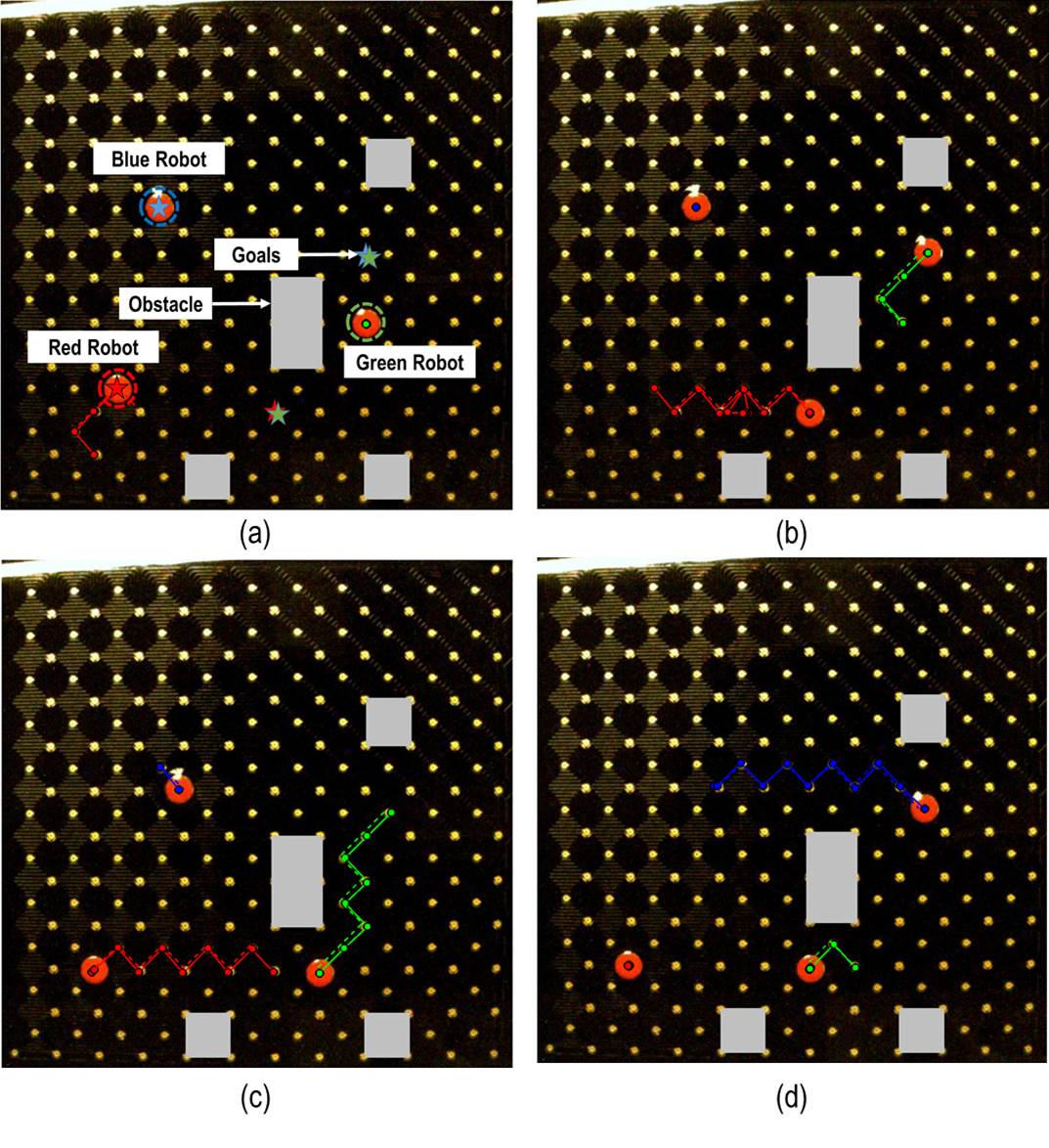}
\caption{Prefix motion of the three robot experiment from Case Study II in Sec.\ref{sec:three_robot_study}. (a),(b),(c), and (d) represent the exact states represented in Fig.\ref{sim2pre}.  The paths of each robot are represented by their color, while the goals are shown as $\star$ in (a). }%\textcolor{red}{I think we are missing blue star at 24 here; also would be good to add both green and red stars at 88.} }
\label{fig:3robot_exp_prefix}
\vspace{-0.15in}
\end{figure}

\begin{figure}[h]
\includegraphics[width=1\columnwidth]{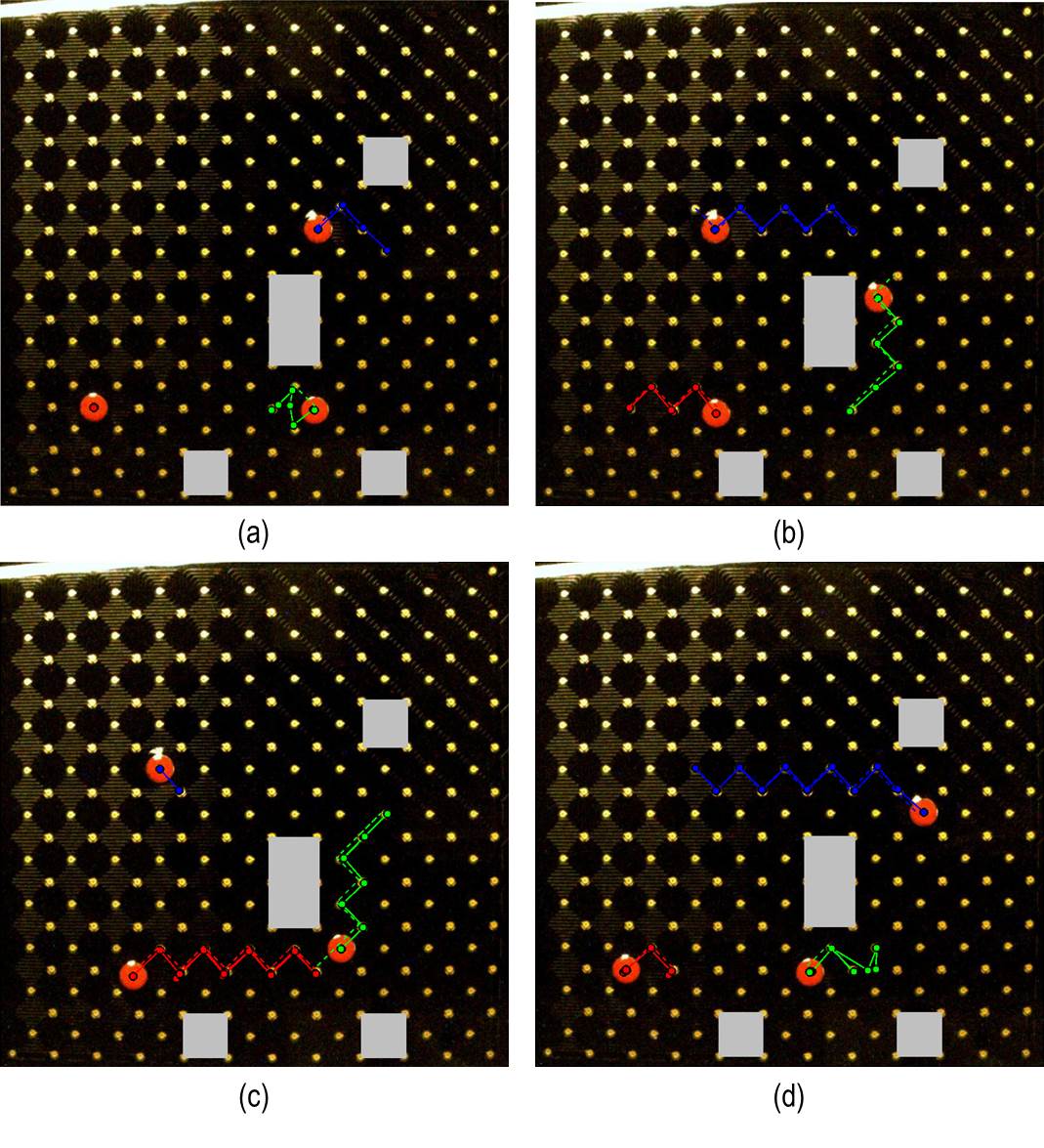}
\caption{Suffix motion of the three Robot Experiment from Case Study II in Sec.\ref{sec:three_robot_study}. (a),(b),(c), and (d) represent the exact states represented in Fig.\ref{sim2suf}.The paths of each robot are represented by their color, while the goals are shown as $\star$ in \ref{fig:3robot_exp_prefix}(a).}
\label{fig:3robot_exp_suffix}
\vspace{-0.15in}
\end{figure}

An experiment of three robots moving in a workspace as in the simulation study in Sec.\ref{sec:three_robot_study} is illustrated in Fig.~\ref{fig:3robot_exp_prefix} and Fig.~\ref{fig:3robot_exp_suffix} and Supplemental Video 4. The three robots in the workspace are represented by the colors red(r), blue(b), and green(g). The goal locations $88$ and $42$ are shown as red color $\star$, $88$ and $25$ are shown as green color $\star$, and $24$ and $12$ are shown as blue color $\star$ in Fig.~\ref{fig:3robot_exp_prefix}(a). The LTL based task is described in \eqref{eq:ex2}.
%
% \textcolor{blue}{\begin{align}\label{eq:ex2_exp}
% \phi_{\text{task}}=&(\square\Diamond \pi_{r}^{{\bf{c}}_{42}})
% \wedge (\square\Diamond \pi_{r}^{{\bf{c}}_{88}})
% \wedge (\square\Diamond \pi_{g}^{{\bf{c}}_{88}})
% \wedge (\square\Diamond \pi_{g}^{{\bf{c}}_{24}})\nonumber\\
% &\wedge (\square\Diamond \pi_{b}^{{\bf{c}}_{24}})
% \wedge (\square\Diamond \pi_{b}^{{\bf{c}}_{12}})
% \wedge (\square (\neg \pi_{\text{obs}})).
% \end{align}}
%
In words, this LTL-based task requires: 
(a) the red robot to move back and forth between locations ${\bf{c}}_{88}$ and ${\bf{c}}_{42}$ infinitely often, (b) the green robot to  move back and forth between locations ${\bf{c}}_{88}$ and ${\bf{c}}_{24}$ infinitely often, (c) the blue robot to move back and forth between locations ${\bf{c}}_{24}$ and ${\bf{c}}_{12}$ infinitely often, and (d) all robots to avoid the obstacles in the workspace. Fig.~\ref{fig:3robot_exp_prefix} and Fig.~\ref{fig:3robot_exp_suffix} show the prefix and suffix at the same states as shown in Case Study II in Sec.~\ref{sec:three_robot_study}, Fig.~\ref{sim2pre} and Fig.~\ref{sim2suf}, respectively.  From the experimental trajectories, it is confirmed that the robots maintained the minimum spacing distance while avoiding the obstacles to reach the respective goals.  These results are also shown in Supplemental Video 4.

\begin{figure*}[h]
\centering
\includegraphics[width=.8\textwidth]{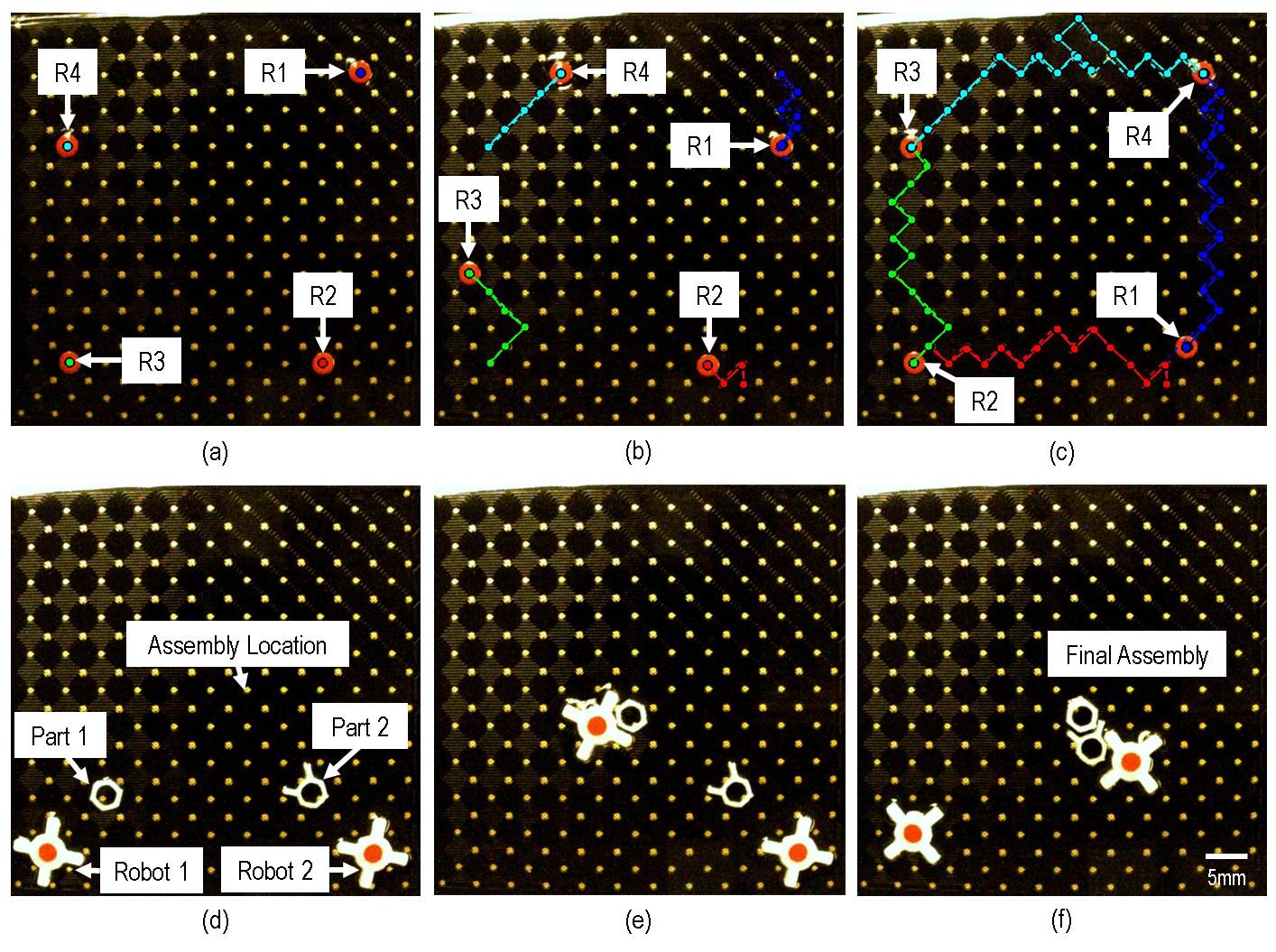}
\caption{Advanced multi-robot experiments:  (a)-(c) Four-robot experiment where the robots R1, R2, R3, and R4 move clockwise to the starting location of the next robot. The paths of each robot are represented in different colors. Two robot assembly task: (d) each robot is fitted with a 3D printed end-effector to push other parts. (e) Robot 1 pushing Part 1 to the assembly location. (f) Robot 2 pushing Part 2 to the assembly location and ensuring correct orientation of part.}
%\textcolor{red}{Label each robot in (a) as R1, R2, R3, R4. Legend in (b) hard to read, just label stuff with text-boxes and arrows.} }
\label{fig:4robot_exp_assembly}
\end{figure*}

\subsection{Advanced Multi-Robot Experiments}
% \textcolor{red}{Benji: insert section here describing the 4 robot experiment and microassembly experiment. It might be best to integrate this into the Discussion section.  Once is written we can see where it fits best.}
A series of advanced multi-robot experiments have been conducted to explore the capabilities of the system and an actual assembly task.  The results of some of these are shown in Fig.~\ref{fig:4robot_exp_assembly}. First, we consider an  experiment of four robots moving independently in the workspace. The task is for the four robots to move clockwise to the starting location of the next robot. The starting positions, intermediate, and final paths of the robots are shown in Fig.~\ref{fig:4robot_exp_assembly}(a),(b), and (c), respectively.

The ability of this system to conduct an assembly task has also been demonstrated, as shown in Fig.~\ref{fig:4robot_exp_assembly}(d)-(f). In this experiment, two robots fitted with 3D printed end-effectors were used to assemble two hexagonal parts. The end effector and the assembly parts were both printed using the Connex 350(Stratasys Ltd.). printer with Vero White material. The robots were able to sequentially push the parts to the assembly location. The paths for the robots were specified on the basis of the positions of the parts. Here, this system cannot control the orientation of the robot, however position control was sufficient to position and orient the parts to form the required assembly. Supplemental Video 4 also shows the results from these advanced multi-robot experiments.

%\textcolor{red}{Can we have a plot of minimum distances similar to figure 9 based on experimental data? }

\subsection{Discussion}
The robots deviate from in their planned paths at times during the experiments, as highlighted for example in Fig.~\ref{fig:LTLexps}. This can be partially attributed to the external forces on the robot due to other robots present in the arena even though they maintain a distance greater than $R_{I_{robot}}$. The original $R_{I_{robot}}$ was determined while the robots were at rest experiencing static friction. Once the robots start moving, the $R_{I_{robot}}$ value increases due to the presence of dynamic friction, that is lower than the static friction, under the robots. Additionally, small differences between the coils due to fabrication limitations also contribute to errors. In the case of path deviations, as highlighted in Fig.~\ref{fig:LTLexps}, the feedback control, which runs at 3Hz, is designed to bring the robot back to its desired waypoint. Another potential cause for errors is the assumption that the net magnetization direction is through the center of the robots; a small offset in the net magnetization can affect the position of the robot at the equilibrium. There can also be a significant impact to the image processing based tracking accuracy due to small changes in the workspace which can affect the background subtraction algorithm. Therefore, the tolerance for reaching a goal location is fixed at $0.5$ mm. It should also be noted that currently
%the coils cannot exert magnetic torques that allow the orientation control of the robot. Hence, 
the closed loop path involves only position control of the robot and not orientation control at this time.  Nevertheless, this type of multi-robot planning and control will be useful in future micromanipulation applications, such as those that require efficient, sequential, and cyclic assembly of microscale components.
 
%\textcolor{red}{Three robot experiment to show why it didn't work}

%% file: Conclusion.tex
%In conclusion, we have presented a global LTL formulas as

In this paper we have presented a novel control framework for teams of magnetic microrobots that need to accomplish temporal micromanipulation tasks.
An optimal control synthesis algorithm was developed to construct discrete plans that satisfy the assigned tasks. The key idea was to combine an existing optimal control synthesis approach with a novel method to reduce the state-space of the transition system that models robot mobility, so that our planning algorithm scales to larger planning problems.
Two case studies were presented in simulation to show the efficacy of this approach in terms of its ability to handle large-scale planning problems that cannot be solved using existing optimal control techniques.  A new diagonal static equilibrium point in the local magnetic field control substrate was identified. Control validation experiments for the workspace static equilibrium points were presented followed by experiments with a team of two magnetic microrobots executing motion plans from the developed algorithm.  This is the first experimental demonstration of LTL-based multi-microrobot control using local magnetic fields.  We showed that the robots are able to satisfy the LTL specifications while also respecting the physical constraints for inter-robot spacing in the workspace.  This is the first step in using large teams of independently controlled magnetic microrobots for efficient, temporal micromanipulation task planning. 

%% file: main.bbl
\begin{thebibliography}{10}
\providecommand{\url}[1]{#1}
\csname url@samestyle\endcsname
\providecommand{\newblock}{\relax}
\providecommand{\bibinfo}[2]{#2}
\providecommand{\BIBentrySTDinterwordspacing}{\spaceskip=0pt\relax}
\providecommand{\BIBentryALTinterwordstretchfactor}{4}
\providecommand{\BIBentryALTinterwordspacing}{\spaceskip=\fontdimen2\font plus
\BIBentryALTinterwordstretchfactor\fontdimen3\font minus
  \fontdimen4\font\relax}
\providecommand{\BIBforeignlanguage}[2]{{%
\expandafter\ifx\csname l@#1\endcsname\relax
\typeout{** WARNING: IEEEtran.bst: No hyphenation pattern has been}%
\typeout{** loaded for the language `#1'. Using the pattern for}%
\typeout{** the default language instead.}%
\else
\language=\csname l@#1\endcsname
\fi
#2}}
\providecommand{\BIBdecl}{\relax}
\BIBdecl

\bibitem{Capp14}
D.~Cappelleri, D.~Efthymiou, A.~Goswami, N.~Vitoroulis, and M.~Zavlanos,
  ``Towards mobile microrobot swarms for additive micromanufacturing,''
  \emph{International Journal of Advanced Robotic Systems}, vol.~11, no. 150,
  2014, dOI: 10.5772/58985.

\bibitem{Chow15}
S.~Chowdhury, W.~Jing, P.~Jaron, and D.~Cappelleri, ``Path planning and control
  for autonomous navigation of single and multiple magnetic mobile
  microrobots,'' in \emph{Proceedings of the ASME 2015 International Design
  Engineering Technical Conferences \& Computers and Informatio in Engineering
  Conference}, Boston, Massachusetts, USA, August 2-5 2015.

\bibitem{Chow15a}
S.~Chowdhury, W.~Jing, and D.~J. Cappelleri, ``Towards independent control of
  multiple magnetic mobile microrobots,'' \emph{Micromachines}, vol.~7, no.~1,
  p.~3, 2015.

\bibitem{kantaros15asilomar}
Y.~Kantaros and M.~M. Zavlanos, ``Intermittent connectivity control in mobile
  robot networks,'' in \emph{49th Asilomar Conference on Signals, Systems and
  Computers}, Pacific Grove, CA, USA, November, 2015, pp. 1125--1129.

\bibitem{david1}
D.~J. Cappelleri and Z.~Fu, ``Towards flexible, automated microassembly with
  caging micromanipulation,'' in \emph{2013 IEEE International Conference on
  Robotics and Automation}, May 2013, pp. 1427--1432.

\bibitem{seven}
D.~Cappelleri, C.~Peng, J.~Fink, B.~Gavrea, and V.~Kumar, ``Automated assembly
  for meso-scale parts,'' \emph{IEEE Transactions on Automation Science and
  Engineering}, vol.~8, no.~3, pp. 598--613, July 2011.

\bibitem{WasonTRO12}
J.~Wason, J.~Wen, J.~Gorman, and N.~Dagalakis, ``Automated multiprobe
  microassembly using vision feedback,'' \emph{Robotics, IEEE Transactions on},
  vol.~28, no.~5, pp. 1090 --1103, oct. 2012.

\bibitem{cappelleri2012cooperative}
D.~J. Cappelleri and Z.~Fu, ``Cooperative micromanipulators for 3d
  micromanipulation and assembly,'' in \emph{ASME 2012 International Design
  Engineering Technical Conferences and Computers and Information in
  Engineering Conference}.\hskip 1em plus 0.5em minus 0.4em\relax American
  Society of Mechanical Engineers, 2012, pp. 177--185.

\bibitem{cappelleri2011two}
D.~J. Cappelleri, G.~Piazza, and V.~Kumar, ``A two dimensional vision-based
  force sensor for microrobotic applications,'' \emph{Sensors and Actuators A:
  Physical}, vol. 171, no.~2, pp. 340--351, 2011.

\bibitem{cappelleri2011caging}
D.~J. Cappelleri, M.~Fatovic, and Z.~Fu, ``Caging grasps for micromanipulation
  \& microassembly,'' in \emph{Intelligent Robots and Systems (IROS), 2011
  IEEE/RSJ International Conference on}.\hskip 1em plus 0.5em minus 0.4em\relax
  IEEE, 2011, pp. 925--930.

\bibitem{YangNelsonGaines2005}
G.~Yang, J.~A. Gaines, and B.~J. Nelson, ``Optomechatronic design of
  microassembly systems for manufacturing hybrid microsystems,'' \emph{IEEE
  Transactions on Industrial Electronics}, vol.~52, no.~4, pp. 1013--1023, Aug
  2005.

\bibitem{Dechev04}
N.~Dechev, W.~Cleghorn, and J.~Mills, ``Microassembly of 3-d microstructures
  using a compliant, passive microgripper,'' \emph{Microelectromechanical
  Systems, Journal of}, vol.~13, no.~2, pp. 176 -- 189, april 2004.

\bibitem{Popa07}
D.~Popa, W.~H. Lee, R.~Murthy, A.~Das, and H.~Stephanou, ``High yield automated
  mems assembly,'' in \emph{Automation Science and Engineering, 2007. CASE
  2007. IEEE International Conference on}, sept. 2007, pp. 1099 --1104.

\bibitem{Ren07}
L.~Ren, L.~Wang, J.~Mills, and D.~Sun, ``3-d automatic microassembly by
  vision-based control,'' in \emph{Intelligent Robots and Systems, 2007. IROS
  2007. IEEE/RSJ International Conference on}, 29 2007-nov. 2 2007, pp. 297
  --302.

\bibitem{Rabenorosoa09}
K.~Rabenorosoa, C.~Clevy, P.~Lutz, S.~Bargiel, and C.~Gorecki, ``A
  micro-assembly station used for 3d reconfigurable hybrid moems assembly,'' in
  \emph{Assembly and Manufacturing, 2009. ISAM 2009. IEEE International
  Symposium on}, nov. 2009, pp. 95 --100.

\bibitem{Tamadazte09a}
B.~Tamadazte, N.~Le~Fort-Piat, S.~Dembele, and E.~Marchand, ``Microassembly of
  complex and solid 3d mems by 3d vision-based control,'' in \emph{Intelligent
  Robots and Systems, 2009. IROS 2009. IEEE/RSJ International Conference on},
  oct. 2009, pp. 3284 --3289.

\bibitem{Venkatesan17}
V.~Venkatesan and D.~Cappelleri, ``Development of an automated flexible
  micro-soldering station,'' in \emph{Proceedings of the ASME 2017
  International Design Engineering Technical Conferences \& Computers and
  Information in Engineering Conference}, August 2017, accepted for
  publication.

\bibitem{Chow15b}
S.~Chowdhury, W.~Jing, and D.~J. Cappelleri, ``Controlling multiple
  microrobots: recent progress and future challenges,'' \emph{Journal of
  Micro-Bio Robotics}, vol.~10, no. 1-4, pp. 1--11, 2015.

\bibitem{Bane12}
A.~G. Banerjee, S.~Chowdhury, W.~Losert, and S.~K. Gupta, ``Real-time path
  planning for coordinated transport of multiple particles using optical
  tweezers,'' \emph{{IEEE Trans. Autom. Sci. Eng.}}, vol.~9, no.~4, pp. 669
  --678, Oct. 2012.

\bibitem{Bane14}
A.~Banerjee, S.~Chowdhury, and S.~Gupta, ``Optical tweezers: Autonomous robots
  for the manipulation of biological cells,'' \emph{Robotics Automation
  Magazine, IEEE}, vol.~21, no.~3, pp. 81--88, Sept 2014.

\bibitem{Chea15}
C.~C. Cheah, Q.~M. Ta, and R.~Haghighi, ``Robotic manipulation of a biological
  cell using multiple optical traps,'' in \emph{Robotics and Automation (ICRA),
  2015 IEEE International Conference on}, May 2015, pp. 803--808.

\bibitem{Chea15a}
U.~K. Cheang and M.~J. Kim, ``Self-assembly of robotic micro-and nanoswimmers
  using magnetic nanoparticles,'' \emph{Journal of Nanoparticle Research},
  vol.~17, no.~3, pp. 1--11, 2015.

\bibitem{Chow13b}
S.~Chowdhury, P.~Svec, C.~Wang, K.~Seale, J.~P. Wikswo, W.~Losert, and S.~K.
  Gupta, ``Automated cell transport in optical tweezers assisted microfluidic
  chamber,'' \emph{{IEEE Trans. Autom. Sci. Eng.}}, vol.~{10}, no.~{4}, pp.
  {980--989}, Oct. 2013.

\bibitem{Chow13a}
S.~Chowdhury, A.~Thakur, C.~Wang, P.~Svec, W.~Losert, and S.~K. Gupta,
  ``Automated manipulation of biological cells using gripper formations
  controlled by optical tweezers,'' \emph{{IEEE Trans. Autom. Sci. Eng.}},
  vol.~{11}, no.~{2}, pp. {338--347}, Apr. 2014.

\bibitem{Haoyao12}
C.~Haoyao and D.~Sun, ``Moving groups of microparticles into array with a
  robot-tweezers manipulation system,'' \emph{IEEE Trans. Robot.}, vol.~28,
  no.~5, pp. 1069--1080, Oct. 2012.

\bibitem{Hu11b}
S.~Hu and D.~Sun, ``Automated transportation of biological cells with a
  robot-tweezer manipulation system,'' \emph{Int. J. Robot. Res.}, vol.~30,
  no.~14, pp. 1681--1694, Dec. 2011.

\bibitem{Hu12}
W.~Hu, K.~S. Ishii, and A.~Ohta, ``Micro-assembly using optically controlled
  bubble microrobots in saline solution,'' in \emph{Robotics and Automation
  (ICRA), 2012 IEEE International Conference on}, May 2012, pp. 733--738.

\bibitem{Rahm17}
M.~A. Rahman, J.~Cheng, Z.~Wang, and A.~T. Ohta, ``Cooperative
  micromanipulation using the independent actuation of fifty microrobots in
  parallel,'' \emph{Scientific Reports}, vol.~7, 2017.

\bibitem{Thak14}
A.~Thakur, S.~Chowdhury, P.~{\v{S}}vec, C.~Wang, W.~Losert, and S.~K. Gupta,
  ``Indirect pushing based automated micromanipulation of biological cells
  using optical tweezers,'' \emph{The International Journal of Robotics
  Research}, p. 0278364914523690, 2014.

\bibitem{Mair17}
L.~O. Mair, A.~Nacev, R.~Hilaman, P.~Y. Stepanov, S.~Chowdhury, S.~Jafari,
  J.~Hausfeld, A.~J. Karlsson, M.~E. Shirtliff, B.~Shapiro \emph{et~al.},
  ``Biofilm disruption with rotating microrods enhances antimicrobial
  efficacy,'' \emph{Journal of Magnetism and Magnetic Materials}, vol. 427, pp.
  81--84, 2017.

\bibitem{Mair17a}
L.~Mair, B.~Evans, A.~Nacev, P.~Stepanov, R.~Hilaman, S.~Chowdhury, S.~Jafari,
  W.~Wang, B.~Shapiro, and I.~Weinberg, ``Magnetic microkayaks: propulsion of
  microrods precessing near a surface by kilohertz frequency, rotating magnetic
  fields,'' \emph{Nanoscale}, vol.~9, no.~10, pp. 3375--3381, 2017.

\bibitem{Jing14}
W.~Jing and D.~J. Cappelleri, ``Towards functional mobile magnetic
  microrobots,'' in \emph{Small-Scale Robotics. From Nano-to-Millimeter-Sized
  Robotic Systems and Applications}.\hskip 1em plus 0.5em minus 0.4em\relax
  Springer, 2014, pp. 81--100.

\bibitem{ufsmm_tase}
W.~Jing, S.~Chowdhury, M.~Guix, J.~Wang, Z.~An, B.~V. Johnson, and D.~J.
  Cappelleri, ``A microforce-sensing mobile microrobot for automated
  micromanipulation tasks,'' \emph{IEEE Transactions on Automation Science and
  Engineering}, pp. 1--13, 2018.

\bibitem{Jing14b}
\BIBentryALTinterwordspacing
W.~Jing and D.~Cappelleri, ``A magnetic microrobot with in situ force sensing
  capabilities,'' \emph{Robotics}, vol.~3, no.~2, pp. 106--119, 2014. [Online].
  Available: \url{http://www.mdpi.com/2218-6581/3/2/106}
\BIBentrySTDinterwordspacing

\bibitem{bi2018design}
C.~Bi, M.~Guix, B.~V. Johnson, W.~Jing, and D.~J. Cappelleri, ``Design of
  microscale magnetic tumbling robots for locomotion in multiple environments
  and complex terrains,'' \emph{Micromachines}, vol.~9, no.~2, p.~68, 2018.

\bibitem{Saka14}
M.~Selman~Sakar \emph{et~al.}, ``Cooperative manipulation and transport of
  microobjects using multiple helical microcarriers,'' \emph{RSC Advances},
  vol.~4, no.~51, pp. 26\,771--26\,776, 2014.

\bibitem{Dill12}
E.~Diller, S.~Floyd, C.~Pawashe, and M.~Sitti, ``Control of multiple
  heterogeneous magnetic microrobots in two dimensions on nonspecialized
  surfaces,'' \emph{IEEE Transactions on Robotics}, vol.~28, no.~1, pp.
  172--182, 2012.

\bibitem{Frut10}
D.~R. Frutiger, K.~Vollmers, B.~E. Kratochvil, and B.~J. Nelson, ``Small, fast,
  and under control: wireless resonant magnetic micro-agents,'' \emph{The
  International Journal of Robotics Research}, vol.~29, no.~5, pp. 613--636,
  2010.

\bibitem{Chea14}
U.~K. Cheang, K.~Lee, A.~A. Julius, and M.~J. Kim, ``Multiple-robot drug
  delivery strategy through coordinated teams of microswimmers,'' \emph{Applied
  Physics Letters}, vol. 105, no.~8, p. 083705, 2014.

\bibitem{Devo09}
D.~DeVon and T.~Bretl, ``Control of many robots moving in the same direction
  with different speeds: a decoupling approach,'' in \emph{American Control
  Conference, 2009. ACC'09.}\hskip 1em plus 0.5em minus 0.4em\relax IEEE, 2009,
  pp. 1794--1799.

\bibitem{Wong15}
D.~Wong, J.~Wang, E.~Steager, and V.~Kumar, ``Control of multiple magnetic
  micro robots,'' in \emph{Proceedings of the ASME 2015 International Design
  Engineering Technical COnferences \& Computers and Information in Engineering
  Conference}, Boston, Massachusetts, USA, August 2 -5 2015.

\bibitem{Pawa09a}
C.~Pawashe, S.~Floyd, and M.~Sitti, ``Multiple magnetic microrobot control
  using electrostatic anchoring,'' \emph{Applied Physics Letters}, vol.~94,
  no.~16, pp. 164\,108--164\,108, 2009.

\bibitem{Pelr12}
R.~Pelrine, A.~Wong-Foy, B.~McCoy, D.~Holeman, R.~Mahoney, G.~Myers, J.~Herson,
  and T.~Low, ``Diamagnetically levitated robots: An approach to massively
  parallel robotic systems with unusual motion properties,'' in \emph{2012 IEEE
  International Conference on Robotics and Automation (ICRA),}, 2012, pp.
  739--744.

\bibitem{Chow16}
S.~Chowdhury, W.~Jing, and D.~J. Cappelleri, ``Independent actuation of
  multiple microrobots using localized magnetic fields,'' in
  \emph{Manipulation, Automation and Robotics at Small Scales (MARSS),
  International Conference on}.\hskip 1em plus 0.5em minus 0.4em\relax IEEE,
  2016, pp. 1--6.

\bibitem{Chow17}
S.~Chowdhury, B.~V. Johnson, W.~Jing, and D.~J. Cappelleri, ``Designing local
  magnetic fields and path planning for independent actuation of multiple
  mobile microrobots,'' \emph{Journal of Micro-Bio Robotics}, vol.~12, no. 1-4,
  pp. 21--31, 2017.

\bibitem{kress2009temporal}
H.~Kress-Gazit, G.~E. Fainekos, and G.~J. Pappas, ``Temporal-logic-based
  reactive mission and motion planning,'' \emph{IEEE Transactions on Robotics},
  vol.~25, no.~6, pp. 1370--1381, 2009.

\bibitem{kress2007s}
------, ``Where's waldo? sensor-based temporal logic motion planning,'' in
  \emph{Robotics and Automation, 2007 IEEE International Conference on}.\hskip
  1em plus 0.5em minus 0.4em\relax IEEE, 2007, pp. 3116--3121.

\bibitem{bhatia2010sampling}
A.~Bhatia, L.~E. Kavraki, and M.~Y. Vardi, ``Sampling-based motion planning
  with temporal goals,'' in \emph{International Conference on Robotics and
  Automation (ICRA)}, Anchorage, AL, May 2010, pp. 2689--2696.

\bibitem{chen2011synthesis}
Y.~Chen, X.~C. Ding, and C.~Belta, ``Synthesis of distributed control and
  communication schemes from global {LTL} specifications,'' in \emph{50th IEEE
  Conference on Decision and Control and European Control Conference}, Orlando,
  FL, USA, December 2011, pp. 2718--2723.

\bibitem{chen2012formal}
Y.~Chen, X.~C. Ding, A.~Stefanescu, and C.~Belta, ``Formal approach to the
  deployment of distributed robotic teams,'' \emph{IEEE Transactions on
  Robotics}, vol.~28, no.~1, pp. 158--171, 2012.

\bibitem{baier2008principles}
C.~Baier and J.-P. Katoen, \emph{Principles of model checking}.\hskip 1em plus
  0.5em minus 0.4em\relax MIT press Cambridge, 2008, vol. 26202649.

\bibitem{clarke1999model}
E.~M. Clarke, O.~Grumberg, and D.~Peled, \emph{Model checking}.\hskip 1em plus
  0.5em minus 0.4em\relax MIT press, 1999.

\bibitem{smith2010optimal}
S.~L. Smith, J.~Tumova, C.~Belta, and D.~Rus, ``Optimal path planning under
  temporal logic constraints,'' in \emph{International Conference on
  Intelligent Robots and Systems (IROS)}.\hskip 1em plus 0.5em minus
  0.4em\relax Taipei, Taiwan: IEEE, October 2010, pp. 3288--3293.

\bibitem{smith2011optimal}
------, ``Optimal path planning for surveillance with temporal-logic
  constraints,'' \emph{The International Journal of Robotics Research},
  vol.~30, no.~14, pp. 1695--1708, 2011.

\bibitem{guo2013revising}
M.~Guo, K.~H. Johansson, and D.~V. Dimarogonas, ``Revising motion planning
  under linear temporal logic specifications in partially known workspaces,''
  in \emph{IEEE International Conference on Robotics and Automation (ICRA)},
  Karlsruhe, Germany, May 2013, pp. 5025--5032.

\bibitem{guo2013reconfiguration}
M.~Guo and D.~V. Dimarogonas, ``Reconfiguration in motion planning of
  single-and multi-agent systems under infeasible local {LTL} specifications,''
  in \emph{IEEE 52nd Annual Conference on Decision and Control (CDC)},
  Florence, December 2013, pp. 2758--2763.

\bibitem{guo2015multi}
------, ``Multi-agent plan reconfiguration under local {LTL} specifications,''
  \emph{The International Journal of Robotics Research}, vol.~34, no.~2, pp.
  218--235, 2015.

\bibitem{kloetzer2010automatic}
M.~Kloetzer and C.~Belta, ``Automatic deployment of distributed teams of robots
  from temporal logic motion specifications,'' \emph{IEEE Transactions on
  Robotics}, vol.~26, no.~1, pp. 48--61, 2010.

\bibitem{ulusoy2013optimality}
A.~Ulusoy, S.~L. Smith, X.~C. Ding, C.~Belta, and D.~Rus, ``Optimality and
  robustness in multi-robot path planning with temporal logic constraints,''
  \emph{The International Journal of Robotics Research}, vol.~32, no.~8, pp.
  889--911, 2013.

\bibitem{ulusoy2014optimal}
A.~Ulusoy, S.~L. Smith, and C.~Belta, ``Optimal multi-robot path planning with
  ltl constraints: guaranteeing correctness through synchronization,'' in
  \emph{Distributed Autonomous Robotic Systems}.\hskip 1em plus 0.5em minus
  0.4em\relax Springer, 2014, pp. 337--351.

\bibitem{kantaros2017Csampling}
\BIBentryALTinterwordspacing
Y.~Kantaros and M.~M. Zavlanos, ``Sampling-based optimal control synthesis for
  multi-robot systems under global temporal tasks,'' \emph{IEEE Transactions on
  Automatic Control}, 2018, (to appear). [Online]. Available:
  \url{https://arxiv.org/pdf/1706.04216.pdf}
\BIBentrySTDinterwordspacing

\bibitem{vardi1986automata}
M.~Y. Vardi and P.~Wolper, ``An automata-theoretic approach to automatic
  program verification,'' in \emph{1st Symposium in Logic in Computer Science
  (LICS)}.\hskip 1em plus 0.5em minus 0.4em\relax IEEE Computer Society, 1986.

\bibitem{holzmann2004spin}
G.~J. Holzmann, \emph{The SPIN model checker: Primer and reference
  manual}.\hskip 1em plus 0.5em minus 0.4em\relax Addison-Wesley Reading, 2004,
  vol. 1003.

\bibitem{cimatti2002nusmv}
A.~Cimatti, E.~Clarke, E.~Giunchiglia, F.~Giunchiglia, M.~Pistore, M.~Roveri,
  R.~Sebastiani, and A.~Tacchella, ``Nusmv 2: An opensource tool for symbolic
  model checking,'' in \emph{International Conference on Computer Aided
  Verification}.\hskip 1em plus 0.5em minus 0.4em\relax Springer, 2002, pp.
  359--364.

\bibitem{gastin2001fast}
P.~Gastin and D.~Oddoux, ``Fast {LTL} to b{\"u}chi automata translation,'' in
  \emph{International Conference on Computer Aided Verification}.\hskip 1em
  plus 0.5em minus 0.4em\relax Springer, 2001, pp. 53--65.

\end{thebibliography}
